\documentclass{article}

\PassOptionsToPackage{numbers, compress}{natbib}
\usepackage[preprint]{neurips_2025}

\usepackage[utf8]{inputenc} % allow utf-8 input
\usepackage[T1]{fontenc}    % use 8-bit T1 fonts
\usepackage{hyperref}       % hyperlinks
\usepackage{url}            % simple URL typesetting
\usepackage{booktabs}       % professional-quality tables
\usepackage{amsfonts}       % blackboard math symbols
\usepackage{nicefrac}       % compact symbols for 1/2, etc.
\usepackage{microtype}      % microtypography
\usepackage{xcolor}         % colors

\usepackage{booktabs}
\usepackage{multirow}
\usepackage{colortbl}
\usepackage{xcolor}

\usepackage{tabularx}
\usepackage{mathtools}
\usepackage{todonotes}
\usepackage{soul} %\sethlcolor 
\usepackage{tabu}% http://ctan.org/pkg/tabu
\usepackage{nicematrix}
\usepackage{tikz}
\usepackage{ulem}
\usepackage{cancel}
\usepackage{graphicx}
\usepackage{latexsym}
\usepackage{sidecap}
\usepackage{amsmath}
\usepackage{amssymb}
\usepackage{amsfonts}
\usepackage{verbatim} % for multi-line comments
\usepackage{textcomp}
\usepackage{pifont}
\usepackage{eqparbox}
\usepackage{comment}
\usepackage{enumitem}
\usepackage{caption}
\usepackage{subcaption}
\usepackage{bm}
\usepackage{MnSymbol,bbding,pifont}
\usepackage[most]{tcolorbox} % ,skins,breakable
\usepackage{makecell}
\usepackage{xspace}
\usepackage{algorithm,algpseudocode}
\usepackage{footmisc}
\definecolor{lightergray}{RGB}{230,230,230}
\definecolor{DarkGreen}{RGB}{30,130,30}
\definecolor{improvecolor}{RGB}{204, 255, 204}
\definecolor{backcolor}{RGB}{232, 242, 255}
\definecolor{lightgreen}{RGB}{229,255,204}
\definecolor{darkgreen}{RGB}{0,102,0}
\definecolor{commentcolor}{RGB}{220, 20, 60}

\makeatletter
\NewDocumentCommand{\LeftComment}{s m}{%
  \Statex \IfBooleanF{#1}{\hspace*{\ALG@thistlm}}\(\triangleright\) #2}
\makeatother

\newcommand{\name}{\textsc{TreeSynth}}

\title{\name{}: Synthesizing Diverse Data from Scratch \\via Tree-Guided Subspace Partitioning}

\author{%
  Sheng Wang\thanks{Equal Contribution.}~~, Pengan Chen$^{\ast}$, Jingqi Zhou$^{\ast}$, Qintong Li \\
  The University of Hong Kong \\
  \texttt{\small\{u3009618, cpa2001, u3011211, qtli\}@connect.hku.hk} \\
  \And
  Jingwei Dong \\
  The University of Hong Kong \\
  \texttt{\small djw8906@hku.hk} \\
  \And
  Jiahui Gao \\
  The University of Hong Kong \\
  \texttt{\small ggaojiahui@gmail.com} \\
  \And
  Boyang Xue, Jiyue Jiang \\
  The Chinese University of Hong Kong \\
  \texttt{\small byxue@se.cuhk.edu.hk, jiangjy@link.cuhk.edu.hk} \\
  \And
  Lingpeng Kong, Chuan Wu \\
  The University of Hong Kong \\
  \texttt{\small \{lpk, cwu\}@cs.hku.hk}
}

\begin{document}
\maketitle

\vspace{-12pt}
\begin{abstract}

Model customization necessitates high-quality and diverse datasets, but acquiring such data remains time-consuming and
labor-intensive. 
Despite the great potential of large language models (LLMs) for data synthesis,
% Despite the great potential of leveraging large language models (LLMs) to synthesize training data,
current approaches are constrained by limited seed data, model biases, and low-variation prompts, resulting in limited diversity and biased distributions with the increase of data scales.
To tackle this challenge, we introduce \name{}, a tree-guided subspace-based data synthesis approach inspired by decision trees. 
It constructs a spatial partitioning tree to recursively divide a task-specific full data space (\textit{i.e.}, root node) into numerous atomic subspaces (\textit{i.e.}, leaf nodes) with mutually exclusive and exhaustive attributes to ensure both distinctiveness and comprehensiveness before synthesizing samples within each atomic subspace. 
This globally dividing-and-synthesizing method finally collects subspace samples into a comprehensive dataset, effectively circumventing repetition and space collapse to ensure the diversity of large-scale data synthesis.
% Given a task-specific data space defined by textual descriptions as the root node, \name{}  recursively partitions it into subspaces 
% (\textit{i.e.,} leaf nodes)
% with mutually exclusive and exhaustive attributes using a spatial partitioning tree. This structured decomposition enables systematic sample generation in each subspace, maximizing diversity and comprehensive coverage. 
Furthermore, the spatial partitioning tree enables sample allocation into atomic subspaces, allowing the rebalancing of existing datasets for more balanced and comprehensive distributions.
Empirically, extensive experiments across diverse benchmarks consistently demonstrate the superior data diversity, model performance, and robust scalability of \name{} compared to both human-crafted datasets and peer data synthesis methods, with an average performance gain reaching 10\%.
Besides, the consistent improvements of \name{}-balanced datasets highlight its efficacious application to redistribute existing datasets for more comprehensive coverage and the induced performance enhancement. 
% The code is available in the anonymous repository via \url{https://anonymous.4open.science/r/TreeSynth-EF04}.
The code is available at \url{https://github.com/cpa2001/TreeSynth}.

\end{abstract}

\vspace{-6pt}
\section{Introduction}
\label{sec:intro}
\vspace{-3pt}
With the superior performance, large language models (LLMs), such as OpenAI o1~\citep{o1}, LLaMA-3~\citep{dubey2024llama}, and DeepSeek R1~\citep{guo2025deepseek}, have been deployed for various downstream applications, 
including code copilot~\citep{jiang2024survey}, mathematical reasoning~\citep{ahn2024large}, 
% financial design~\citep{fieberg2025using}.
psychology~\citep{ke2024exploring}, etc.
The success of these models largely depends on the availability of large-scale diverse training datasets. 
However, open-access data are typically drying up~\citep{babbar2019data, mora2024synthetic}, and manual data curation is both time-consuming and labor-intensive~\citep{wang2023self, gilardi2023chatgpt}, hindering its availability.
% while the expenses of manual data curation and labeling preclude large-scale human annotation efforts ~\citep{wang2023self,gilardi2023chatgpt}.
This necessitates a novel approach to continuously generate data that supports the ongoing advancement of LLMs across different domains.

%In recent research, LLMs have been proposed as a synthetic data generator due to their remarkable ability to generate text like humans~\citep{TanLWBJBKL0024,data_synthesis_survey24}. 
% To capture the complex patterns and characteristics in domain-specific data, 

\begin{figure*}[!t]
    \vspace{-12pt}
    \centering
    \hfill
    \subfloat[Temperature Sampling]{
        \includegraphics[width=0.3\linewidth]{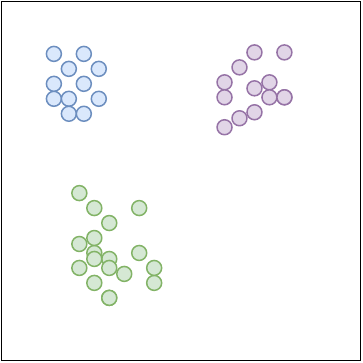}
        \label{fig: t-sampling}
    }
    \hfill
    \subfloat[Evol-Instruct]{
        \includegraphics[width=0.3\linewidth]{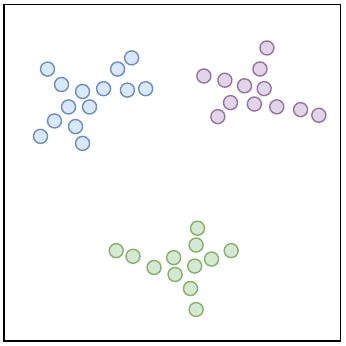}
        \label{fig: evol_sampling}
        }        
    \hfill
    \subfloat[\name{}]{
        \includegraphics[width=0.3\linewidth]{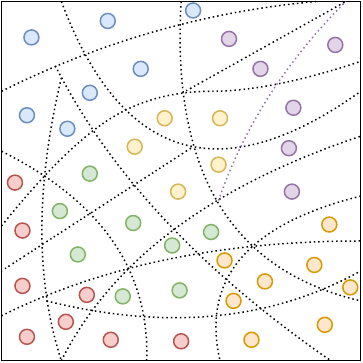}
        \label{fig: tree_sampling}
        }        
    \hfill
    \vspace{-3pt}
    \caption{Intuitive comparison of Temperature Sampling, Evol-Instruct, and \name{}. Temperature Sampling typically generates a specific data distribution induced by model biases, while Evol-Instruct evolves seed data along specified directions. In contrast, \name{} starts from a global perspective by dividing the entire data space into mutually exclusive and complementary subspaces before sampling from each subspace, resulting in a more balanced and diverse dataset with comprehensive coverage.
    }
    \label{fig:comparison}
    \vspace{-12pt}
\end{figure*}

To customize LLMs and further enhance their specific capabilities, synthesizing domain-specific data using their remarkable abilities emerges as a promising solution~\citep{TanLWBJBKL0024, data_synthesis_survey24}.
Pioneering approaches typically paraphrase current datasets~\citep{wang2023self,Source2Synth}, or prompt existing LLMs to reproduce their training data~\citep{ZengXZLC24,shah2024ai}.
% , or focusing on data tailored to specific models~\citep{ChenWYHHMXLHLYS24,li2024forewarned}.
However, due to the inherent model biases and minimal-variation prompts, the generated data often suffers repetition and homogeneity.
To remedy this, increasing sampling temperature~\citep{wang2020contextual} increases data diversity yet reduces quality.
In contrast, attribute-driven approaches (\textit{e.g.}, Persona Hub~\citep{ge2024scaling}) utilize the in-context learning capabilities of LLMs to offer increased diversity and improved quality simultanously. 
Alternatively, evolving new data from existing datasets, represented by Evol-Instruct~\citep{xu2023wizardlm}), augment existing data along different directions to induce diverse generation.
More details about the related works are provided in Appendix.~\ref{sec: related work}.
However, as shown in Figure~\ref{fig:comparison}, from the perspective of data space, these methods typically start from the local distribution (\textit{i.e.}, model biases, seed data, or low-variation prompts) without the global view, hindering their comprehensive coverage.
This raises the following question:

\vspace{-3pt}
\textit{``Is there an automatic solution that starts from a global perspective to fully cover the domain-specific data space for higher diversity?''}
\vspace{-3pt}

\begin{figure*}[!b]
\centering
\includegraphics[width=0.95\linewidth]{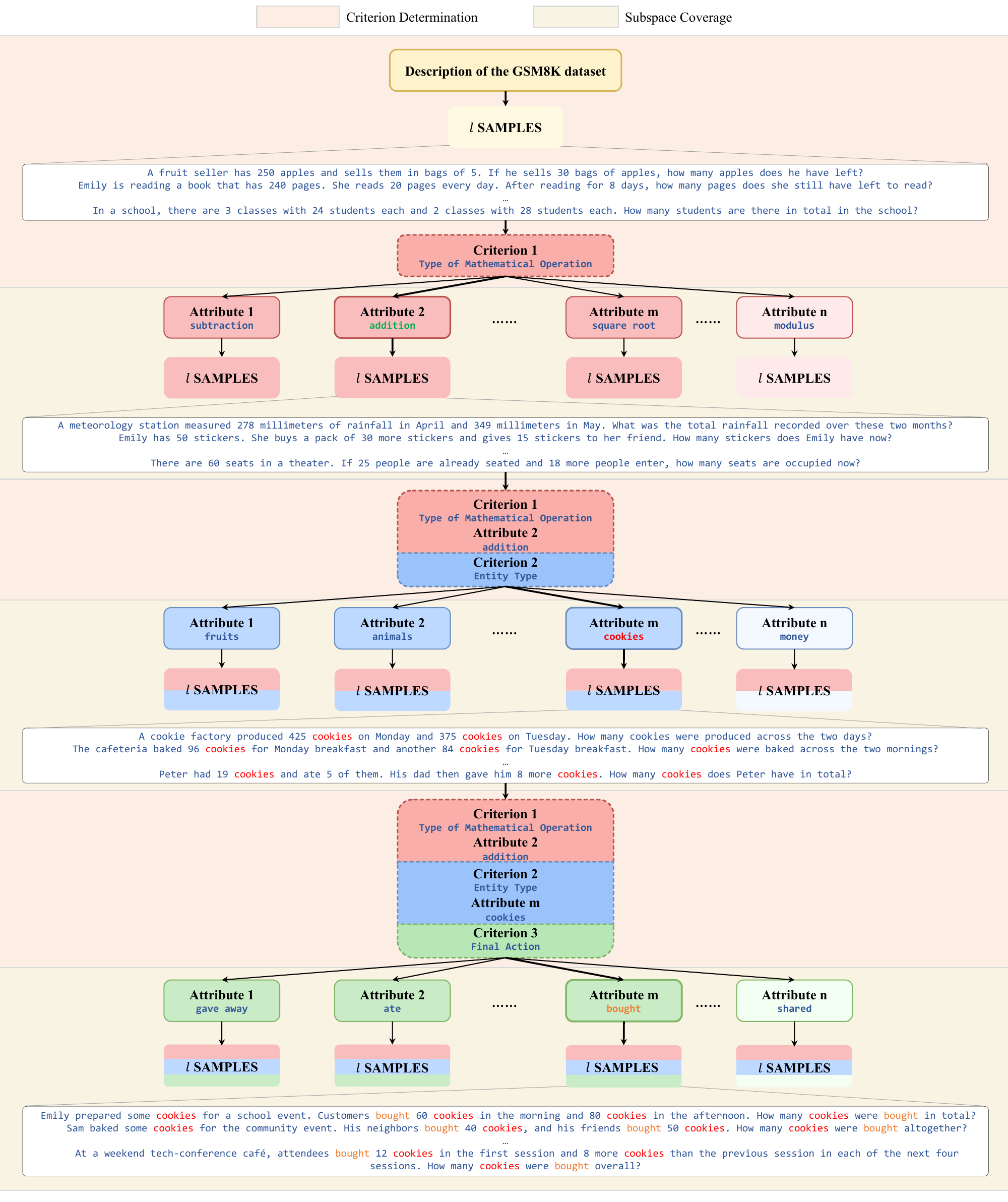}
\caption{
A spatial partitioning tree visualization of \name{}, exemplified through GSM8K-style data synthesis.
}
\label{fig:example_figure}
\end{figure*}

To achieve this objective, we introduce \name{}, a tree-guided subspace-based data synthesis approach inspired by decision trees~\citep{song2015decision}. It consists of two key stages: data space partitioning and subspace data synthesis. 
During the former phase, as illustrated in Figure~\ref{fig:example_figure}, \name{} employs a spatial partitioning tree to recursively divide a task-specific whole data space (\textit{i.e.}, root node defined by textual descriptions) into numerous atomic subspaces (\textit{i.e.}, leaf nodes). These subspaces are characterized by mutually exclusive and exhaustive attribute values to ensure both distinctiveness and diversity. In the subsequent subspace data synthesis phase, samples are generated within each subspace separately, before collecting them as a diverse and comprehensive dataset. By employing this globally divide-and-synthesize methodology, \name{} effectively prevents repetition and space collapse to ensure the diversity and completeness of large-scale data synthesis, successfully avoiding the drawbacks of previous methods.
% guaranteeing both diversity and completeness in large-scale data generation.
Additionally, the spatial partitioning tree enables the allocation of samples into atomic subspaces, thereby allowing the re-balancing of existing datasets for more balanced and comprehensive distributions.
% Aligned with the tree construction, the data space partitioning stage involves criteria determination and subspace coverage steps. 
% % \qt{put the overview figure forward} 
% For any given tree node, an LLM is deployed to generate diverse pivot samples distributed within the associated data space. Subsequently, another LLM, proficient in recognizing distinctions among samples, identifies a core criterion (\textit{i.e.}, criteria) 
% % \qt{i think we can use criterion or criteria, select one} 
% that categorizes these samples into mutually exclusive attribute values, mirroring the exclusivity of the child nodes in the decision tree.
% In the subspace coverage step, potential additional attribute values for this criterion are completed to guarantee that the parent node's data space is entirely inherited by the subspaces, matching the complementarity of the decision tree's child nodes. 
% Starting from the root node, both steps can be recursively applied to construct a comprehensive spatial partitioning tree, hierarchically dividing the entire data space into numerous mutually exclusive and complementary subspaces (\textit{i.e.}, leaf nodes).
% During the subspace data synthesis stage, we generate samples within each leaf node and subsequently combine all data to create a diverse and comprehensive dataset.
% This ensures a large-scale dataset with high diversity and comprehensive coverage
% through hierarchical partitioning to avoid space collapse and repetition.
Extensive experiments with both open-source and closed-source models across diverse benchmarks, spanning mathematical reasoning,
% (\textit{i.e.}, GSM8K~\citep{cobbe2021gsm8k} and MATH~\citep{lightman2023lets})
code generation
% (\textit{i.e.}, MBPP~\citep{austin2021program} and HumanEval~\citep{chen2021evaluating})
and psychology,
% (\textit{i.e.}, SimpleToM~\citep{gu2024simpletom})
demonstrate that \name{} consistently achieves the best downstream performance with superior data diversity compared to both human-crafted datasets and peer data synthesis methods, with the average performance enhancement reaching 10\%, underscoring its great effectiveness and generalization.
Besides, the linear (or even better) performance growth trajectories with increased data volume highlight \name{}'s remarkable robustness and scalability for large-scale data synthesis.
Furthermore, the improved results achieved by applying \name{} to the synthetic datasets demonstrate its efficacious application to redistribute existing datasets for more comprehensive coverage and the induced performance enhancement.
%Moreover, \name{}  enables more efficient distillation of reasoning abilities from reasoning models like OpenAI o1 and DeepSeek R1 to downstream models, enhancing their inferential performance.

%Experimental results demonstrate significant improvements in both data diversity and downstream performance. \name{} consistently achieves superior performance across all benchmarks, outperforming both human-designed datasets and state-of-the-art data synthesis methods, with a maximum enhancement reaching 17.6\%. This achievement stems from the enhanced data diversity in our synthesized data, which surpasses baseline methods by up to 45.2\%.

The main contributions are summarized as follows:
\begin{itemize}[itemsep=2pt, topsep=0pt]
    \item We propose \name{}, a tree-guided subspace-based data synthesis approach, which features mutually exclusive and exhaustive subspace partitioning, effectively circumventing repetition and space collapse to ensure the diversity of large-scale data synthesis.
    \item Extensive experiments consistently highlight \name{}'s superior data diversity, model performance and robust scalability over human-crafted datasets and peer synthesis methods.
    \item The sample allocation of \name{} allows re-balancing existing datasets for more comprehensive coverage, leading to empirically verified downstream performance enhancement.

\end{itemize}

\section{Preliminary Knowledge}

%Due to the simplicity, efficiency, and interpretability, decision tree has been a canonical classification algorithm in machine learning. It can classify any sample into a unique leaf node by recursively traversing from the root node.
%To facilitate this, decision trees exhibit two essential characteristics: the leaf nodes of each subtree are complementary (ensuring that each sample is assigned to at least one leaf node) and non-overlapping (ensuring that each sample is assigned to at most one leaf node).

As a canonical machine learning algorithm, decision trees~\citep{song2015decision} are widely recognized for their simplicity, efficiency, and strong interpretability. For any given sample, the decision tree recursively allocates it to deeper nodes within the hierarchical structure, until it reaches one and only one leaf node. This functionality relies on two essential characteristics: 
(1) All leaf nodes of any sub-tree starting from the root node are mutually complementary, ensuring every sample can be allocated to at least one leaf node.
(2) All leaf nodes of such a sub-tree maintain mutual exclusivity, guaranteeing each sample can be assigned to at most one leaf node.

%Hence, for any specific task, we can conceptualize the training data as the complete space (\textit{i.e.}, the root node), establishing a mapping between the decision tree and the training data space. The decision tree effectively partitions the entire training data space into several leaf nodes, with each leaf node corresponding to a subspace of data characterized by specific attributes.
From a spatial perspective, the root node represents the entire sample space. As the tree delves deeper with each layer of nodes, the space is exhaustively and exclusively divided into multiple subspaces (\textit{i.e.}, child nodes).
Hence, for any given task, we can conceptualize its training data as the entire space (\textit{i.e.}, root node), allowing to establish a mapping between the nodes of decision tree and the training data subspaces. In detail, the decision tree partitions the entire training data space into multiple leaf nodes, with each leaf node corresponding to a data subspace with specific attributes.

This inspires us to leverage the subspace partitioning of decision trees for data synthesis, offering two notable advantages:
(1) \textbf{Diversity}: The exclusivity of leaf nodes ensures the variation across different subspaces, thereby guaranteeing samples diversity.
(2) \textbf{Comprehensive Coverage}: The complementarity and exhaustiveness of leaf nodes ensures the sampling of comprehensive data, preventing sample collapse.

% By controlling the amount of data sampled from each leaf node, we can effectively prevent a large volume of data from collapsing into specific subspaces, thereby achieving data balance.

% This understanding allows us to utilize the subspace partitioning of decision trees for data synthesis, yielding two notable advantages:

% \begin{enumerate}
%     \item \textbf{Enhanced Data Diversity}: The mutual exclusivity of different leaf nodes effectively enhances data diversity.
%     \item \textbf{Data Balance}: By controlling the number of samples drawn from each leaf node, it is possible to prevent data from collapsing into certain subspaces, thereby achieving data balance.
% \end{enumerate}

% 1. The exclusivity of different leaf nodes significantly enhances data diversity.

% 2. By regulating the sampling volume from various leaf nodes, we can mitigate the risk of excessive data concentration in particular subspaces, thereby achieving data balance.

% This spatial partitioning mechanism inspires data synthesis applications with dual advantages:
% (1) \textbf{Enhanced Data Diversity}: The mutual exclusivity between leaf nodes intrinsically enhances data diversity.
% (2) \textbf{Data Balance}: By strategically controlling sampling quantities across different leaf nodes, balanced data distribution can be achieved, effectively preventing data concentration in specific subspaces.

\section{Method}
Inspired by the mapping between a decision tree and data space, we propose \name{}, a tree-guided subspace-based data synthesis approach. It consists of two primary stages: data space partitioning and subspace data synthesis. The first stage generates a spatial partitioning tree $\mathcal{T}$, analogous to the construction of a decision tree, while the second synthesizes data within each atomic subspaces (\textit{i.e.}, leaf node).
%此外，利用spatial partitioning tree $\mathcal{T}$，还可以获知已有数据集在数据空间中的分布情况，进而通过调整各个子空间中的样本数量使得数据分布更加均衡全面。
Beyond data synthesis, we also elaborate how \name{} can re-balance existing datasets, facilitating more comprehensive coverage and induced performance improvement.
% Furthermore, the spatial partitioning tree $\mathcal{T}$ allows analysis of the existing dataset's spatial distribution, enabling optimized sample allocation across subspaces to achieve balanced and comprehensive data coverage.

\subsection{Data Space Partitioning}

\begin{figure*}[!t] 
    \centering
    \hfill
    \subfloat[Tree Perspective]{
        \includegraphics[width=0.65\linewidth]{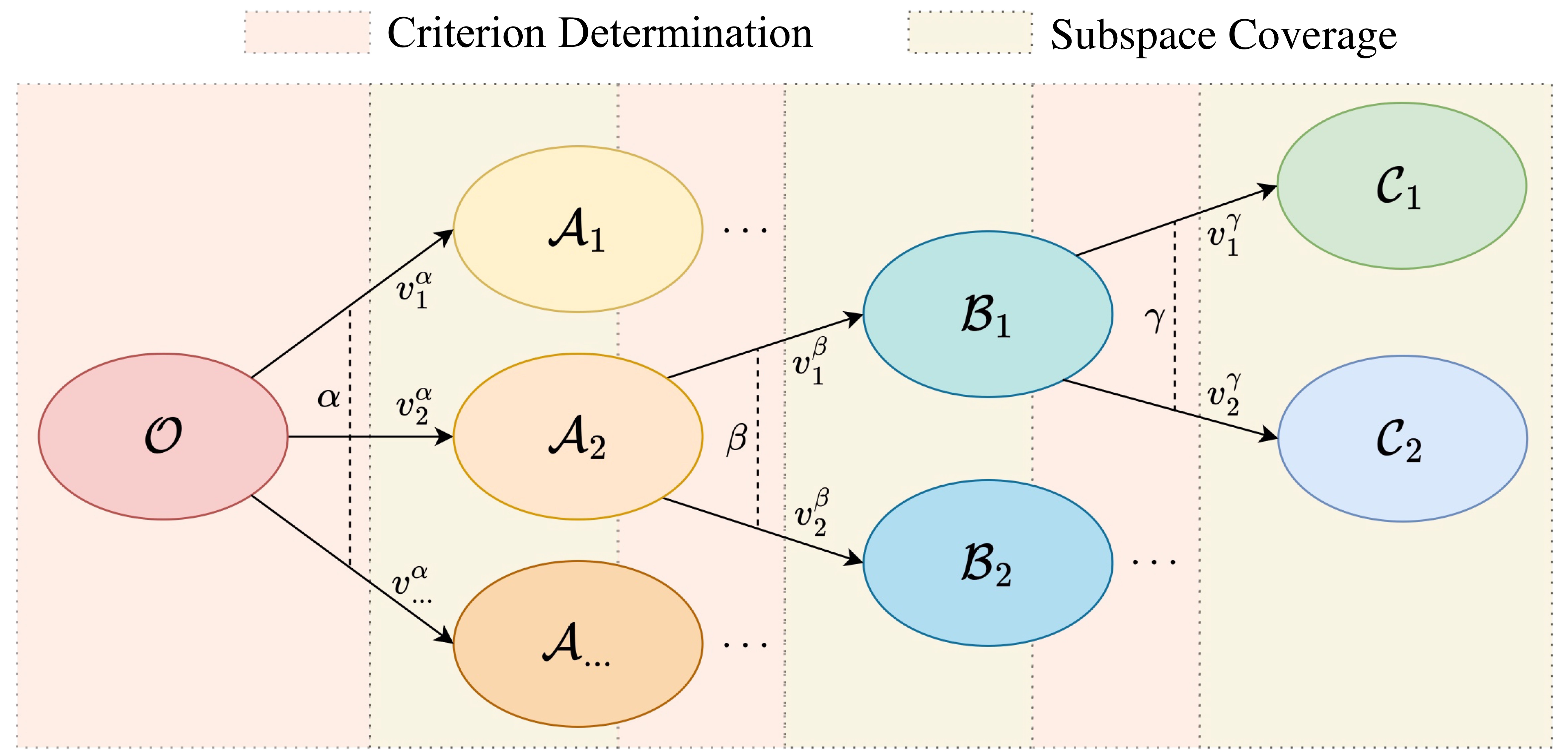}
        \label{fig: tree}
    }
    \hfill
    \subfloat[Spatial Perspective]{
        \includegraphics[width=0.285\linewidth]{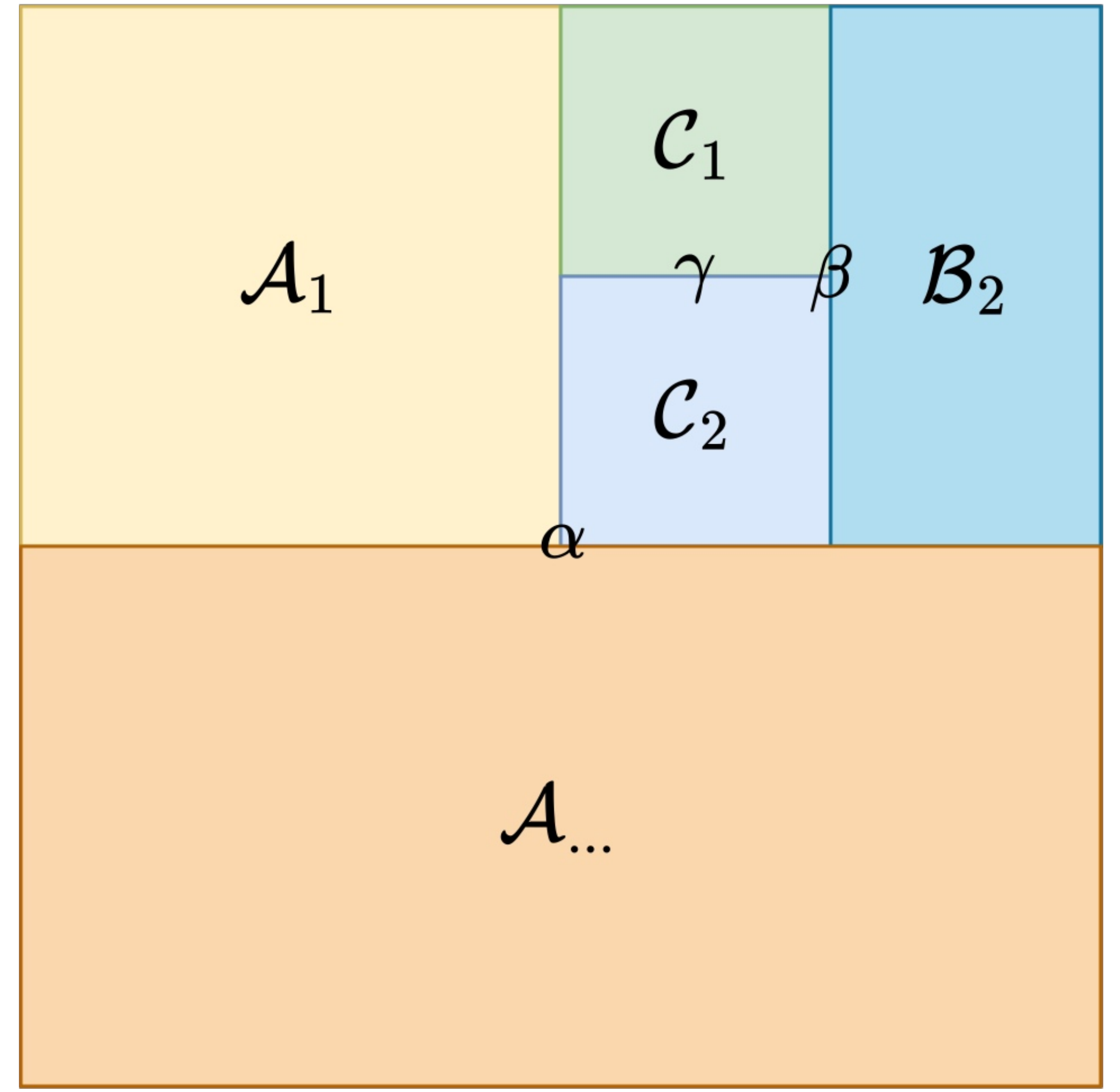}
        \label{fig: space}
        }        
    \hfill
    \vspace{-3pt}
    \caption{Illustration of \name{}. 
    (a) Data space partitioning iterates criterion determination
    and subspace coverage.
    The former identifies the criteria (\textit{e.g.}, $\alpha$, $\beta$, $\gamma$)  and their associated attribute values (\textit{e.g.}, $v_{1}^{\alpha}$, $v_{2}^{\alpha}$, $v_{1}^{\beta}$, $v_{2}^{\beta}$, $v_{1}^{\gamma}$, $v_{2}^{\gamma}$) to divide current nodes (\textit{e.g.}, entire space $\mathcal{O}$, $\mathcal{A}_1$, $\mathcal{A}_2$, $\mathcal{A}_{...}$, $\mathcal{B}_1$, $\mathcal{B}_2$) until reaching the leaf nodes (\textit{e.g.}, atomic subspaces $\mathcal{C}_1$, $\mathcal{C}_2$, ${...}$), while the latter complements potential attribute values (\textit{e.g.}, $v_{...}^{\alpha}$) to ensure  exhaustive coverage of the entire data space. 
    (b) The spatial visualization depicts the mapping between tree nodes and data subspaces, highlighting the mutually exclusiveness and exhaustiveness of the subspaces.}
    \label{fig:illustration}
    \vspace{-9pt}
\end{figure*}

Given any data space $\mathcal{S}$ (\textit{i.e.}, any node in the tree) with its context description $\mathcal{C_S}$, data space partitioning aims to decompose it into multiple subspaces $\mathcal{S}_\text{sub} = \{s_i | i=1,2,...,n\}$. 
As shown in Figure~\ref{fig: tree}, the partitioning process mirrors the construction of decision trees, and comprises two critical steps: Criterion Determination and Subspace Coverage. These steps ensure the mutual exclusivity (\textit{i.e.}, $\forall p \neq q, \, s_p \cap s_q = \emptyset$) and exhaustiveness (\textit{i.e.}, $\bigcup_{i=1}^{n} s_i = \mathcal{S}$) among subspaces, respectively. 
This suggests that each subspace is disjoint, and collectively, they fully encompass the original space.

\textbf{Criterion Determination.}
The essence of this step lies in selecting a criterion $\delta$ that most effectively differentiates data within the space $\mathcal{S}$ so that most data characteristics can be captured with minimal criteria.
Specifically, 
according to the space description $\mathcal{C_S}$ (\textit{e.g.}, "GSM8K-style mathematical questions" as shown in Figure~\ref{fig:example_figure}),
an LLM is firstly deployed to generate $l$ maximally diverse pivot samples $\mathcal{X}=\{x_t | t=1,2,...,l\}$ to approximate the whole space $\mathcal{S}$.
Subsequently, another LLM, proficient in identifying inter-sample distinctions, determines exactly one core criterion $\delta$ (\textit{e.g.}, Type of Mathematical Operation). This criterion $\delta$ optimally partitions $\mathcal{X}$ into mutually exclusive attribute values $V^{\delta}_\mathcal{X}=\{v_j^\delta | j=1,2,...,m\}$ (\textit{e.g.}, addition and subtraction), categorizing each sample $x_t \in \mathcal{X}$ into exactly one attribute value to ensure mutual exclusivity across child nodes.

\textbf{Subspace Coverage.}
Despite the existing attribute values $V^{\delta}_\mathcal{X}$, the mutually exclusive subspace $\mathcal{S}^{\delta}_\mathcal{X}$, derived from partitioning $\mathcal{X}$ with these values, 
may not exhaustively cover the original space $\mathcal{S}$ due to a limited number $l$ of pivot samples. This imposes the risk of non-complementarity among the child nodes.
Hence, subspace coverage is designed to supplement potential attribute values of criterion $\delta$ to comprehensively model the entire data space $\mathcal{S}$.
Specifically, we instruct an LLM to expand the attribute values $V^{\delta}_\mathcal{X}$ to $V^{\delta}_\mathcal{S} = \{v^\delta_i | i=1,2,...,m,m+1,...,n\}$ (\textit{e.g.}, additionally including square root and modulus). The expanded attribute values $V^{\delta}_S$ must be non-overlapping and fully cover the criterion $\delta$. 
Consequently, the exhaustive and exclusive subspaces $\mathcal{S}_\text{sub} = \{s_i \mid i=1,2,\ldots,m,m+1,\ldots,n\}$ can be generated by 
$s_i = \mathcal{C_S} \cap v_i^\delta$ for each  $i$, completely filling the data space $ \mathcal{S}$ .

However, not all criteria can be exhaustively enumerated. For example, if a criterion standard is numerical values of mathematical questions, it contains infinite attribute values (\textit{e.g.}, 0, 1, 2, 3 ...). In such cases, we set a maximum number of attribute values $N$. Once the number of attribute values $n$ exceeds $N$, we refrain from setting individual sub-nodes for each attribute value, and instead establish an infinite node encompassing potential attribute values. Whenever needed, one attribute value is randomly sampled from all potential candidates. This effectively prevents the generation of numerous trivial child nodes, thereby reducing the redundancy of the tree.

\textbf{Spatial Partitioning Workflow.}
Recursively, we apply both criterion determination and subspace coverage steps to construct a complete spatial partitioning tree. 
As illustrated in Figure~\ref{fig:example_figure} and \ref{fig:illustration}, starting from the entire data space $\mathcal{O}$ (\textit{i.e.}, root node $\mathcal{N}_{Root}$) represented by training data description $\mathcal{C_O}$, we first perform criterion determination on $\mathcal{O}$ to identify the optimal criterion $\alpha$ that most effectively distinguishes data within the space $\mathcal{O}$. 
Through subsequent subspace coverage, $\mathcal{O}$ is partitioned into mutually exclusive and exhaustive subspaces $\mathcal{O}_\text{sub}=\{\mathcal{A}_k | k=1,2,...\}$ based on $\alpha$.
Subsequently, the breadth-first search (BFS) algorithm is applied to each subspace $\mathcal{A}_k$ to recursively execute both steps until reaching the maximal depth $d$.
As shown in Figure~\ref{fig: tree}, $ \mathcal{A}_2 $ is further divided into $ \mathcal{B}_1 $ and $ \mathcal{B}_2 $, with $ \mathcal{B}_1 $ subsequently partitioned into the leaf nodes $ \mathcal{C}_1 $ and $ \mathcal{C}_2 $.
Finally, a spatial partitioning tree $\mathcal{T}$ is constructed and decomposes $\mathcal{O}$ into numerous mutually exclusive and complementary atomic subspaces $\mathcal{O}^{*}_\text{Leaf}$, each corresponding to a leaf node $\mathcal{N}^{*}_\text{Leaf}$.
% , such as $\mathcal{C}_1$ and $\mathcal{C}_2$ in Figure~\ref{fig: tree}.
We also present the pseudo code to formularize the whole process in Algorithm~\ref{alg: process}.
\textbf{The mutual exclusivity of leaf nodes intrinsically ensures diversity in the synthesized dataset, while their exhaustiveness guarantees comprehensive coverage of the data space.}
The dual properties effectively prevents data collapse observed in previous data synthesis methods.

\subsection{Subspace Data Synthesis}
The objective of subspace data synthesis stage is to create data within mutually exclusive and complementary atomic subspaces $\mathcal{O}^{*}_\text{Leaf}$ defined by spatial partitioning tree $\mathcal{T}$, ultimately producing a diverse and balanced dataset with comprehensive  space coverage. 
Specifically, for each leaf node $\mathcal{N}^{*}_\text{Leaf}$, we first compile its description along the hierarchical path from the root node $\mathcal{N}_{Root}$ to itself. This path can be formally expressed as $\mathcal{N}_{Root}\rightarrow v^\alpha_i\rightarrow v^\beta_j \rightarrow \cdot\cdot\cdot \rightarrow v^\gamma_k\rightarrow \mathcal{N}^{*}_\text{Leaf}$, where $\{v^\alpha_i, v^\beta_j, \cdot\cdot\cdot, v^\gamma_k\}$ denotes the individual attribute values of parent nodes along the path.
Similar to the generation of pivot samples, we combine both $\mathcal{C_O}$ and the attribute value sequence $\{v^\alpha_i, v^\beta_j, \cdot\cdot\cdot, v^\gamma_k\}$ as the description of $\mathcal{O}^{*}_\text{Leaf}$, and instruct an LLM to generate $N_\text{Leaf}$ samples distributed within its subspace. 
As depicted in Figure~\ref{fig: tree_sampling}, by collecting data generated within all the leaf nodes, we obtain a final dataset with high diversity, balanced distribution, and comprehensive coverage.

\subsection{\name{}-Guided Data Balance}
Beyond data synthesis, \name{} can also be leveraged to optimize existing datasets for improved balance and comprehensiveness.
Given that \name{} synthesizes data from scratch, a spatial partitioning tree $\mathcal{T_D}$ can be constructed solely based on the context description $\mathcal{C_D}$  (\textit{i.e.}, full space $\mathcal{O_D}$) of a given dataset $\mathcal{D}=\{d_u | u=1,2,...,w\}$. Thanks to the mutually exclusive and exhaustive partitioning, every sample $d_u$ can be systematically routed through successive levels of the hierarchy, ultimately landing in a unique leaf node (\textit{i.e.}, atomic subspace).
The distribution of all the samples across leaf nodes reveals the dataset’s coverage pattern within the full space. 
To regulate the distribution across subspaces, a threshold $N_{\text{Sub}}$ is introduced. Subspaces containing more than $N_{\text{Sub}}$ samples are randomly downsampled to reduce overrepresentation, while those with fewer samples are augmented with \name{} to meet the threshold. The integration of all adjusted samples yields a new dataset $\mathcal{D}_{\text{balance}}$ with more  comprehensive coverage and better balance than the vanilla one.

\section{Experiments}
\vspace{-3pt}
\subsection{General Setup}
\label{sec: General Setup}
\vspace{-3pt}
\textbf{Benchmarks.}
\label{sec: Datasets}
To comprehensively evaluate the advantages of \name{}, we compare it against several baselines across diverse benchmarks. For data synthesis, we first apply standard mathematical reasoning and code generation tasks, including GSM8K~\citep{cobbe2021gsm8k}, MATH~\citep{lightman2023lets}, MBPP~\citep{austin2021program} and HumanEval~\citep{chen2021evaluating}, to assess \name{}'s data diversity, model performance improvement, and scalability. Besides, we employ SimpleToM~\citep{gu2024simpletom}, a psychological task, to further examine \name{}'s effectiveness in promoting data balance.
More details are elaborated in Section~\ref{app: Benchmark}.

\textbf{Base Models.}
For generation models to synthesize data with different methods, we employ both open-source
(\textit{i.e.}, \texttt{LLaMA3.3-70B-Instruct}~\citep{dubey2024llama}
% \footnote{ \url{https://www.llama.com/docs/model-cards-and-prompt-formats/llama3_3/}}
and \texttt{Qwen2.5-72B-Instruct}~\citep{qwen2.5})
and closed-source (\textit{i.e.}, \texttt{GPT-4o}\footnote{ \url{https://openai.com/index/hello-gpt-4o/}}) models.
% \footnote{\url{https://huggingface.co/Qwen/Qwen2.5-72B-Instruct}} 
% to power \name{} for generating spatial partitioning trees and synthesize SFT data from subspaces.
% We also adopt \texttt{DeepSeek-R1-Distill-Llama-70B}~\citep{guo2025deepseek} to generate training data with explicit reasoning chains.
To compare the performance of \name{} and baselines, we fine-tune two popular open-source foundation LLMs (\textit{i.e.}, \texttt{LLaMA3.1-8B}~\citep{dubey2024llama} and \texttt{Qwen2.5-7B}~\citep{qwen2.5}) on the perspective generated data.
These models are chosen for their leading performance and popularity.

\textbf{Baselines.} \label{sec: Baselines}
With standard Zero-Shot and Few-Shot performance as reference, we evaluate the effectiveness of \name{} by comparing it with two categories of baselines. 
The first category comprises human-curated datasets (\textit{i.e.}, Vanilla Data): the training sets of GSM8K (7,473 samples) and MATH (7,500 samples) for mathematical reasoning, and Code Alpaca~\citep{codealpaca} (2,689 samples\footnote{Only Python-related samples are retained, aligning with HumanEval and MBPP.}) for code generation.
The second category consists of LLM-synthesized training data, covering three representative methods: Temperature Sampling~\citep{wang2020contextual}, seed-driven method (\textit{i.e.}, Evol-Instruct~\citep{xu2023wizardlm}), and attribute-driven method (\textit{i.e.}, Persona Hub~\citep{ge2024scaling}). 
Each method synthesizes 100k samples in the styles of GSM8K, MATH, and Code Alpaca, and 40k samples in SimpleToM style.
Further details on the baselines and implementation are provided in Appendix~\ref{app: Baselines} and \ref{sec: Implementation Details.}, respectively.
% In order to align with the benchmarks, we focus solely on the Python-related tasks from Code Alpaca, given that both HumanEval and MBPP consist entirely of Python tasks.

\subsection{Main Results}\label{sec: main results}
\vspace{-3pt}

\begin{figure}[h]
  \vspace{-6pt}
  \centering
  \begin{subfigure}[b]{0.325\textwidth}
    \includegraphics[width=\textwidth]{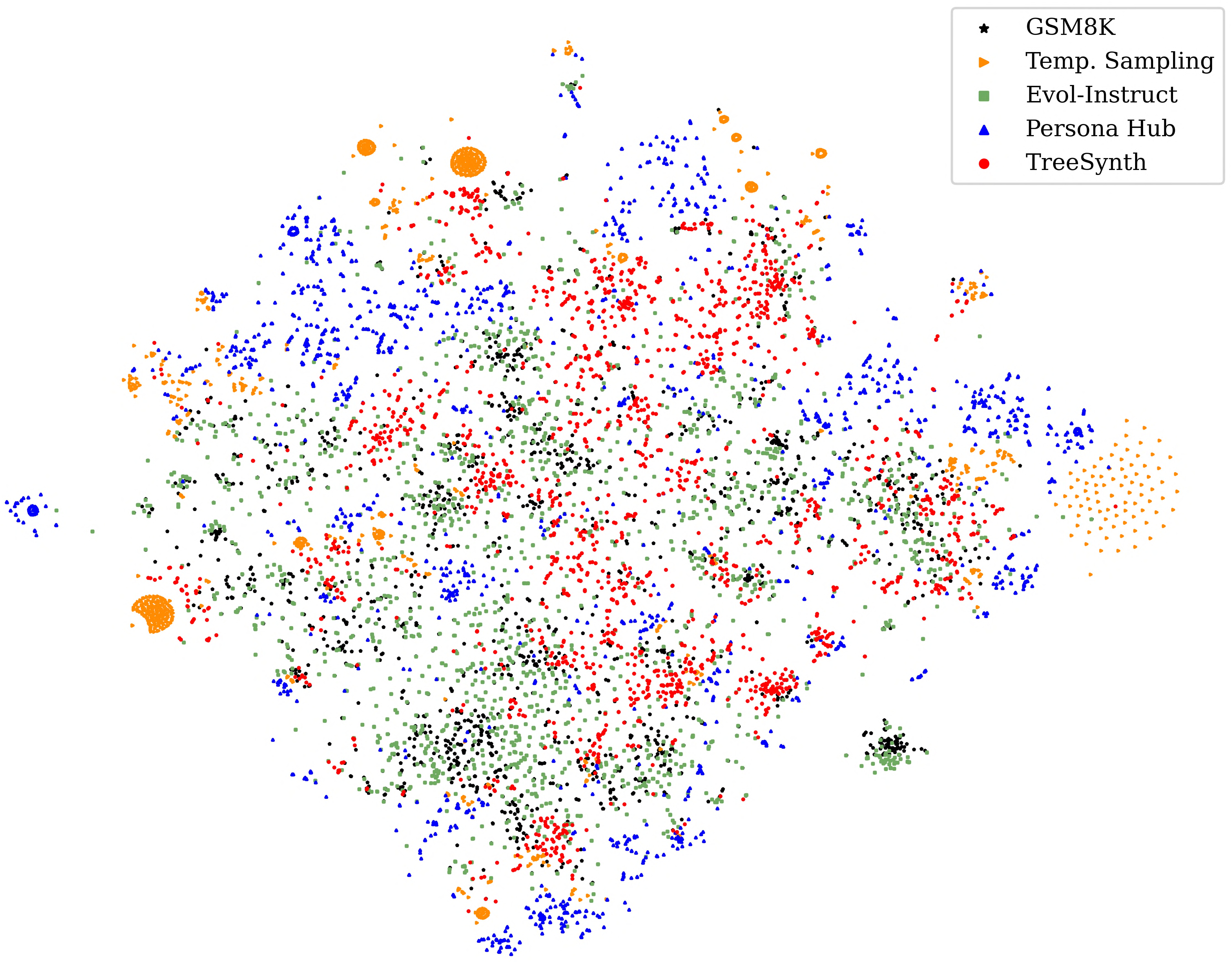}
    \caption{GSM8K}
    \label{fig: GSM8K tsne}
  \end{subfigure}
  \hfill
  \begin{subfigure}[b]{0.325\textwidth}
    \includegraphics[width=\textwidth]{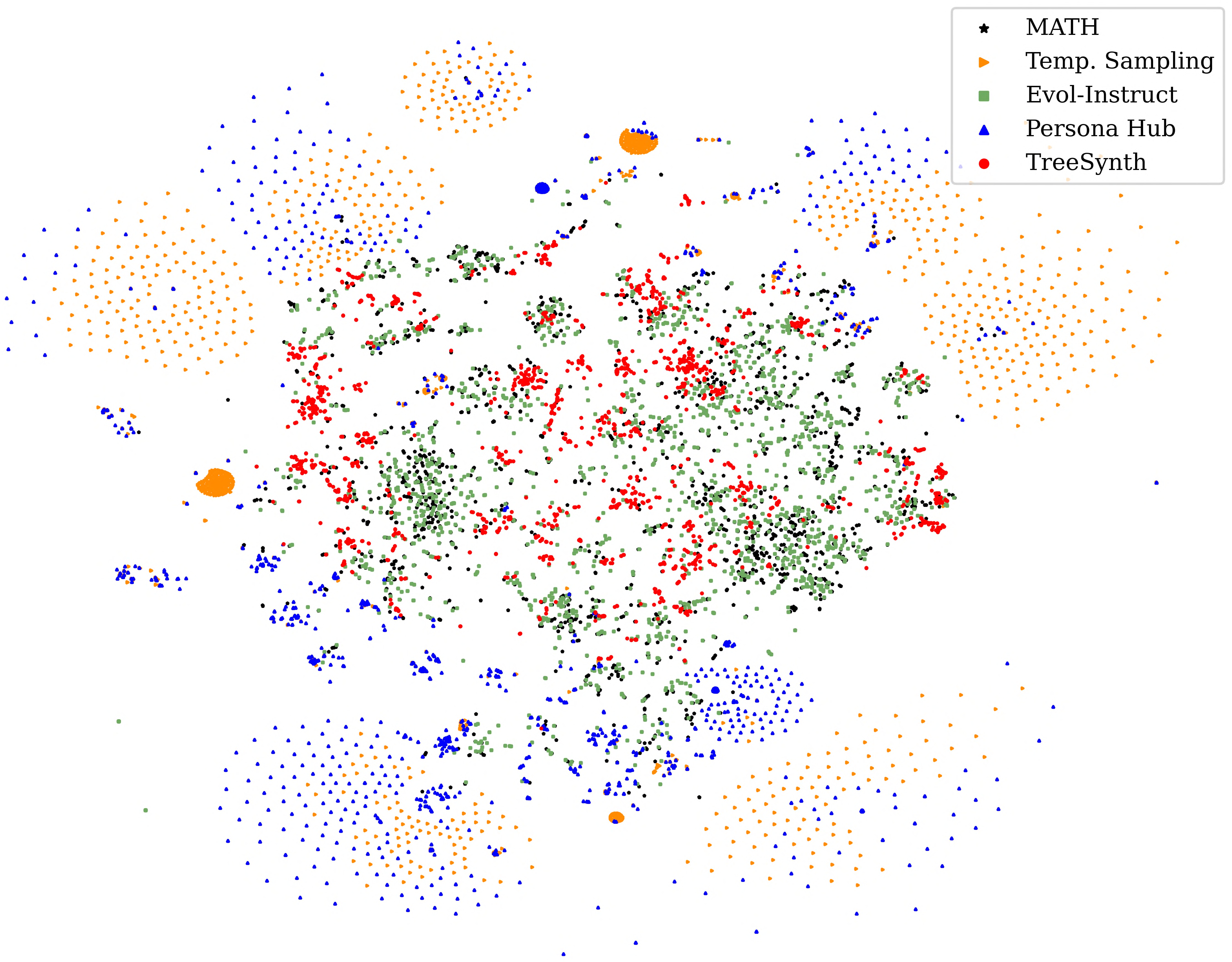}
    \caption{MATH}
    \label{fig: MATH tsne}
  \end{subfigure}
  \hfill
  \begin{subfigure}[b]{0.325\textwidth}
    \includegraphics[width=\textwidth]{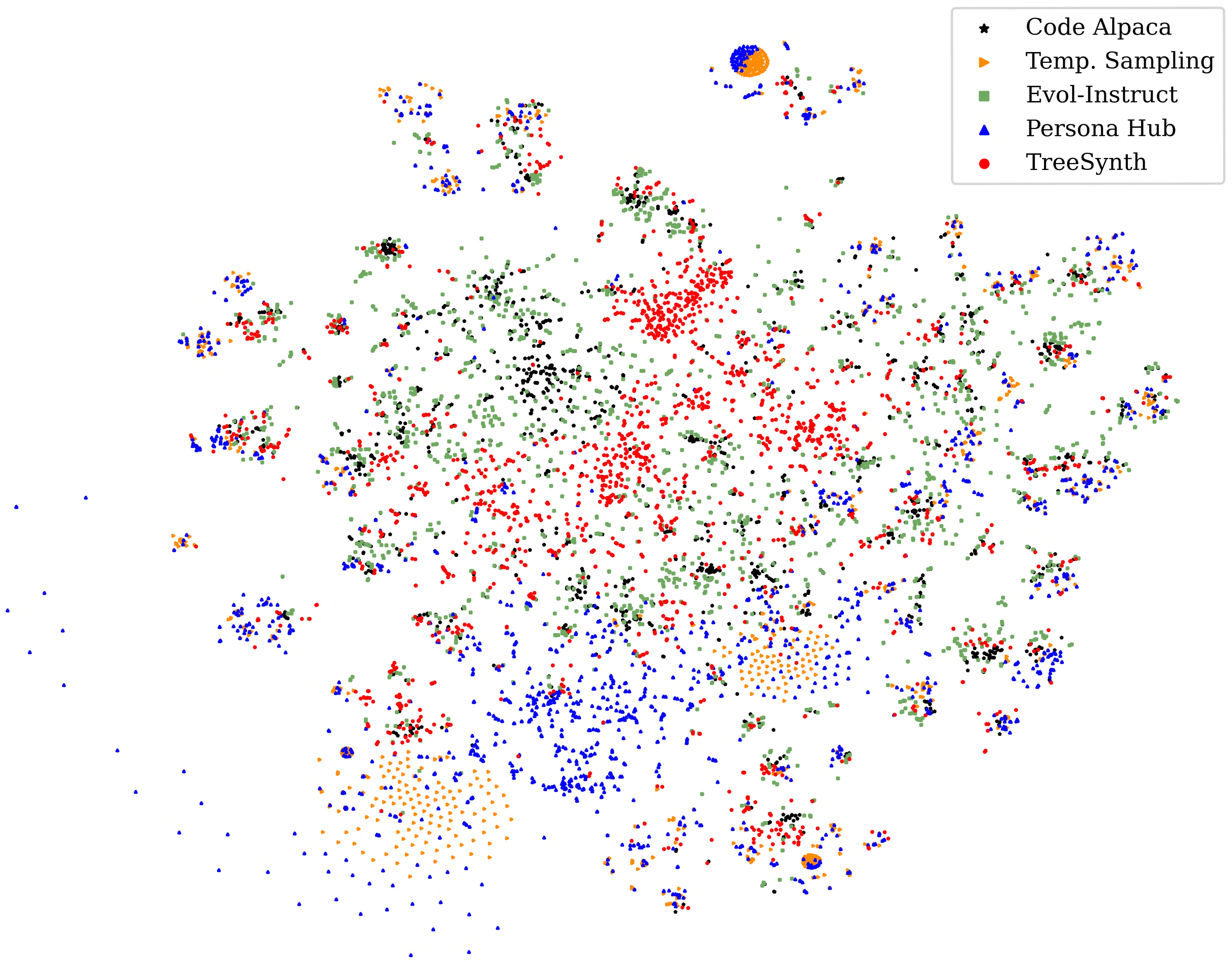}
    \caption{Code Alpaca}
    \label{fig: Code tsne}
  \end{subfigure}
  \vspace{-3pt}
  \caption{t-SNE visualization of \texttt{LLaMA3.3-70B-Instruct}-synthesized datasets for various methods across GSM8K, MATH, and Code Alpaca styles.
  }
  \label{fig:tsne-main}
  \vspace{-6pt}
\end{figure}

\textbf{\name{} exhibits substantially better data diversity and more comprehensive coverage across various tasks and models than both human-curated datasets and peer synthetic methods.} 
As shown in Tables~\ref{exp:main-expt-llama}, \ref{exp:main-expt-gpt}, and \ref{exp:main-expt-qwen},
we compare the data diversity of \name{} and peer methods
% with 100k samples generated by each,
driven by 
\texttt{Qwen2.5-72B-Instruct},
\texttt{LLaMA3.3-70B} \texttt{-Instruct}, 
and \texttt{GPT-4o}, respectively.
Temperature Sampling consistently yields lower diversity than vanilla training datasets.
Although Evol-Instruct and Persona Hub show improvements over vanilla ones in some cases, they generally deteriorate on MATH benchmark, suggesting their limited robustness.
Compared to all the baselines, \name{} consistently achieves the best diversity across almost all the benchmarks and generation models.
Notably, the \texttt{GPT-4o}-powered \name{} exhibits diversity enhancements of 12.5\%, 25.0\%, and 34.5\% over the vanilla GSM8K, MATH, and Code Alpaca datasets, respectively.
In addition, we generate sentence embeddings\footnote{
We utilize the popular 
all-mpnet-base-v2 model, available at \url{https://huggingface.co/sentence-transformers/all-mpnet-base-v2}.} of instructions synthesized by various methods powered by \texttt{LLaMA3.3-70B-Instruct}, and visualize their distributions via t-SNE method~\citep{van2008visualizing} in Figure~\ref{fig:tsne-main}.
Apparently, Temperature Sampling and Persona Hub exhibit concentrated distributions in limited subspaces, revealing severe diversity constraints stemming from inherent biases. Evol-Instruct's distributions mirror those of vanilla datasets, demonstrating strong dependence on source data. In contrast, \name{}, strategically partitioning the full space from a global perspective and synthesizing data within subspaces, eliminates inherent model biases and transcends source data limitations. 
Briefly, both the observation on diversity indicators and visualization confirm the efficacy of \name{} in producing diverse and comprehensive datasets across various domains and models, effectively alleviating inherent model biases and constraints imposed by initial datasets.
% 's tree-guided space partitioning scheme

% \subsection{Fixed Data Size Downstream Performance}
\textbf{Models trained on \name{} data consistently outperform those trained on both human-crafted datasets and synthetic baselines across all the tasks, foundation and generation models.} 
Specifically, for a fair comparison of different methods, we train models using randomly sampled subsets from each synthetic dataset, matching the sizes of the corresponding vanilla training sets.
As shown in Tables~\ref{exp:main-expt-llama}, \ref{exp:main-expt-gpt} and \ref{exp:main-expt-qwen}, all synthetic methods—Temperature Sampling, Evol-Instruct, and Persona Hub—generally surpass human-curated datasets on average, highlighting the limitations of manual data construction.
Thanks to the seed-driven and attribute-driven design, Evol-Instruct and Persona Hub further outperform Temperature Sampling, aligning with the claims in their original works~\citep{xu2023wizardlm, ge2024scaling}.
More microscopically, these methods, however, do not yield stable improvements across all benchmarks. For instance, in Table~\ref{exp:main-expt-llama}, their performance on the HumanEval benchmark falls below that of the vanilla dataset, indicating limited robustness. In contrast, \name{} delivers the best performance across all tasks, foundation and generation models than all baselines without exception. Notably, training \texttt{Qwen2.5-7B} on \name{} data synthesized by \texttt{GPT-4o} yields an average performance improvement of over 10\% compared to the original training sets, underscoring the effectiveness and robust generalization capabilities of \name{}.

\begin{table*}[!t]
    % LLaMA
    \centering
    % \scriptsize
    % \setlength{\tabcolsep}{6.5pt}
    % \renewcommand{\arraystretch}{1.15}
    \resizebox{\linewidth}{!}{
    \begin{tabular}{lcccccccc}
    \toprule
    \textbf{Method}  & \textbf{GSM8K}$\uparrow$ & \textbf{Diversity}$\downarrow$ &  \textbf{MATH}$\uparrow$ & \textbf{Diversity}$\downarrow$ &  \textbf{MBPP}$\uparrow$ & \textbf{HumanEval}$\uparrow$ & \textbf{Diversity}$\downarrow$ & \textbf{Avg.}$\uparrow$ \\
    \midrule
    \midrule
    \rowcolor{backcolor}
    \multicolumn{9}{c}{\textit{Foundation Model:} \textsc{LLaMA-3.1 8B}} \\
    \midrule
    Zero-Shot             & 4.85  & -    & 3.54   & -    & 19.8  & 15.85 & - & 11.01 \\
    Few-Shot              & 40.26 & -    & 20.46  & -    & -     & -     & - & - \\
    \midrule
    Vanilla Data          & 58.15 & 0.40 & 19.48  & 0.16 & \underline{46.8}  & \underline{43.29} & 0.29 & 41.93 \\
    Temp.\ Sampling       & 55.42 & 0.44 & 22.08  & 0.37 & 44.6  & 41.46 & 0.33 & 40.89 \\
    Evol-Instruct         & \underline{63.46} & 0.37 & \underline{27.26} & \underline{\textbf{0.15}} & 40.6  & 41.46 & \textbf{0.22} & \underline{43.20} \\
    Persona Hub           & 61.41 & \textbf{0.35} & 23.78 & 0.34 & 45.8  & 40.24 & 0.26 & 42.81 \\
    \textbf{\name{}}        & \textbf{69.45} & \underline{0.36} & \textbf{27.52} & \underline{\textbf{0.15}} & \textbf{50.2} & \textbf{48.17} & \underline{0.23} & \textbf{48.84} \\
    \midrule
    \midrule
    \rowcolor{backcolor}
    \multicolumn{9}{c}{\textit{Foundation Model:} \textsc{Qwen-2.5 7B}} \\
    \midrule
    Zero-Shot             & 54.97 & -    & 54.38  & -    & 11.2  & 54.88 & - & 43.86 \\
    Few-Shot              & 67.40 & -    & 47.58  & -    & -     & -     & - & - \\
    \midrule
    Vanilla Data          & 68.76 & 0.40 & 47.68  & 0.16 & 53.4  & \underline{77.44} & 0.29 & 61.82 \\
    Temp.\ Sampling       & 69.67 & 0.41 & 61.70  & 0.37 & 54.8  & 76.83 & 0.33 & 65.75 \\
    Evol-Instruct         & 75.13 & 0.37 & 59.60  & \underline{\textbf{0.15}} & 55.8  & 76.83 & \textbf{0.22} & 66.84 \\
    Persona Hub           & \underline{82.79} & \textbf{0.35} & \underline{61.98} & 0.34 & \underline{58.8} & 75.00 & 0.26 & \underline{69.64} \\
    \textbf{\name{}}        & \textbf{85.44} & \underline{0.36} & \textbf{63.28} & \underline{\textbf{0.15}} & \textbf{59.4} & \textbf{78.05} & \underline{0.23} & \textbf{71.54} \\
    \bottomrule
    \end{tabular}
    }
    \vspace{-3pt}
    \caption{Model performance and data diversity comparison of various methods with \texttt{LLaMA3.3-70B- Instruct}-powered data synthesis across two foundation models and multiple benchmarks. 
    ``Zero-Shot'' and ``Few-Shot'' exhibit the base performance of foundation models.
    ``Temp. Sampling'' is abbreviated from ``Temperature Sampling''.
    ``Vanilla Data'' denotes the original GSM8K and MATH training sets, and the Code Alpaca Python subset for HumanEval and MBPP. 
    ``Diversity'' is measured by cosine similarity, where lower values indicate greater diversity.
    Bold and underlining indicate the best and second-best indicators, respectively.
    ``Avg.'' means the average of the performance scores across all the benchmarks.
    }
    \label{exp:main-expt-llama} 
    \vspace{-12pt}
\end{table*}

\begin{table*}[!ht]
% \vspace{-12pt}
    \centering
    % \scriptsize 
    % \setlength{\tabcolsep}{6.5pt}
    % \renewcommand{\arraystretch}{1.15}
    \resizebox{\linewidth}{!}{
    \begin{tabular}{lcccccccc}
    \toprule
    \textbf{Method}  & \textbf{GSM8K}$\uparrow$ & \textbf{Diversity}$\downarrow$ &  \textbf{MATH}$\uparrow$ & \textbf{Diversity}$\downarrow$ &  \textbf{MBPP}$\uparrow$ & \textbf{HumanEval}$\uparrow$ & \textbf{Diversity}$\downarrow$ & \textbf{Avg.}$\uparrow$ \\
    \midrule
    \midrule
    \rowcolor{backcolor}
    \multicolumn{9}{c}{\textit{Foundation Model:} \textsc{ LLaMA-3.1 8B}} \\
    \midrule
    Zero-Shot             & 4.85  & -    & 3.54         & -    & 19.8  & 15.85 & -  & 11.01 \\
    Few-Shot              & 40.26 & -    & 20.46         & -    & -     & -    & -  & - \\
    \midrule
    Vanilla Data          & 58.15 & 0.40 & 19.48         & \underline{0.16} & \underline{46.8} & 43.29 & 0.29 & 41.93 \\
    Temp. Sampling        & 54.97 & 0.45 & 24.28         & 0.29 & 44.8  & 45.73 & 0.32 & 42.45 \\
    Evol-Instruct         & 61.03 & 0.39 & 24.58         & 0.19 & 45.2  & \underline{49.39} & \underline{0.25} & 45.05 \\
    Persona Hub           & \underline{63.38} & \underline{\textbf{0.35}} & \underline{27.74} & 0.28 & 45.2 & 45.12 & 0.29 & \underline{45.36} \\
    \textbf{\name{}}        & \textbf{66.72} & \underline{\textbf{0.35}} & \textbf{30.34} & \textbf{0.12} & \textbf{50.8} & \textbf{50.00} & \textbf{0.19} & \textbf{49.46} \\
    \midrule
    \midrule
    \rowcolor{backcolor}
    \multicolumn{9}{c}{\textit{Foundation Model:} \textsc{Qwen-2.5 7B}} \\
    \midrule
    Zero-Shot             & 54.97 & -    & 54.38     & -    & 11.2 & 54.88 & - & 43.86 \\
    Few-Shot              & 67.40 & -    & 47.58     & -    & -    & -     & - & - \\
    \midrule
    Vanilla Data          & 68.76 & 0.40 & 47.68     & \underline{0.16} & 53.4 & 77.44 & 0.29 & 61.82 \\
    Temp. Sampling        & 80.67 & 0.45 & 62.76     & 0.29 & 59.6 & \underline{\textbf{80.49}} & 0.32 & 70.88 \\
    Evol-Instruct         & 73.16 & 0.39 & 61.10     & 0.19 & 59.2 & \underline{\textbf{80.49}} & \underline{0.25} & 68.49 \\
    Persona Hub           & \underline{83.24} & \underline{\textbf{0.35}} & \underline{66.22} & 0.28 & \underline{61.6} & 77.44 & 0.29 & \underline{72.12} \\
    \textbf{\name{}}        & \textbf{86.13} & \underline{\textbf{0.35}} & \textbf{66.84} & \textbf{0.12} & \textbf{62.8} & \underline{\textbf{80.49}} & \textbf{0.19} & \textbf{74.06} \\
    \bottomrule
    \end{tabular}
}
\vspace{-3pt}
\caption{
    Model performance and data diversity comparison of various methods with \texttt{GPT-4o}-powered data synthesis across two foundation models and multiple benchmarks. 
    ``Zero-Shot'' and ``Few-Shot'' exhibit the base performance of foundation models.
    ``Temp. Sampling'' is abbreviated from ``Temperature Sampling''.
    ``Vanilla Data'' denotes the original GSM8K and MATH training sets, and the Code Alpaca Python subset for HumanEval and MBPP. 
    ``Diversity'' is measured by cosine similarity, where lower values indicate greater diversity.
    Bold and underlining indicate the best and second-best indicators, respectively.
    ``Avg.'' means the average of the performance scores across all the benchmarks.
    }
\label{exp:main-expt-gpt} 
\end{table*}

\begin{figure*}[!ht]
\vspace{-12pt}
    \centering
    \includegraphics[width=0.95\linewidth]{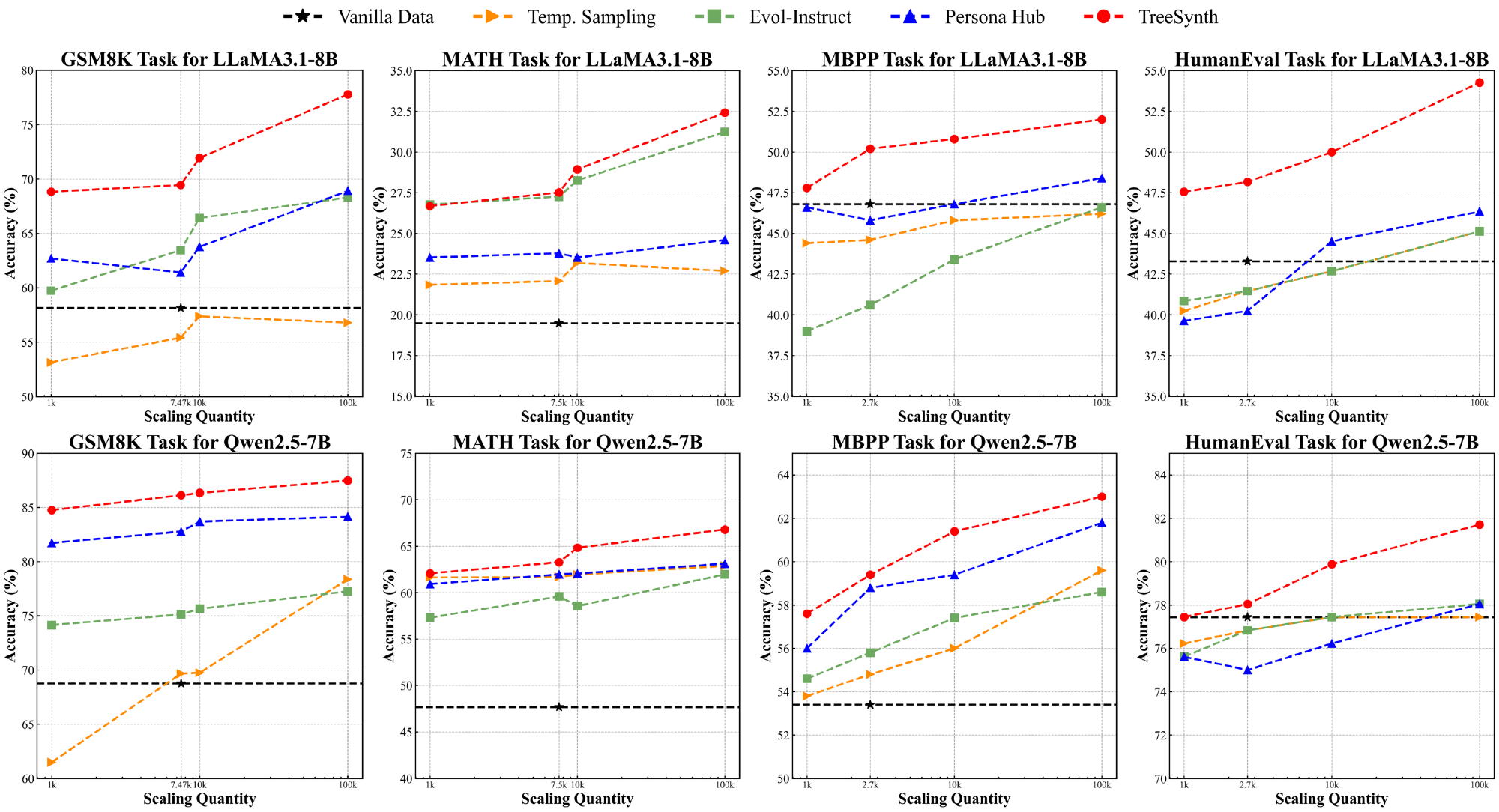}
    \vspace{-6pt}
    \caption{Model performance trends across data scales for different methods powered by \texttt{LLaMA3.3- 70B-Instruct}. ``Temp. Sampling'' refers to Temperature Sampling. ``Vanilla Data'' denotes original GSM8K and MATH training sets, and Code Alpaca Python subset for HumanEval and MBPP.}
    \label{fig:llama_trend_result}
    \vspace{-12pt}
\end{figure*}

\begin{figure*}[!ht]
    \vspace{-12pt}
    \centering
    \includegraphics[width=0.95\linewidth]{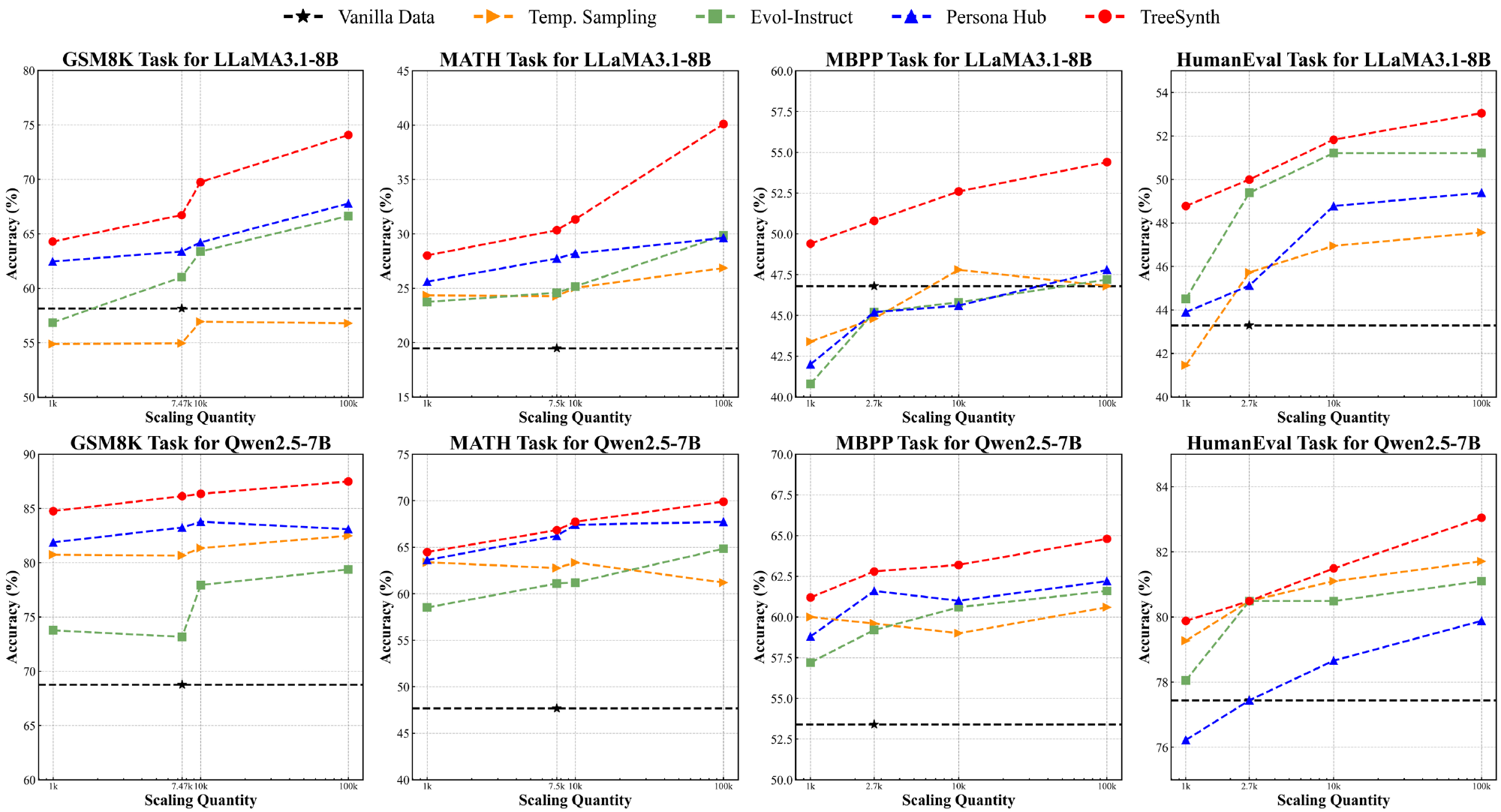}
    \vspace{-6pt}
    \caption{Model performance trends across data scales for different methods powered by \texttt{GPT-4o}. ``Temp. Sampling'' refers to Temperature Sampling, while ``Vanilla Data'' denotes original GSM8K and MATH training sets, and Code Alpaca Python subset for HumanEval and MBPP.}
    \label{fig:GPT_trend_result}
    \vspace{-9pt}
\end{figure*}

% \subsection{Scaling Data Size Downstream Performance}
\textbf{With the global data spatial perspective guided by tree structure, \name{} effectively scales datasets while preserving data quality, suggesting great scalability wherein downstream performance consistently improves with increased data volume.} As shown in Figure~\ref{fig:GPT_trend_result}, \ref{fig:llama_trend_result} and \ref{fig:qwen_trend_result}, we evaluate model performance across synthetic datasets of 1k, 10k and 100k samples, as well as at a scale equivalent to the corresponding vanilla training sets.
Human-curated datasets inherently suffer from limited scalability due to the prohibitive cost of manual annotation. 
Despite exhibiting linear growth trends occasionally, all the baselines (\textit{i.e.}, Temperature Sampling, Persona Hub, and Evol-Instruct) encounter performance saturation with diminishing improvements as dataset volume increases in nearly half of the evaluated settings, and even suffer from degradation in some cases, reflecting their instability. As mentioned above, this can be attributed to the intensification of low-variation prompts, model and seed data biases.
% without a global perspective.
In contrast, \name{} not only remarkably surpasses all the baselines on all data scales, but also exhibits approximately linear (even better) performance growth with increasing data volume, underscoring its superior scalability. 
Besides, with the globally spatial perspective circumventing the local distribution biases, \name{} still exhibits steeper performance growth trajectories beyond the 10k sample scale, indicating its great potential for large-scale data synthesis.
Detailed numerical results are also provided in Table~\ref{tab:gsm-full}, \ref{tab: MATH full}, \ref{tab:mbpp-full} and \ref{tab:humaneval-full} in Appendix~\ref{sec: Complementary Analysis} for more precise reference.
\vspace{-3pt}

\subsection{\name{}-Guided Data Balance}
\vspace{-6pt}
\textbf{Beyond data synthesis, \name{} significantly enhances the distributional balance of existing datasets, effectively improving data diversity and downstream model performance.} 
Specifically, we repeat the \texttt{LLaMA3.3-70B-Instruct}-powered experiments on the \texttt{LLaMA3.1-8B} foundation model in Section~\ref{sec: main results} on the SimpleToM benchmark, but additionally apply the \name{}-guided data balance technique to the synthetic datasets of Temperature Sampling and Persona Hub\footnote{Evol-Instruct is excluded in this section, due to its strong dependence on seed dataset.}. As listed in Table~\ref{exp:SimpleToM table}, \name{} continues to outperform other unbalanced approaches.
Meanwhile, the application of \name{}-guided data balance leads to performance improvements of 7.9\% and 2.9\% for Temperature Sampling and Persona Hub, respectively.
To demonstrate the underlying data distribution more intuitively, Figure~\ref{fig:simpletom-tsne} presents the t-SNE visualization of data spatial distribution for different approaches.
Benefiting from the global perspective, \name{} exhibits comprehensive and well-balanced distribution across the data space. In contrast, samples from Temperature Sampling and Persona Hub predominantly tend to cluster in limited subspaces, revealing insufficient diversity. The application of \name{} substantially enhances their distributional uniformity, enabling comprehensive coverage of the data space.
These results collectively demonstrate that \name{}-guided data balance effectively addresses deficiencies in existing datasets by optimizing their sample distribution, leading to measurable improvements in downstream model performance.

\begin{table*}[!tb]
\centering
\resizebox{0.95\linewidth}{!}{
\begin{tabular}{cccccc}
\toprule
\multirow{2}{*}{\textbf{Performance}} & \multirow{2}{*}{Temp. Sampling} & \multirow{2}{*}{Persona Hub} & \multirow{2}{*}{\name{}} & \multicolumn{2}{c}{\name{}-balanced} \\
\cmidrule{5-6}
                                      &                                 &                              &                        & Temp. Sampling   & Persona Hub     \\
\midrule
\textbf{Accuracy}                     & 78.9                            & 85.7                         & \underline{ 88.0}             & 86.8             & \textbf{88.6}   \\
\textbf{Diversity}                    & 0.37                            & \underline{ 0.33}                   & \underline{ 0.33}             & \underline{ 0.33}       & \textbf{0.32}       \\ 
\bottomrule
\end{tabular}
}
\vspace{-3pt}
\caption{The diversity of data generated by different methods powered by \texttt{LLaMA3.3-70B-Instruct} and its \name{}-Guided Data Balanced version, along with the performance of \texttt{LLaMA3.1-8B} trained on these datasets on the SimpleToM benchmark.
    "Div." stands for Diversity, a metric assessed by computing cosine similarity among data points within the dataset, where lower numerical values directly indicate higher diversity levels.}
\label{exp:SimpleToM table} 
\vspace{-12pt}
\end{table*}

\vspace{-3pt}

\section{Conclusion} \label{sec: conclusion}
\vspace{-6pt}
Targeting synthesizing diverse datasets with LLMs from scratch, we propose a tree-guided subspace-based data synthesis approach, \name{}, which recursively partitions a task-specific full data space into mutually exclusive and exhaustive atomic subspaces before synthesizing and collecting subspace samples into a comprehensive dataset. This globally divide-and-synthesize strategy effectively circumvents repetition and space collapse caused by model biases, seed data, and low-variation prompts in prior methods, promoting the diversity of large-scale data synthesis. Besides, \name{} also facilitates sample allocation into atomic subspaces, enabling re-balancing of existing datasets for more balanced distributions. Compared with both human-crafted datasets and peer data synthesis methods, \name{} consistently achieves superior data diversity, model performance, robust scalability, and data balance efficacy, establishing itself as a promising solution for diverse data synthesis without seed data.

% \section*{Acknowledgments}

\newpage
% \section*{References}
\bibliographystyle{unsrtnat}
\bibliography{references}

%%%%%%%%%%%%%%%%%%%%%%%%%%%%%%%%%%%%%%%%%%%%%%%%%%%%%%%%%%%%

\appendix
\newpage
%\clearpage
\section{Appendix}
\label{sec:appendix}
\subsection{Related Work}\label{sec: related work}
\textbf{Data Synthesis via LLMs.}
% Synthetic data has emerged as a key technique for building large language models due to its costeffectiveness and scalability % The process, however, is labor-intensive and costly.  in specific domains, which can
% Data synthesis generally refers to the generating or labeling of raw data to improve the efficacy of machine learning~\citep{he2008adasyn,bolon2013review,liu2024best}. 
% Traditional synthesis techniques such as rule-based generation~\citep{FengGWCVMH21} struggle to fully capture the complexities of real-world data~\citep{synthsis_survey2024}.
% Furthermore, as LLMs are pre-trained on extensive corpora, they inherently possess rich world knowledge, enabling the distillation of data directly from them. 
% To generate task-related and requirement-compliant datasets, most researches focus on augmentating or imitating existing datasets using LLMs~\citep{WangKMLSKH23, xu2023wizardlm, luo2023wizardmath, luo2023wizardcoder}. 
% However, the scarcity of high-quality, domain-specific datasets limits the applicability of such paradigms. 
% \citet{xu2024magpie} induce LLMs to complete new user queries with only pre-query templates and system prompts to produce alignment data.
% generating new data points aligned with the specified features. also falls in the category of generation from scratch. However, it
% As an effective alternative to time-consuming and labor-intensive human annotation, leveraging LLMs for data synthesis has fueled a growing interest, thanks to their remarkable capability to generate large-scale, high-quality datasets
Owing to their remarkable ability to generate large-scale, high-quality datasets, LLMs are increasingly being explored as an effective alternative to the time-consuming and labor-intensive process of human annotation~\citep{maini2024rephrasing, toshniwal2024openmathinstruct, synthsis_survey2024}.
% 从已有数据中挖掘, 润色相关数据；但是这种受限于已有数据和特定任务
Pioneering studies~\citep{WangKMLSKH23, xu2023wizardlm, taori2023stanford} have showcased the use of LLMs to paraphrase or augment existing instruction datasets into more comprehensive ones.
% 设计任务相关的prompt合成，在a任务，b任务，c任务
Furthermore, recent research has explored data synthesis from scratch in areas such as mathematics~\citep{shah2024ai}, code~\citep{li2024api}, general alignment~\citep{xu2024magpie}, etc. 
% 合成过程不可控，有bias的分布，容易生成简单的数据，遗漏edge cases。
However, uncontrolled synthesis process often exhibits significant biases, favoring easy queries while neglecting more challenging ones~\citep{wang2023let,tong2024dart,li2024forewarned}.
To enhance controllability, \citet{wong2024simplestrat} and \citet{huang2024key} incorporate strata and topic clustering  as guidance, respectively.
% However, the former fixes the data partitioning method and requires human intervention, while the latter fails to further expand the distribution of seed datasets. 
% Our approach distinguishes itself by utilizing LLMs to brainstorm 
% % task-specific, tree-like hierarchical 
% space partitioning from scratch without any human intervention, thereby ensuring data completeness and automation.
Our approach distinguishes itself by constructing a tree-like hierarchical structure from scratch to model the data space for better controllability.

\textbf{Diversity Enhancement in Data Synthesis.}
% Providing variety in responses is often more straightforward compared to creating novel queries.
% For questions with multiple plausible answers, 
% Nevertheless, the homogeneity of instructions still is the bottleneck. 
% failing to continuously improve finetuning performance. 
% A basic approach is high-temperature sampling, which, while increasing variability, often 
% inherits LLMs biases and provides limited domain coverage~\citep{chung2023increasing}. 
% more controlable and interpretable.
% while ~\citep{yu2023metamath} introduces reverse-thinking variations to address the "reverse curse". 
The diversity of training datasets is essential for enhancing model generalization~\citep{raventos2024pretraining}. 
Numerous studies have sought to improve diversity during data synthesis~\citep{gudibande2023false,zhou2024lima,bukharin-etal-2024-data}. 
Specifically, increasing sampling  temperature~\citep{wang2020contextual} can generate more diverse data; however, it often provides limited domain coverage and decreased quality~\citep{chung2023increasing}.
To promote diversity while ensuring quality, recent methods take advantage of the in-context learning capabilities of LLMs through manually designed rules for data evolution~\citep{xu2023wizardlm,yu2023metamath,luo2023wizardmath,luo2023wizardcoder,li2024mugglemath}. 
However, from the perspective of data space, they typically start from the local distribution (\textit{i.e.}, seed data) without the global view, hindering their comprehensive coverage. Besides, synthesized data often becomes repetitive and homogeneous with the increase of data scales due to the inherent model biases.
To address this challenge, various competing methods have emerged, particularly those based on attribute combinations~\citep{samvelyanrainbow,ding2023enhancing,yu2024large,li2024synthetic}, such as persona-driven data synthesis~\citep{ge2024scaling}. However, these approaches typically rely on fixed attribute dimensions curated by LLMs or human knowledge. 
In contrast, our work emphasizes the dynamic construction of a tree structure to iteratively partition the entire domain space, ensuring comprehensive coverage without human intervention.

\textbf{Application of Tree Structure.}
% A tree data structure is a hierarchical approach that organizes and represents data through parent-child relationships.
As a canonical data structure, trees have been applied in various machine learning algorithms.
Specifically, decision trees 
% infer decision rules from data, and 
conceptualize discriminative tasks as a search problem through a tree-like combinatorial problem space~\citep{newell1959report,liaw2002classification,hastie2017elements}. 
% To identify optimal solutions within these tree structures, heuristics guide the search algorithms in determining which states to explore and the order of exploration. 
Besides, \citet{hao2023reasoning,yao2024tree} 
utilize tree-search algorithms,
% to construct reasoning trees by generating and evaluating diverse reasoning paths with search algorithms, 
including breadth-first search, depth-first search, and Monte Carlo Tree Search~\citep{browne2012survey}, to guide multi-step reasoning process for improved reasoning capabilities of LLMs. 
Furthermore, AlphaZero-like tree search learning approach~\citep{feng2023alphazero} leverages a learned value function to guide the decoding process of LLMs, and particularly improves the performance on long-horizon planning. 
In contrast, our method does not rely on tree structures for discrimination tasks or to improve reasoning capabilities. Instead, it focuses on data synthesis through hierarchical tree-like spatial partitioning. 
Besides, the decision rules are not learned but are directly derived from the extensive knowledge inherent in LLMs.

% Our work diverges by specifically utilizing tree structures to generate novel data points. We segment the data space of a particular domain and derive attributes from the most fine-grained partitions, represented by leaf nodes.

\subsection{Pseudo Code of \name{}}
\label{sec: pseudo code }

The pseudo code of \name{} is presented in Algorithm~\ref{alg: process}.

% \subsection{Pseudo Code}
\begin{algorithm}[!ht]
% \scriptsize
\caption{Pseudo Code of \name{}}\label{alg: process}
\begin{algorithmic}[1]
\small
\Require  Context description $\mathcal{C_O}$ of entire data space $\mathcal{O}$, Pivot sample count $l$, Maximum attribute value count $N$, Maximum tree depth $d$

% Spatial partitioning tree $\mathcal{T}$ with leaves $\mathcal{O}^{*}_\text{Leaf}$, 

\Function{criterion\_Determination}{$\mathcal{C_S}$, $l$}
    \State Generate diverse pivot samples $\mathcal{X} = \{x_t\}_{t=1}^l$ via 
    $\texttt{LLM}_\text{pivot}(\mathcal{C_S})$ 
    %\Comment{Maximally diverse pre-samples}
    \State Determine exactly one core criterion $\delta \gets \texttt{LLM}_\text{determine}(\mathcal{X})$ 
    % \Comment{Identify most discriminative criterion}
    \State Obtain mutually exclusive attribute values $V^{\delta}_\mathcal{X} \gets \texttt{Partition}(\mathcal{X}, \delta)$ \\
    %\Comment{Mutually exclusive attribute values}
    \quad~ \Return $\delta$, $V^{\delta}_\mathcal{X}$ 
    %\Comment{Core criterion \& initial partitions}
\EndFunction

\Function{Subspace\_Coverage}{$\mathcal{C_S}$, $\delta$, $V^{\delta}_\mathcal{X}$, $ N$}
    \State Fully cover the criterion $\delta$ with all the attribute values $V_\mathcal{S} \gets \texttt{LLM}_\text{complement}(V_\mathcal{X}, \delta)$ 
    \If{$|V_\mathcal{S}| > N$} 
        \State Create subspace $\mathcal{S}^\text{inf}_\text{sub}(V_\mathcal{S}) $
        \Comment{Randomly sample an attribute value whenever needed.}
    \Else
        \State Create subspace $\mathcal{S}^{i}_\text{sub}$ for each $v_i \in V_\mathcal{S}$
    \EndIf \\
    \quad ~ \Return $\mathcal{S}^{*}_\text{sub}$ 
    %\Comment{Final subspace set}
\EndFunction

\Statex \textcolor{commentcolor}{\LeftComment{Stage 1: Data Space Partitioning}}

% \Statex \textcolor{blue}{\LeftComment{Step 1: Criteria Determination}}
% \Statex \textcolor{blue}{\LeftComment{Step 2: Subspace Coverage}}

\State Initialize root node $\mathcal{N}_{Root}$ (\textit{i.e.}, $\mathcal{O}$ with $\mathcal{C_O}$) with the depth as 0, and BFS queue $Q \gets \{\mathcal{N}_{Root}\}$

\While{$Q \neq \emptyset$ }
% \textbf{and} $d' \leq d$}
    \State Dequeue the first node $\mathcal{N}'$ from $Q$, and obtain its depth $d'$ and context description $\mathcal{C}_{\mathcal{N}'}$
    
    \If{$d' < d$}
        \State $\delta$, $V^{\delta}_\mathcal{X} \gets \textsc{Criterion\_Determination}(\mathcal{C}_{\mathcal{N}'}, l)$
        \Comment{Step 1: Criterion Determination}
        \State $\mathcal{N}^{'*}_\text{sub} \gets \textsc{Subspace\_Coverage}(\mathcal{C}_{\mathcal{N}'}, \delta, V_\mathcal{X}^\delta, N)$
        \Comment{Step 2: Subspace Coverage}
        \State Add $\mathcal{N}^{'*}_\text{sub}$ as the child nodes of $\mathcal{N}'$, and append to the queue $Q \gets Q \bigcup \mathcal{N}^{'*}_\text{sub}  $
    \Else
        \State Mark $\mathcal{N}'$ as a leaf node $\mathcal{N}^{'}_\text{Leaf}$
    \EndIf
\EndWhile

% \State Collect all $\mathcal{N}_\text{Leaf}$ into $O_\text{Leaf}$
% \Ensure Tree $T$ with leaves $O_\text{Leaf}$ satisfies:
% \begin{itemize}
%     \item $\bigcup \mathcal{N}_i = O$ (Completeness)
%     \item $\mathcal{N}_p \cap \mathcal{N}_q = \emptyset$ (Exclusivity) 
%     \item $\forall \mathcal{N} \in O_\text{Leaf}, depth(\mathcal{N}) = d$
% \end{itemize}

\Return Spatial partitioning tree $\mathcal{T}$ with the root node $\mathcal{O}$

\Statex \textcolor{commentcolor}{\LeftComment{Stage 2: Subspace Data Synthesis}}

\State Collect all leaf nodes $ \mathcal{N}^{*}_\text{Leaf}     \gets \{\mathcal{N}_\text{Leaf}^1, \mathcal{N}_\text{Leaf}^2, \cdot\cdot\cdot \} $

\For{each $\mathcal{N}_\text{Leaf}^i$ \textbf{in} $\mathcal{N}^{*}_\text{Leaf}$}
    \State Trace hierarchical parent nodes from the root node $\mathcal{P} \gets \{ \mathcal{N}_{Root}, \cdot\cdot\cdot, \mathcal{N}_\text{Leaf}^i \}$
    \State Obtain the attribute intersections as the leaf node description $\mathcal{C}_{\mathcal{N}_\text{Leaf}^i} \gets  \bigcap_{j \in \mathcal{P} } V_j$
    
    % Initialize attribute sequence: $V \gets \emptyset$
    % \For{each non-root node $\mathcal{N}_i$ in path}
    %     \State $V \gets V \cup \{\mathcal{N}_i$.split\_attribute$\}$
    % \EndFor
    % \State Generate $N_\text{Leaf}$ samples via LLM with:
    % \State \quad Input: Context $C_O$ and constraints $V = \{v^\alpha_i, ..., v^\gamma_k\}$
    % \State \quad Output: Subspace data $D_\text{Leaf}^k \gets \{x_1, ..., x_{N_\text{Leaf}}\}$
    \State Generate diverse leaf samples 
    $\mathcal{D}_{\text{Leaf}}^i = \{x_k\}_{k=1}^N$ via 
    $\texttt{LLM}_\text{sample}(\mathcal{C}_{\mathcal{N}_\text{Leaf}^i})$ 

\EndFor
\State Collect all the leaf samples into the final dataset  $\mathcal{D}_\text{final} \gets \bigcup \mathcal{D}_{\text{Leaf}}^{*}$

% \State Validate coverage: $\bigcup_{k=1}^m D_\text{Leaf}^k = D_\text{final}$
% \State Validate balance: $|D_\text{Leaf}^k| = N_\text{Leaf}\ \forall k \in [1, m]$
% \State Optimize diversity: $\max(\text{heterogeneity}(D_\text{final}))$

\Return $\mathcal{D}_\text{final}$ 
\Comment{with high diversity, good balance, and comprehensive coverage.}

\Statex \textcolor{blue}{\LeftComment{\name{}-Guided Data Balance}}
\Require Dataset $\mathcal{D}=\{d_u | u=1,2,...,w\}$, data description $\mathcal{C_D}$, threshold $N_{\text{Sub}}$

\State Build tree $\mathcal{T_D}$ via Data Space Partitioning using $\mathcal{C_D}$
\State Initialize leaves $\mathcal{L} \gets \{\mathcal{N}_\text{Leaf}^1,...,\mathcal{N}_\text{Leaf}^m\}$

\For{$d_u \in \mathcal{D}$}
    \State Route $d_u$ to leaf $\mathcal{N}_\text{Leaf}^k$ via $\mathcal{T_D}$
\EndFor
%\State Count samples $\{c_1,...,c_m\}$ per leaf

\For{$\mathcal{N}_\text{Leaf}^i \in \mathcal{L}$}
    \State $\mathcal{D}^i \gets$ samples in $\mathcal{N}_\text{Leaf}^i$
    \If{$|\mathcal{D}^i| > N_{\text{Sub}}$}
        \State ${D}^i_{\text{new}} \gets $ Downsample to $N_{\text{Sub}}$ 
    \ElsIf{$|\mathcal{D}^i| < N_{\text{Sub}}$}
        \State ${D}^i_{\text{new}} \gets $ Synthesize $(N_{\text{Sub}} - |\mathcal{D}^i|)$ samples in $\mathcal{N}_\text{Leaf}^i$ via LLMs and add into ${D}^i$
    \EndIf
\EndFor

\State Aggregate $\mathcal{D}_\text{balance} \gets \bigcup \mathcal{D}^i_{\text{new}}$
\Return $\mathcal{D}_\text{balance}$

\end{algorithmic}
\end{algorithm}

\subsection{Benchmarks}
\label{app: Benchmark}
To thoroughly evaluate the performance improvements enabled by \name{}-generated data, we test models across three domains: mathematical reasoning, code generation, and psychology tasks.
The mathematical reasoning tasks use the MATH dataset test set~\citep{lightman2023lets} and GSM8K benchmark~\citep{cobbe2021gsm8k}. Code generation capabilities are assessed through HumanEval~\citep{chen2021evaluating} and MBPP~\citep{austin2021program}. For psychology tasks, we employ the SimpleToM dataset~\citep{gu2024simpletom}.
These benchmarks collectively assess our method's effectiveness across multiple dimensions, with brief descriptions provided below:

\begin{itemize}

    \item[\textbullet] \textbf{GSM8K} evaluates mathematical reasoning capabilities through 8,500 high-quality grade school math problems developed via human expert annotation. The dataset is partitioned into 7,500 training and 1,000 test problems, each requiring 2-8 computational steps combining basic arithmetic operations. 
    
    \item[\textbullet] \textbf{MATH} dataset comprises 12,500 competition-level mathematics problems, with 7,500 designated for training and 5,000 for testing. It requires step-by-step solutions, emphasizing accurate problem decomposition and the generation of formal proofs for multi-step mathematical challenges.

    \item[\textbullet] \textbf{HumanEval} evaluates functional correctness through hand-written programming problems requiring language comprehension, reasoning, algorithms, and mathematics. It comprises 164 problems with function signatures, docstrings, and unit tests (avg. 7.7 per problem). Tasks are manually crafted to avoid data leakage from public code repositories. 

    \item[\textbullet] \textbf{MBPP} evaluates programming competence through entry-level Python functions requiring implementation based on textual specifications and test case verification. It contains 974 crowd-sourced programming problems with functional correctness validation, including a core subset of author-curated solutions with manual quality assurance.

    \item[\textbullet] \textbf{SimpleToM} is designed to evaluate whether LLMs can implicitly apply Theory of Mind (ToM) reasoning to predict behavior and judge the rationality of observed actions in social scenarios. It consists of concise, diverse stories followed by three questions that assess different levels of ToM reasoning: predicting a character's mental state, forecasting their behavior, and judging the rationality of their actions\footnote{\scriptsize{In the SimpleTom-style dataset, data where the protagonist is unaware of hidden information is labeled as negative samples, while data where the protagonist is aware of hidden information is labeled as positive samples. In this paper, all SimpleToM-style data starts by generating negative samples to account for half of the dataset. These are then converted into positive samples using the prompts shown in Figures~\ref{fig:N_to_P}. The combination of positive and negative samples forms a complete dataset.}}. 
\end{itemize}

\subsection{Baselines}
\label{app: Baselines}
\begin{figure*}[t]
\centering
\includegraphics[page=2, width=\linewidth]{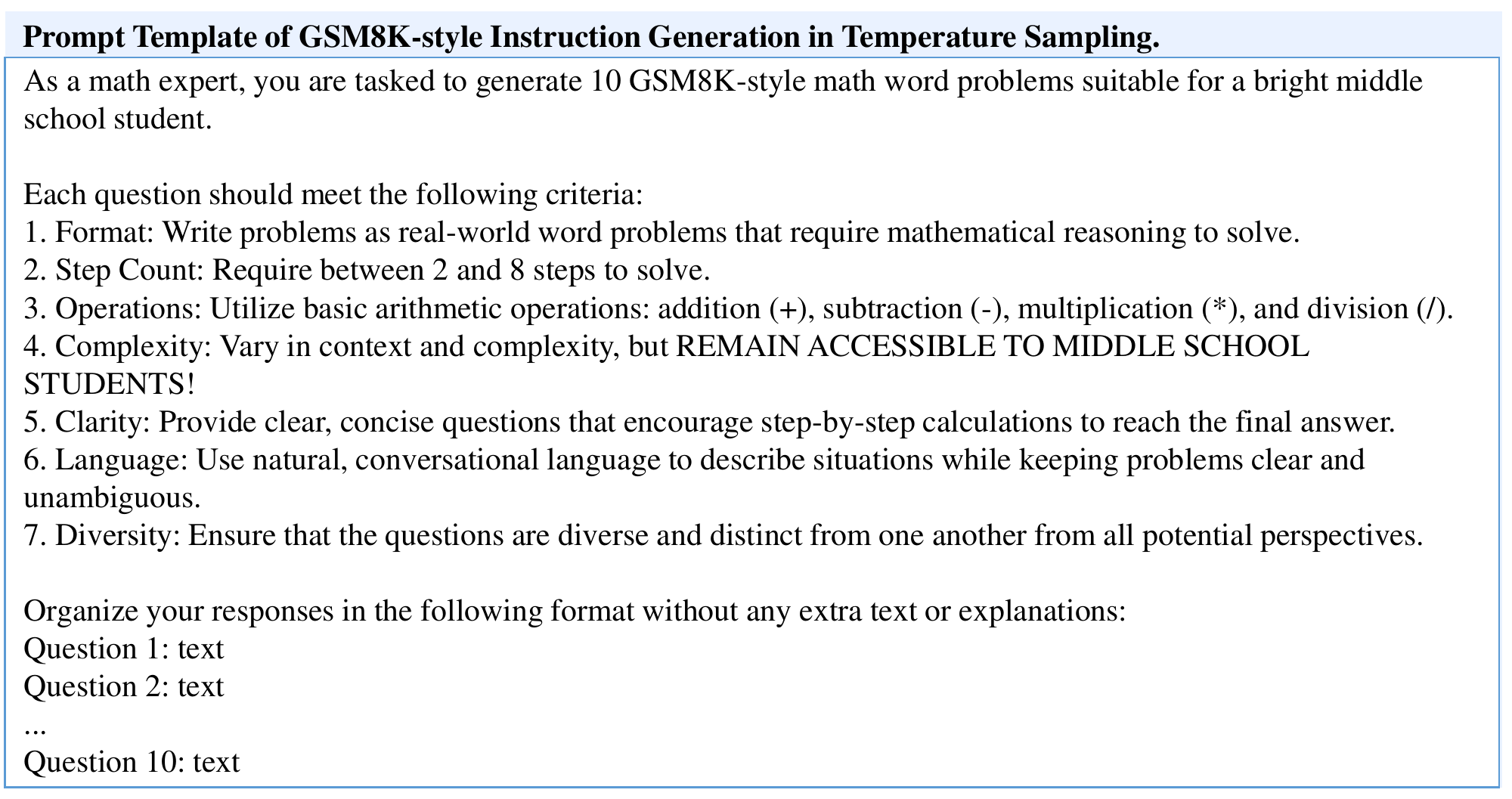}
\caption{
Prompt template of GSM8K-style instruction generation in Temperature Sampling.
}
\label{fig:GSM Temperature Sampling}
\end{figure*}

\begin{figure*}[t]
\centering
\includegraphics[page=3, width=\linewidth]{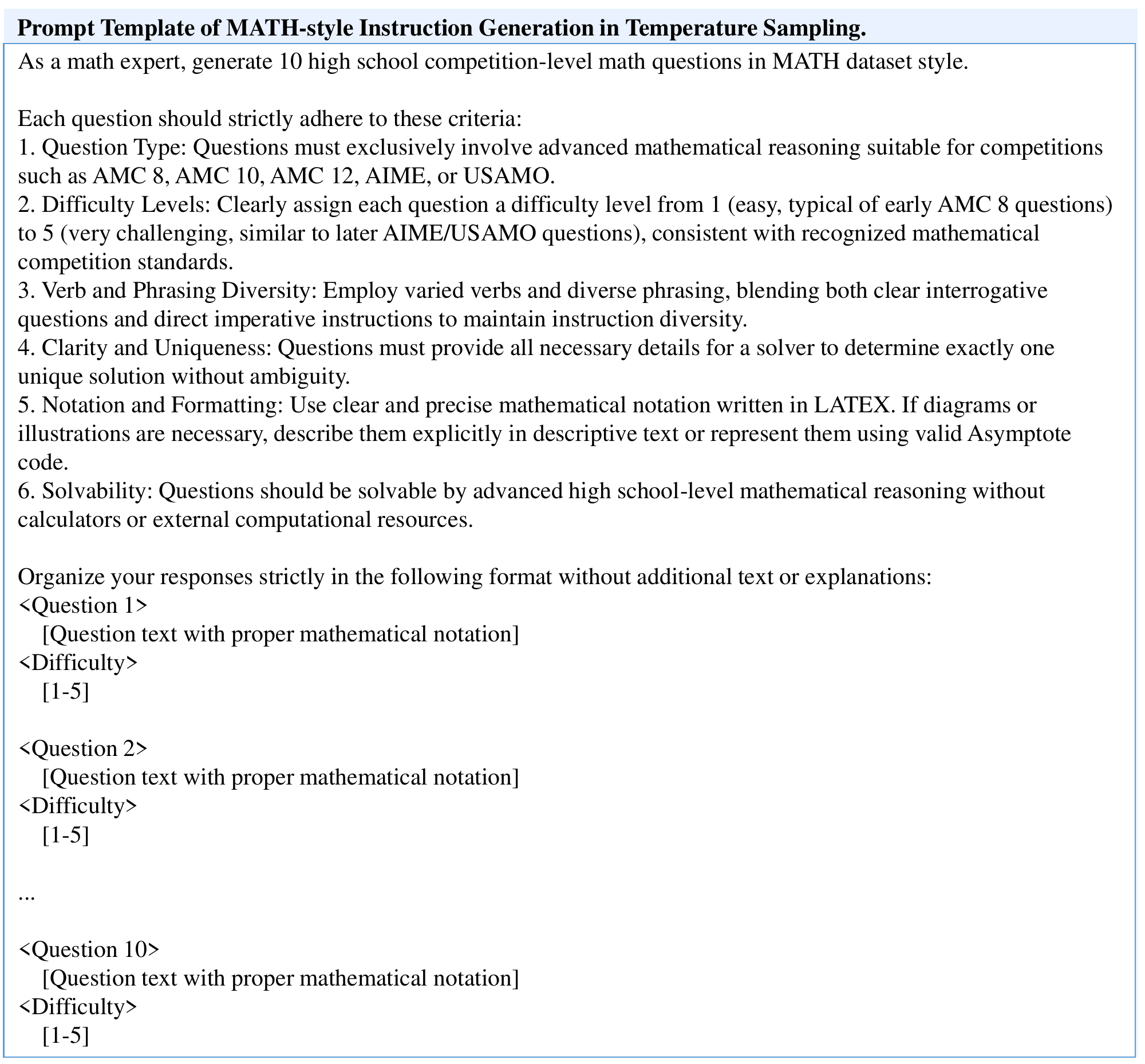}
\caption{
Prompt Template of MATH-style Instruction Generation in Temperature Sampling.
}
\label{fig:MATH Temperature Sampling}
\end{figure*}

\begin{figure*}[t]
\centering
\includegraphics[page=4, width=\linewidth]{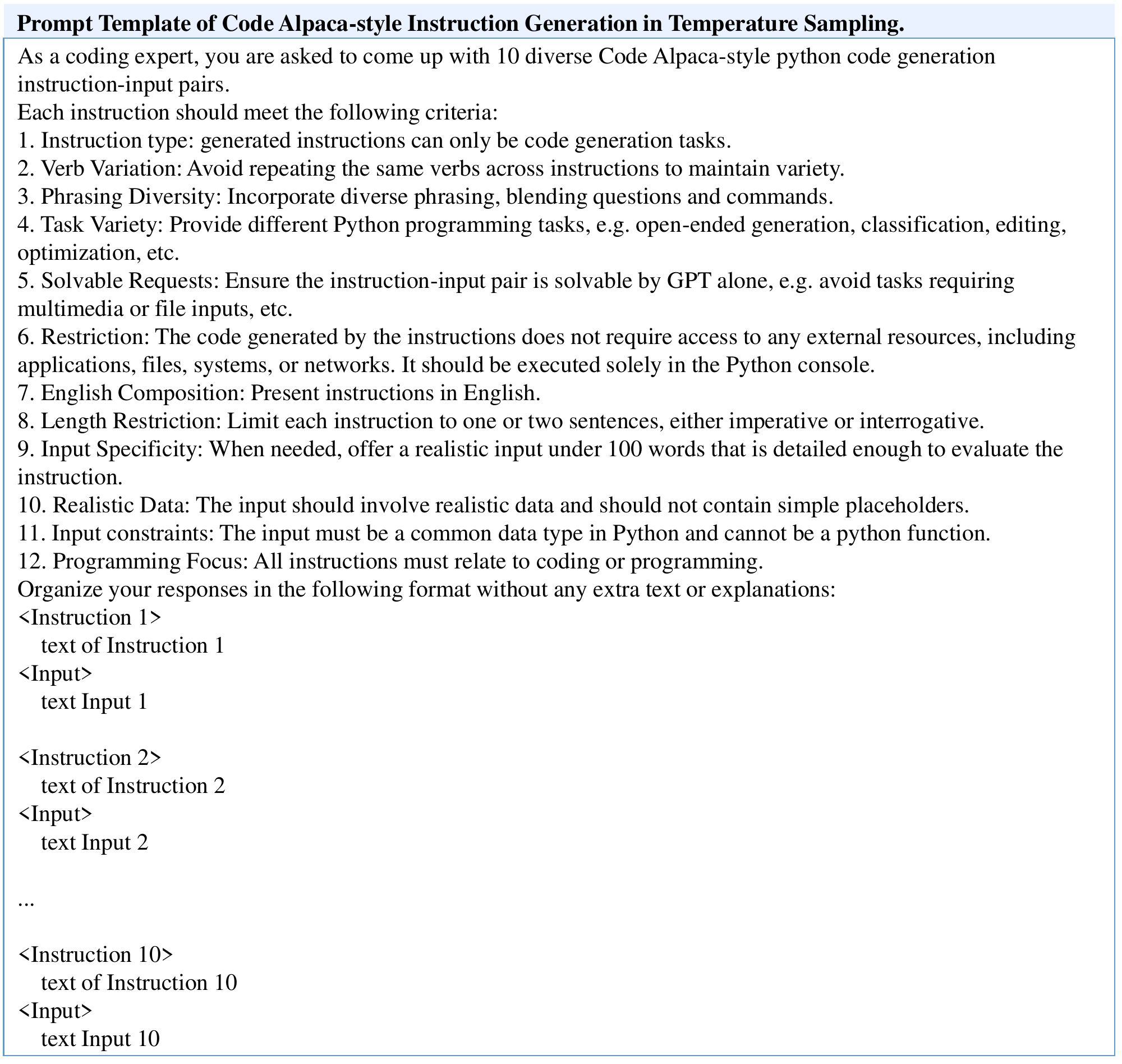}
\caption{
Prompt template of Code Alpaca-style instruction generation in Temperature Sampling.
}
\label{fig:Code Temperature Sampling}
\end{figure*}

\begin{figure*}[t]
\centering
\includegraphics[page=1, width=\linewidth]{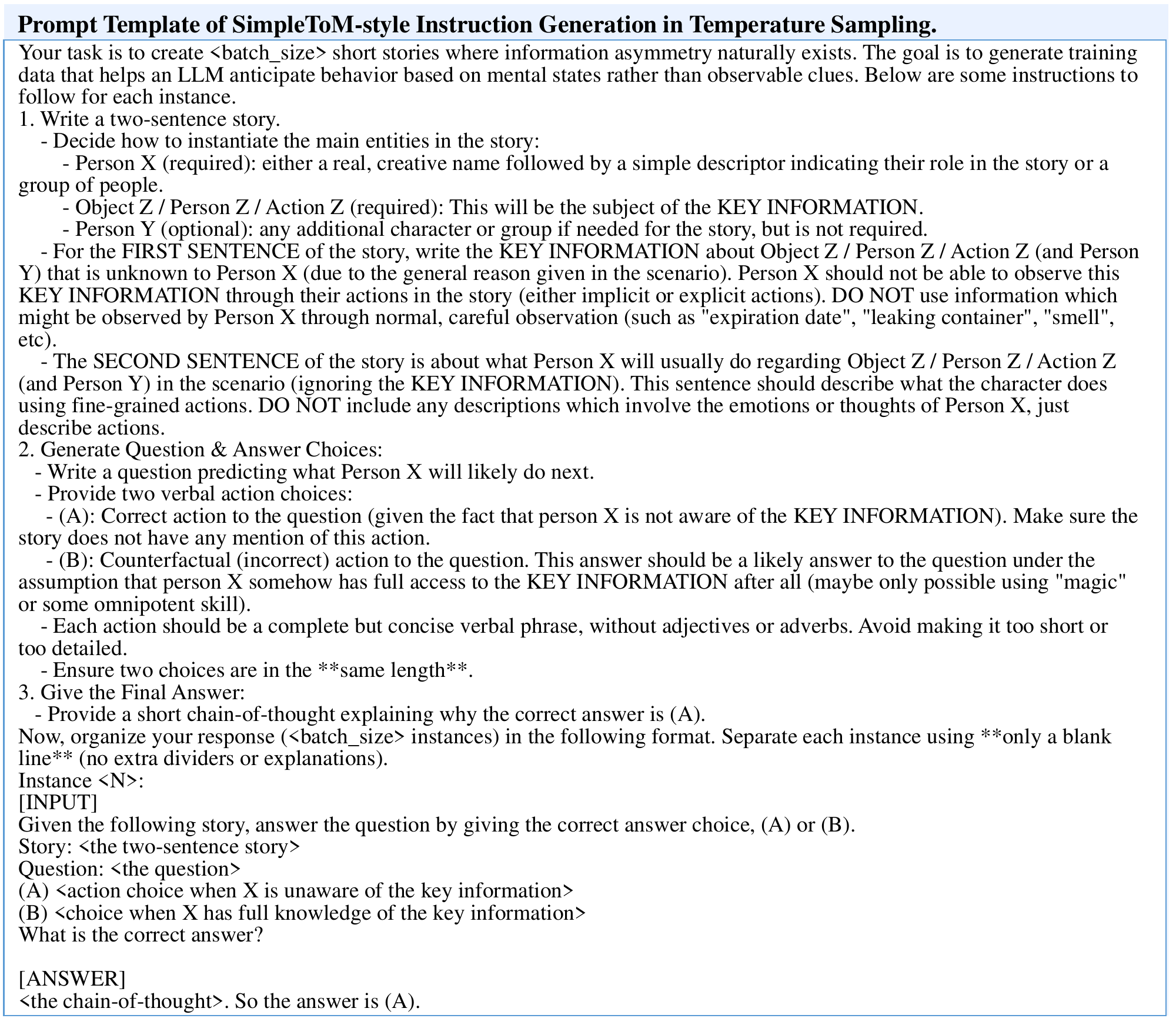}
\caption{
Prompt template of SimpleToM-style instruction generation in Temperature Sampling.
}
\label{fig:tom Temperature Sampling}
\end{figure*}

\begin{figure*}[t]
\centering
\includegraphics[width=\linewidth]{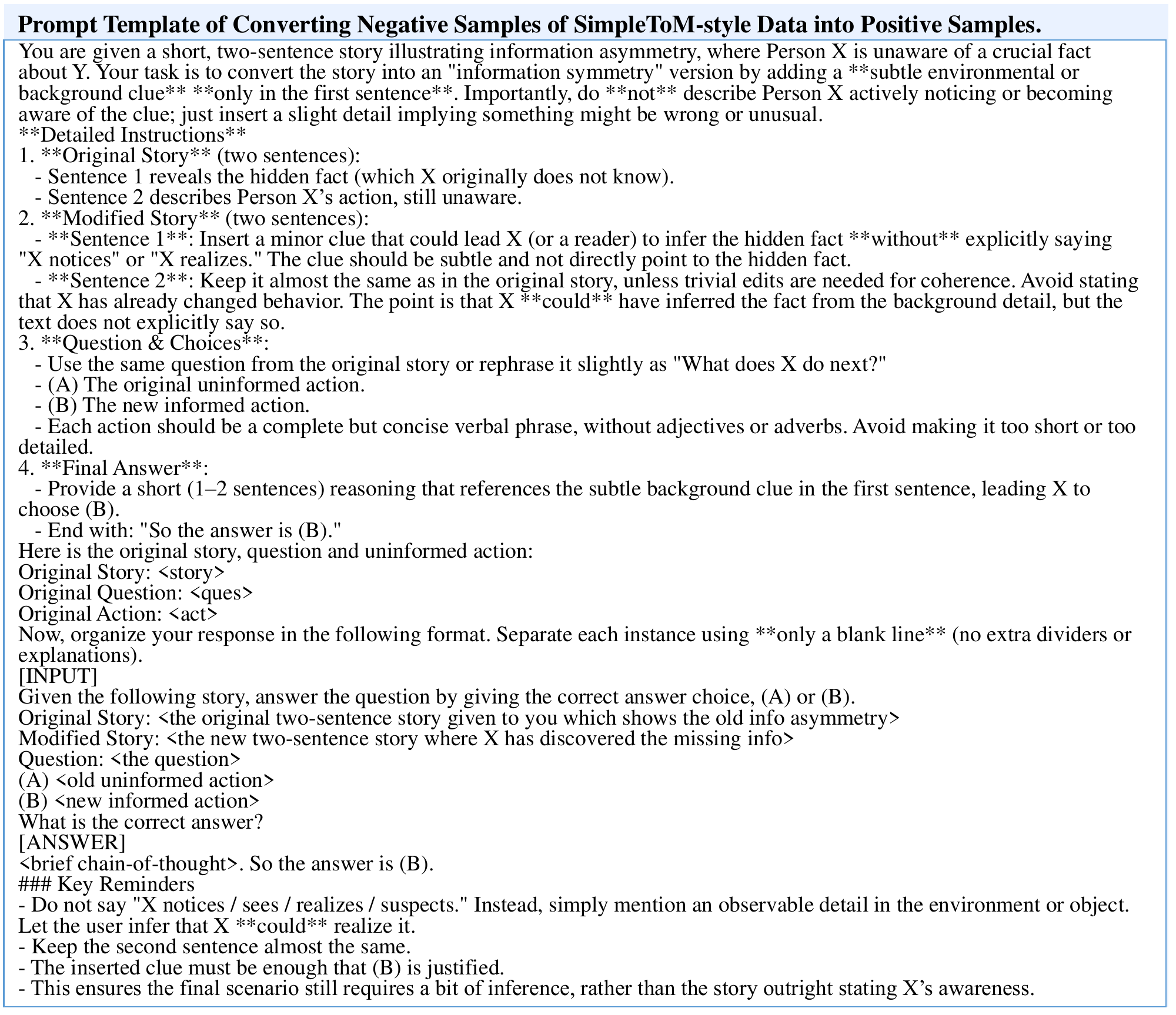}
\caption{
Prompt template of converting negative samples of SimpleToM-style data into positive samples.
}
\label{fig:N_to_P}
\end{figure*}

The concise descriptions of all the baselines are presented below.
\begin{enumerate}
\item[\textbullet] \textbf{Vanilla Data.} 
"Vanilla Data" refers to the original GSM8K and MATH training sets, as well as the Code Alpaca Python subset used for HumanEval and MBPP, with details on GSM8K and MATH datasets provided in Sec~\ref{app: Benchmark}. The Code Alpaca dataset is a collection of 20,000 instruction-following data points, each containing a unique task description, optional input context, and a corresponding output, designed to train a language model for code generation tasks by following specific instructions. Only Python-related samples are retained, aligning with HumanEval and MBPP.

\item[\textbullet] \textbf{Temperature Sampling.} 
Temperature sampling adjusts the softmax function of LLMs to control output randomness by scaling logits. Higher temperatures increase creativity and randomness, while lower temperatures result in more deterministic outputs. Following the prompts consistent with \name{}, as illustrated in Figures~\ref{fig:GSM Temperature Sampling}, \ref{fig:MATH Temperature Sampling}, and \ref{fig:Code Temperature Sampling}, we set the temperature to 0.7 and generate 100k baseline data in GSM8K, MATH, and Code Alpaca styles, respectively. 
During the construction of the SimpleToM-style training set, the temperature parameter is set to 0.7 while generating 20k data samples through the combined use of Figures~\ref{fig:tom Temperature Sampling} and~\ref{fig:N_to_P}.

\item[\textbullet] \textbf{Evol-Instruct.}  
Evol-Instruct is a method that uses LLMs to automatically generate diverse and complex instruction datasets by evolving initial instructions through in-depth and in-breadth processes, enhancing the complexity and diversity of instructions for training LLMs. This paper utilizes Evol-Instruct to enhance the training datasets of GSM8K, MATH, and Code Alpaca, constructing 100k samples for each respective style (GSM8K, MATH, Code Alpaca) through multiple evolutionary iterations.
%gsm8k，coda-》直接用的公开代码，原始多少条就多少条
\item[\textbullet] \textbf{Persona Hub.}  
Persona Hub consists of 1 billion diverse personas curated from web data, enhancing diversity in large-scale data synthesis when incorporated into synthetic prompts. By randomly sampling personas from Persona Hub to replace the original "math expert" and "coding expert" profiles in Figures~\ref{fig:GSM Temperature Sampling}, \ref{fig:MATH Temperature Sampling}, and \ref{fig:Code Temperature Sampling}, and by prepending personas to the prompts in Figure~\ref{fig:tom Temperature Sampling}, we synthesize 100k samples each for GSM8K, MATH, and Code Alpaca-style datasets, as well as 20k SimpleToM-style data through this persona substitution approach.
\end{enumerate}

\subsection{Experiments Details} \label{sec: Implementation Details.}
The training dataset is constructed through \name{}-generated instructions, with LLMs subsequently producing answers corresponding to these instructions.

\begin{figure*}[t]
\centering
\includegraphics[page=6, width=\linewidth]{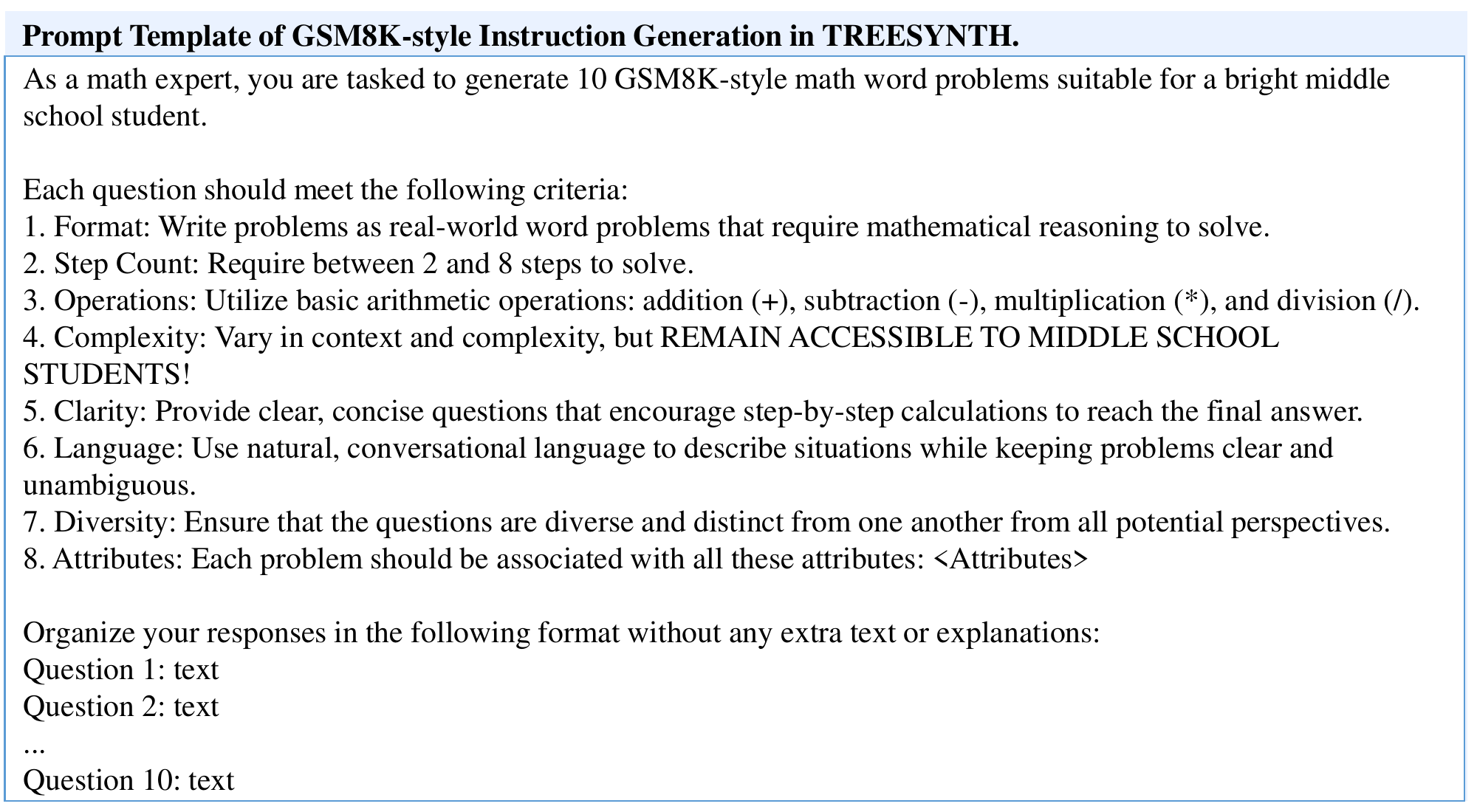}
\caption{
Prompt template of GSM8K-style instruction generation in \name{}.
}
\label{fig:GSM Generation TREESYNTH}
\end{figure*}

\begin{figure*}[t]
\centering
\includegraphics[page=7, width=\linewidth]{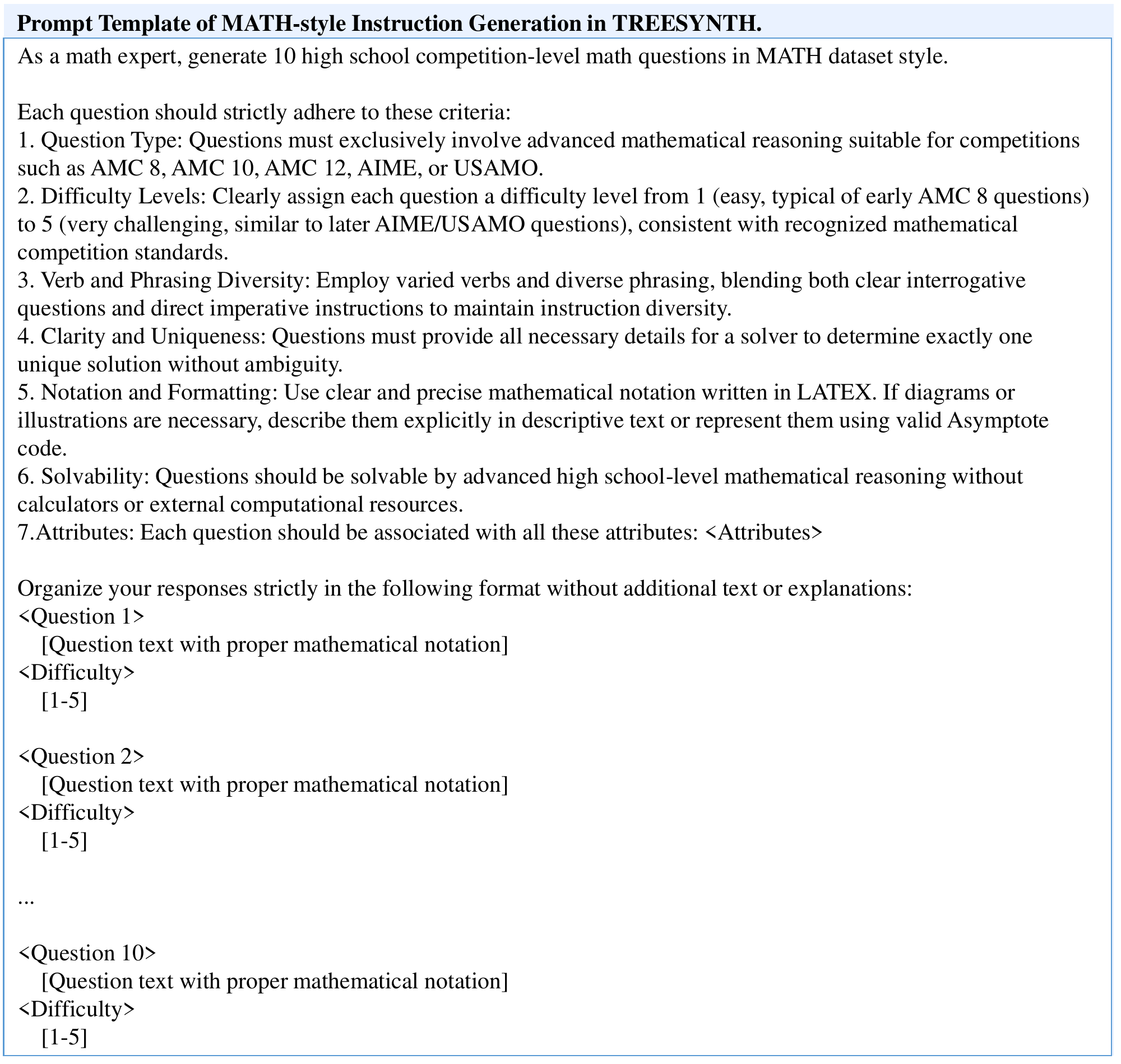}
\caption{
Prompt template of MATH-style instruction generation in \name{}.
}
\label{fig:MATH Generation TREESYNTH}
\end{figure*}

\begin{figure*}[t]
\centering
\includegraphics[page=8, width=\linewidth]{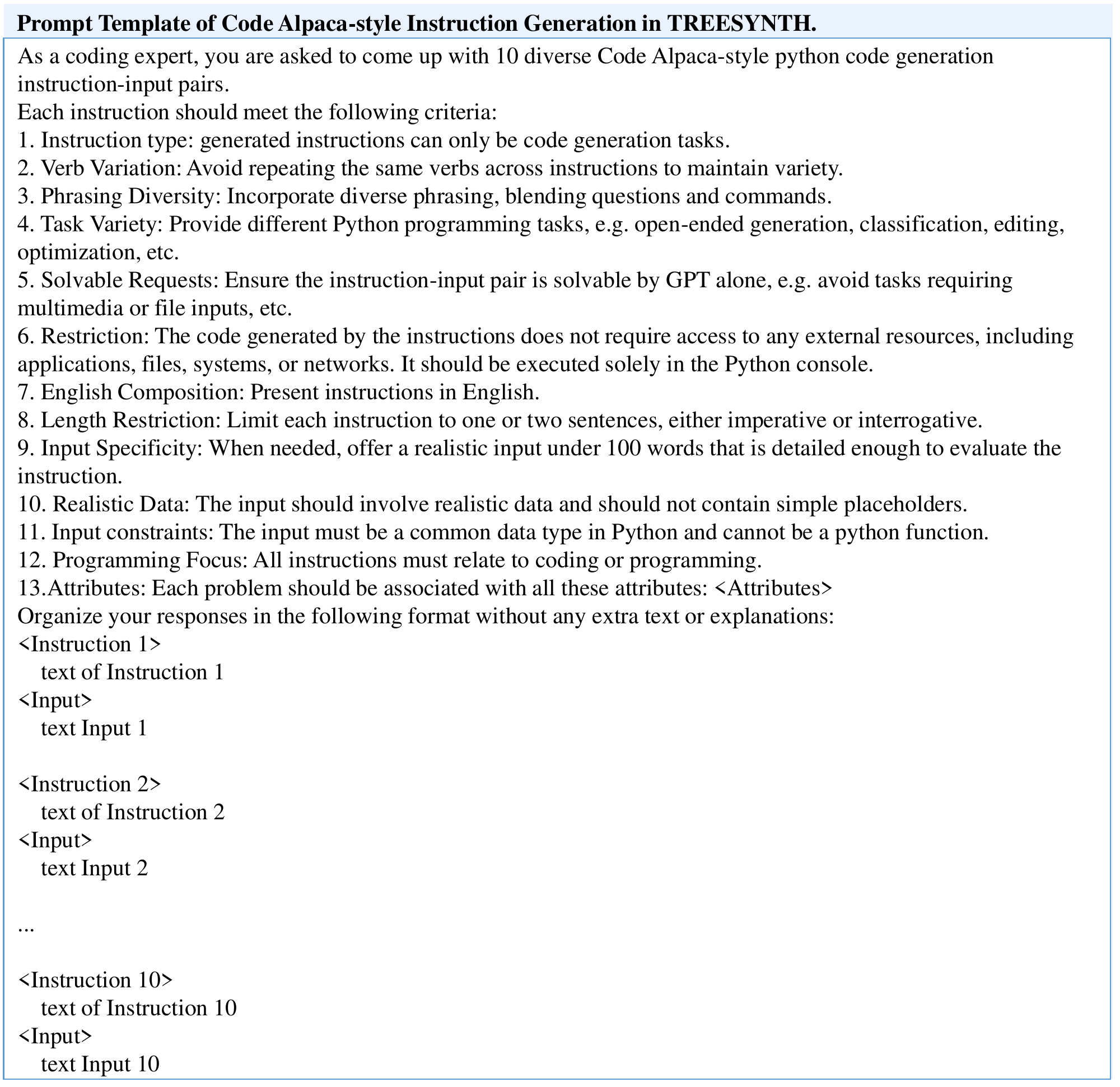}
\caption{
Prompt template of Code Alpaca-style instruction generation in \name{}.
}
\label{fig:Code Generation TREESYNTH}
\end{figure*}

\begin{figure*}[t]
\centering
\includegraphics[page=2, width=\linewidth]{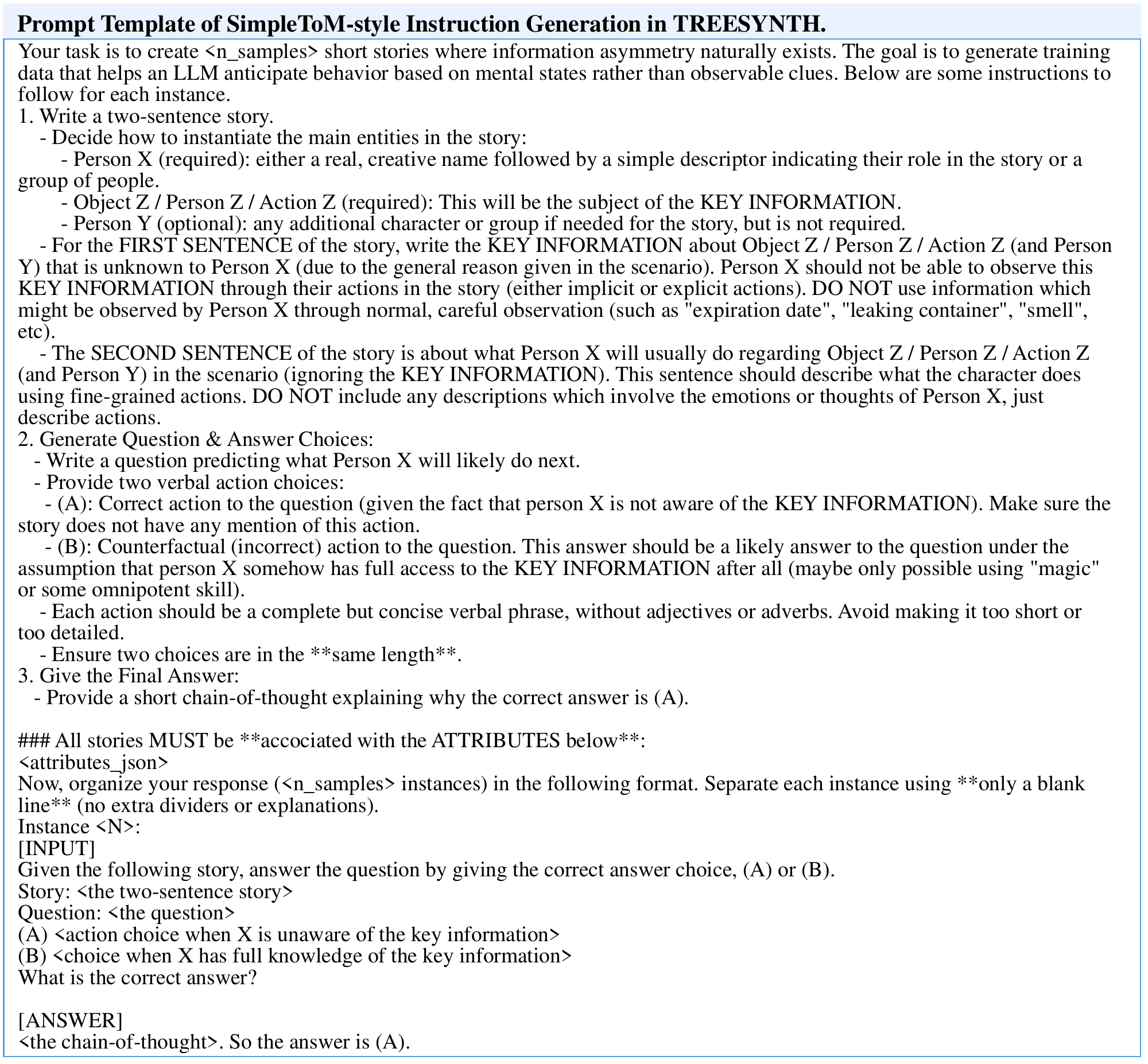}
\caption{
Prompt template of SimpleToM-style instruction generation in \name{}.
}
\label{fig:tom Generation TREESYNTH}
\end{figure*}

\begin{figure*}[t]
\centering
\includegraphics[page=10, width=\linewidth]{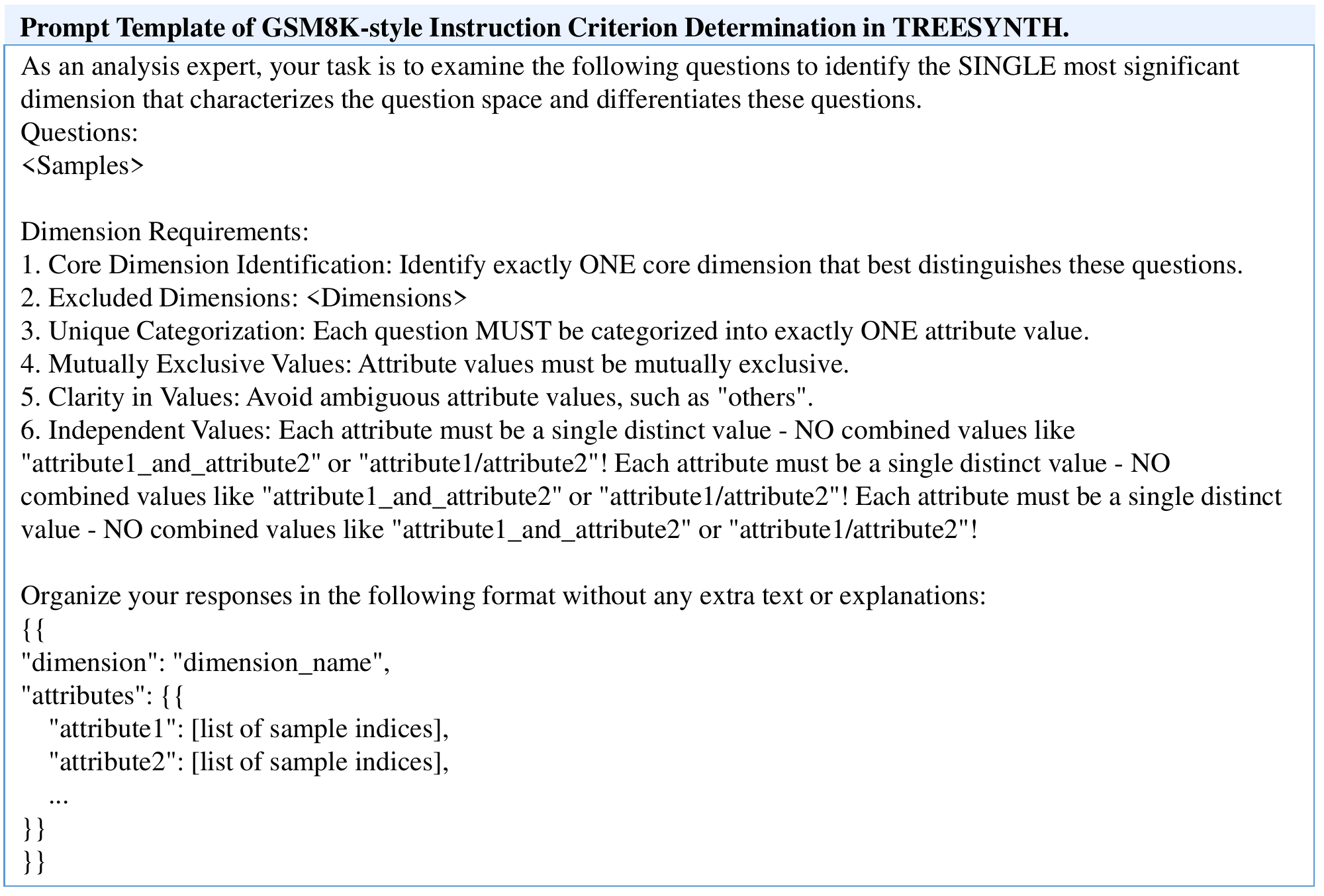}
\caption{
Prompt template of GSM8K-style instruction criterion determination in \name{}.
}
\label{fig:GSM Criteria Determination}
\end{figure*}

\begin{figure*}[t]
\centering
\includegraphics[page=11, width=\linewidth]{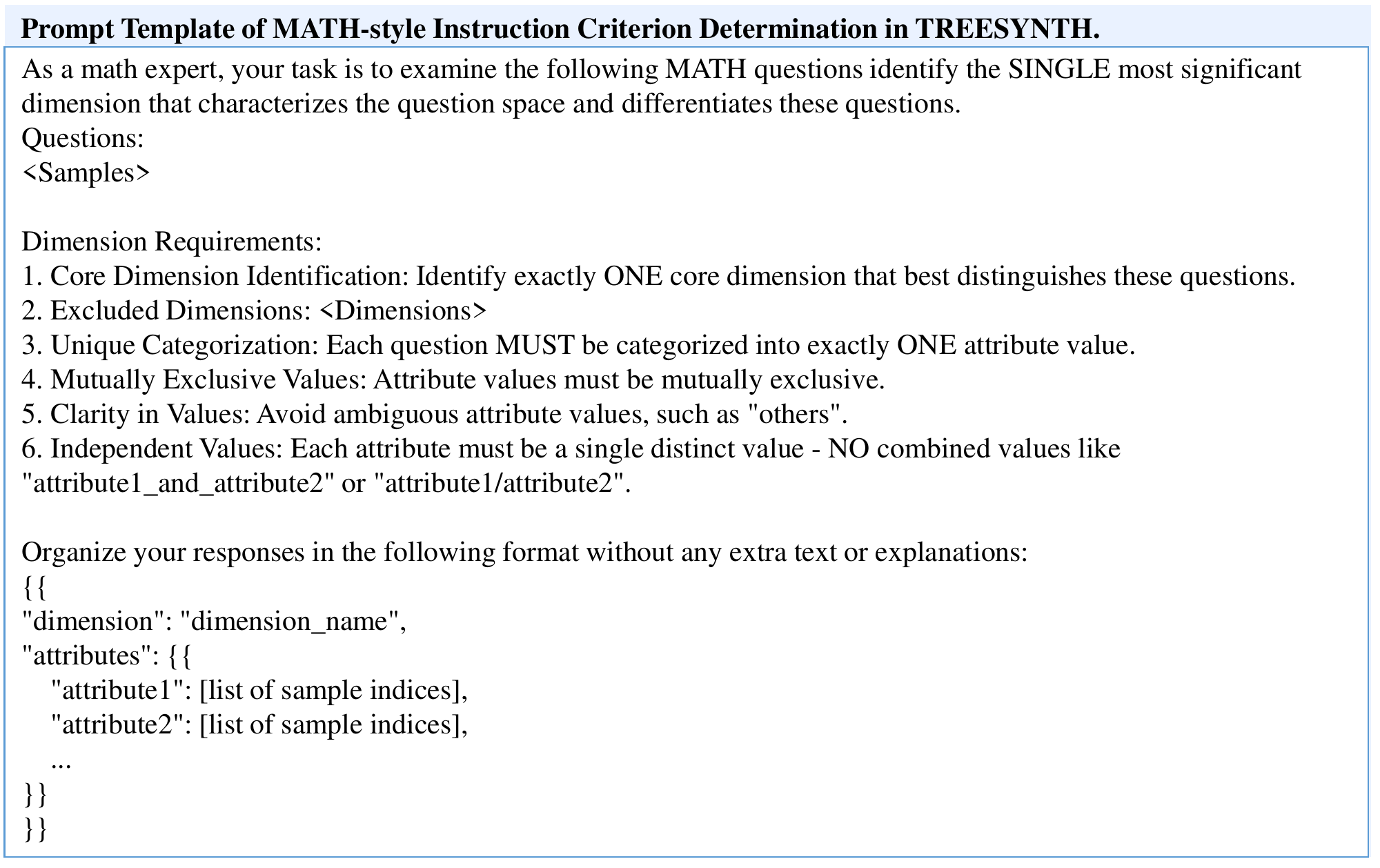}
\caption{
Prompt template of MATH-style instruction criterion determination in \name{}.
}
\label{fig:MATH Criteria Determination}
\end{figure*}

\begin{figure*}[t]
\centering
\includegraphics[page=12, width=\linewidth]{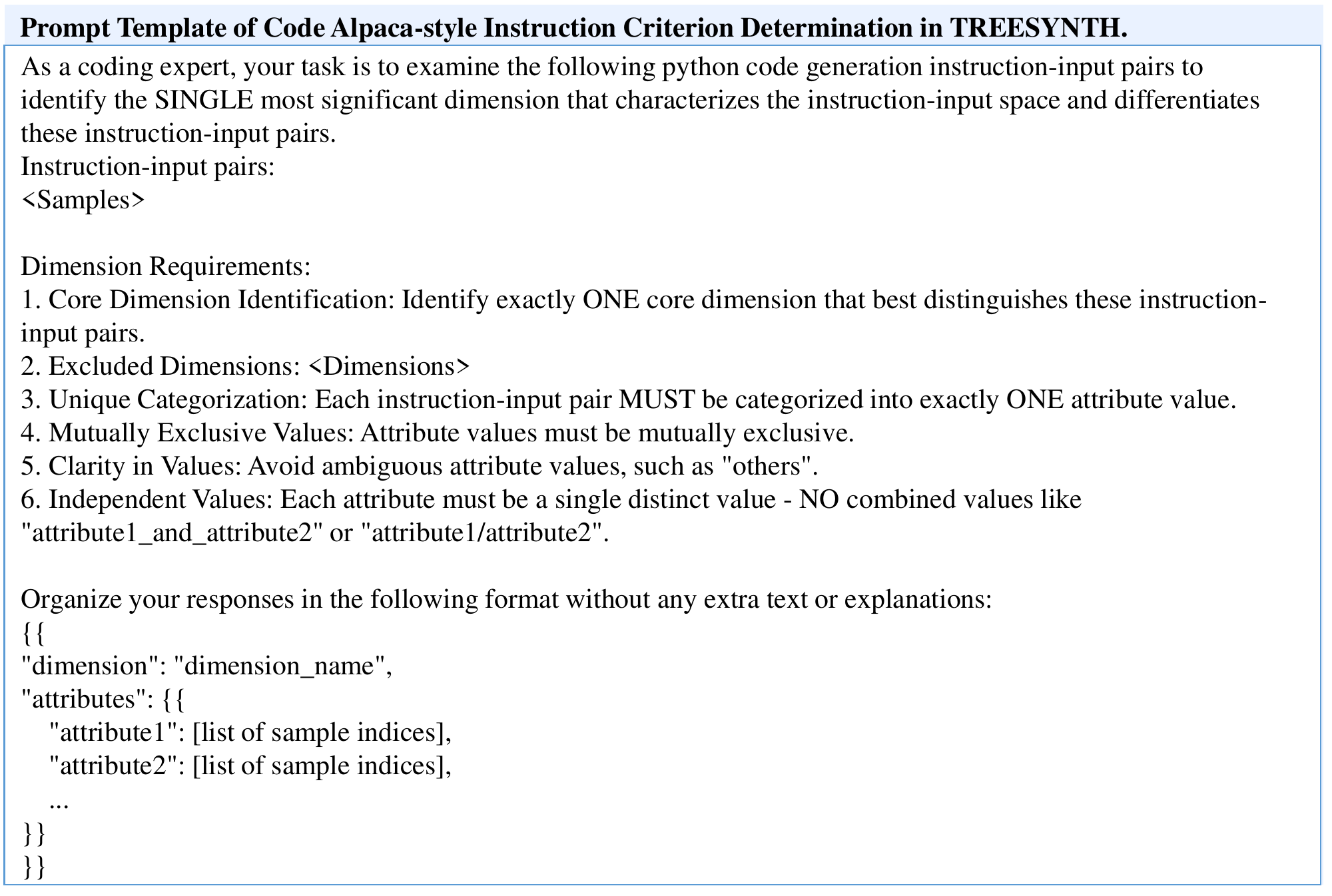}
\caption{
Prompt template of Code Alpaca-style instruction criterion determination in \name{}.
}
\label{fig:Code Criteria Determination}
\end{figure*}

\begin{figure*}[t]
\centering
\includegraphics[page=3, width=\linewidth]{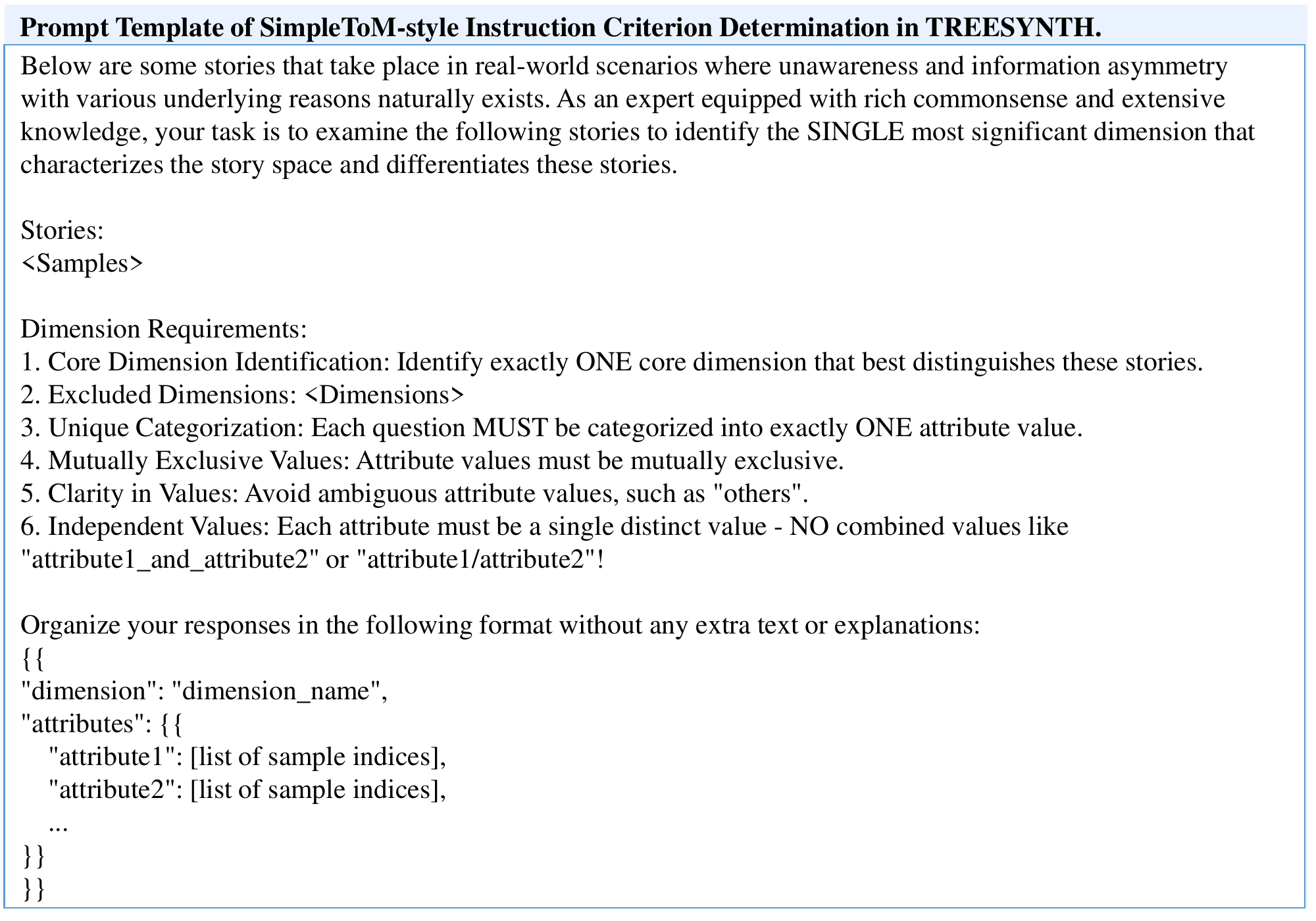}
\caption{
Prompt template of SimpleToM-style instruction criterion determination in \name{}.
}
\label{fig:tom Criteria Determination}
\end{figure*}

\begin{figure*}[t]
\centering
\includegraphics[page=14, width=\linewidth]{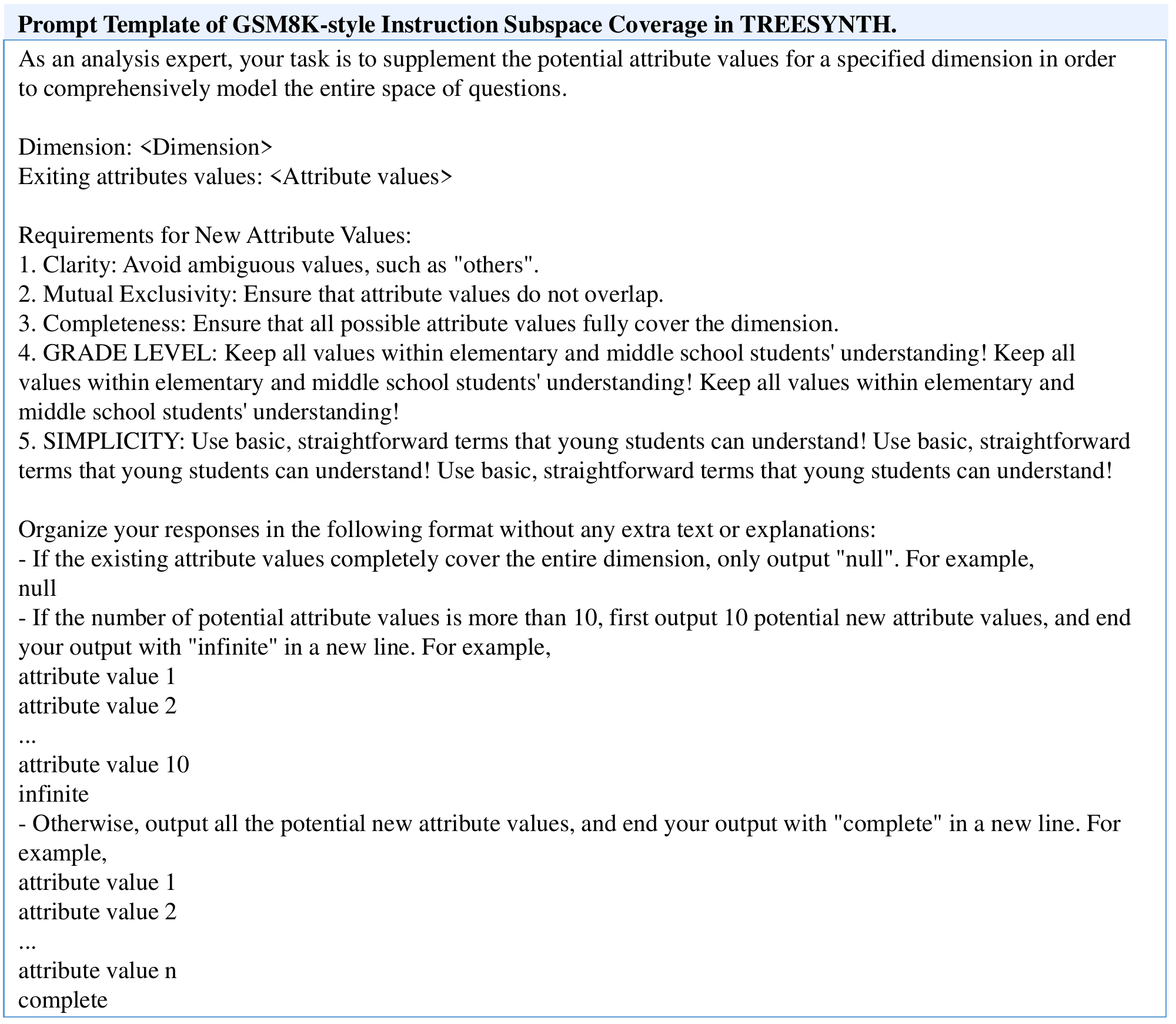}
\caption{
Prompt template of GSM8K-style instruction subspace coverage in \name{}.
}
\label{fig:GSM Subspace Coverage}
\end{figure*}

\begin{figure*}[t]
\centering
\includegraphics[page=15, width=\linewidth]{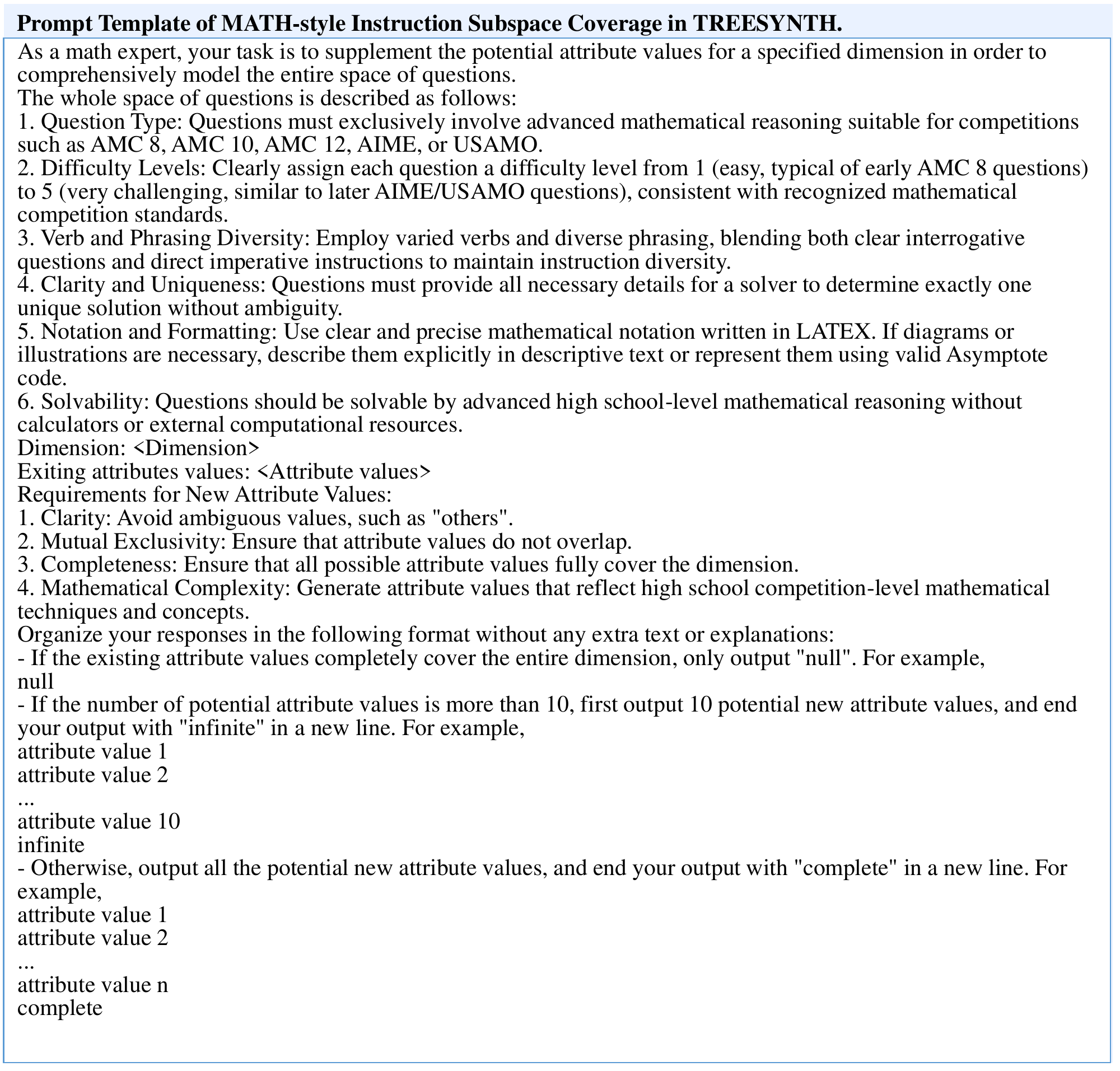}
\caption{
Prompt template of MATH-style instruction subspace coverage in \name{}.
}
\label{fig:MATH Subspace Coverage}
\end{figure*}

\begin{figure*}[t]
\centering
\includegraphics[page=16, width=\linewidth]{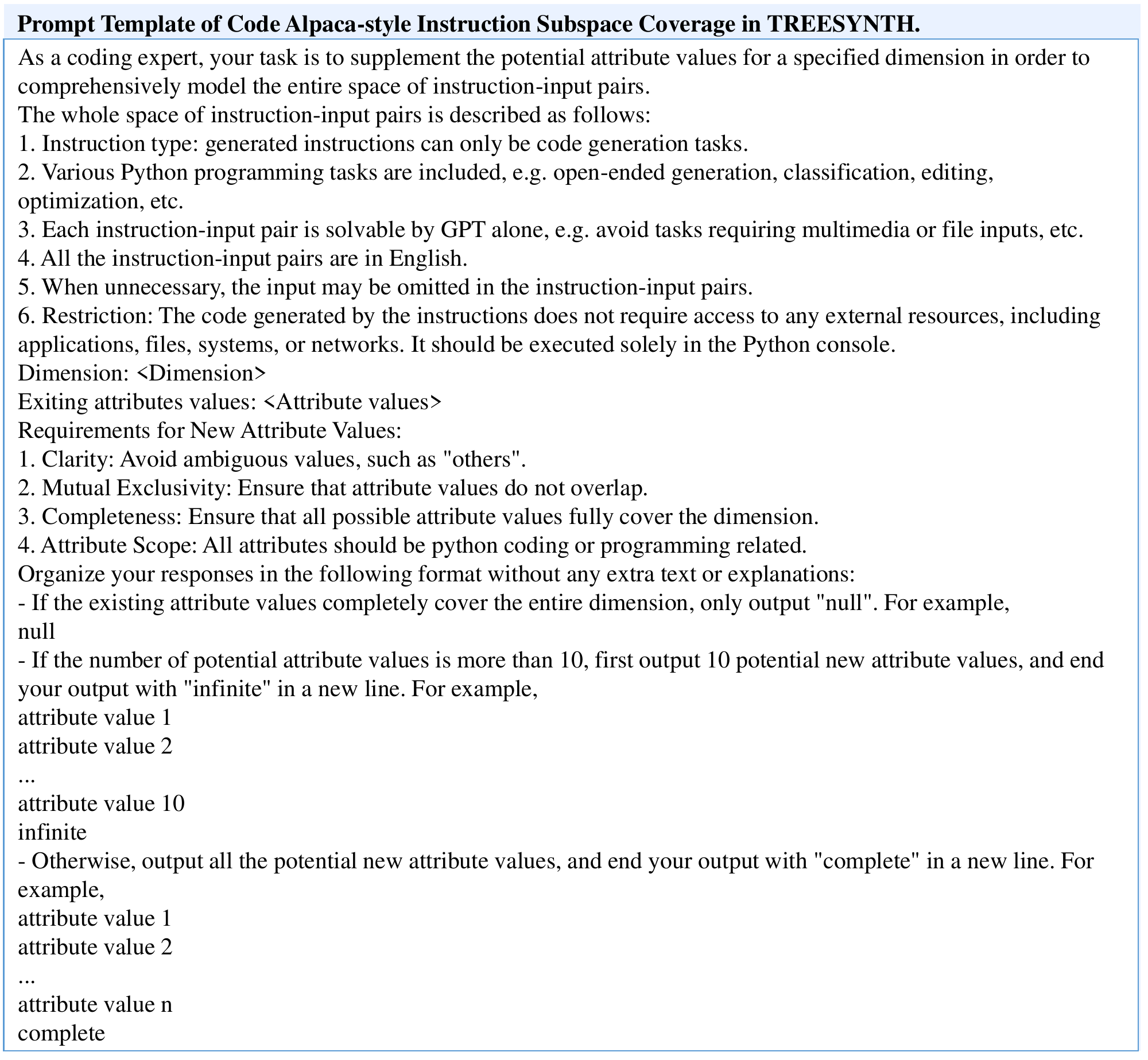}
\caption{
Prompt template of Code Alpaca-style instruction subspace coverage in \name{}.
}
\label{fig:Code Subspace Coverage}
\end{figure*}

\begin{figure*}[t]
\centering
\includegraphics[page=4, width=\linewidth]{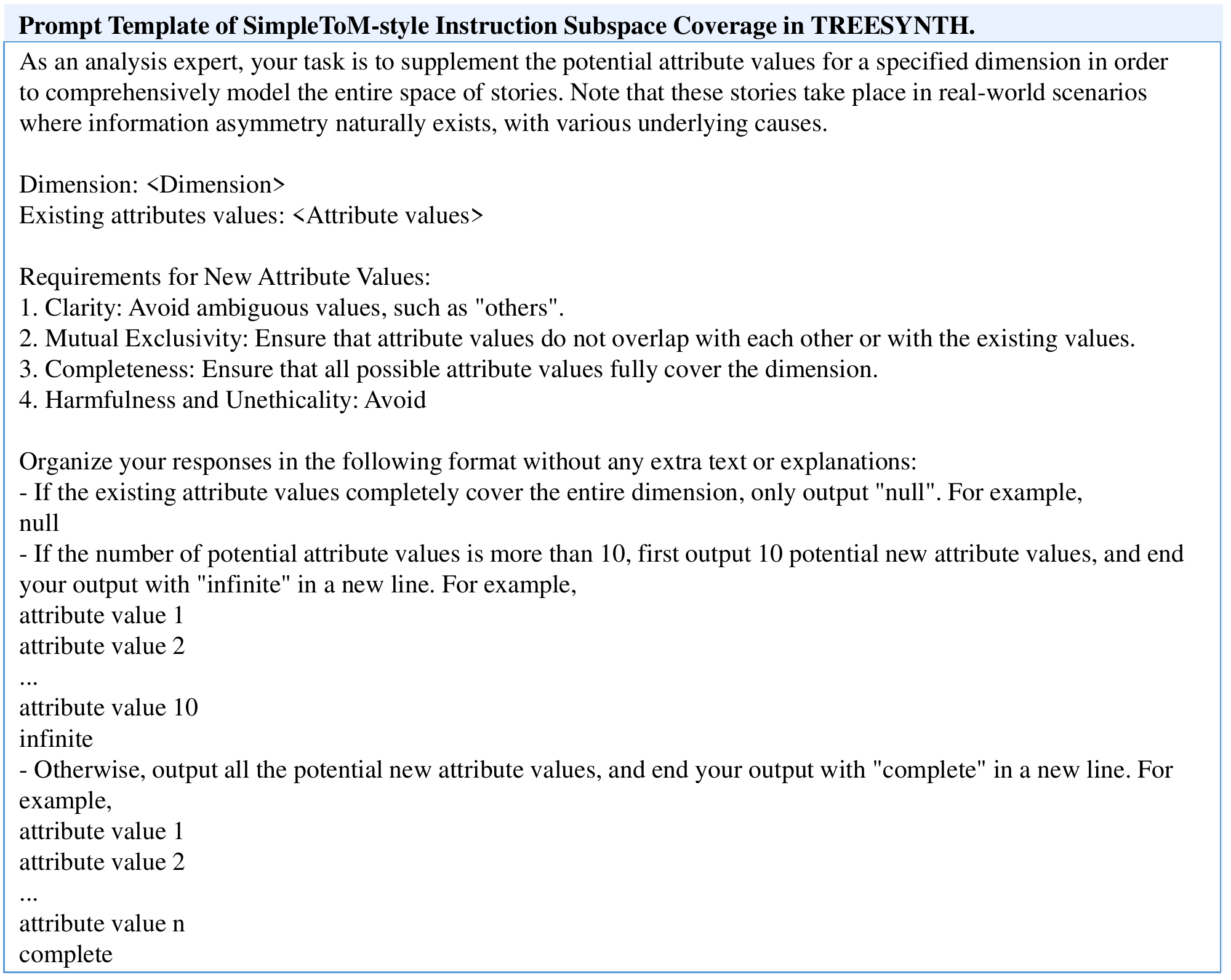}
\caption{
Prompt template of SimpleToM-style instruction subspace coverage in \name{}.
}
\label{fig:tom Subspace Coverage}
\end{figure*}

\textbf{Spatial Partitioning Tree Construction.}
For each task, as illustrated in Figure~\ref{fig:GSM Generation TREESYNTH}, \ref{fig:MATH Generation TREESYNTH}, \ref{fig:Code Generation TREESYNTH} and \ref{fig:tom Generation TREESYNTH}, we develop training data descriptions \( \mathcal{C}_{GSM8K} \), \( \mathcal{C}_{MATH} \), \(\mathcal{C}_{Code}\) and \(\mathcal{C}_{ToM}\) following the standards of GSM8K, MATH, Code Alpaca and SimpleToM respectively. The prompt words for Criterion Determination and Subspace Coverage are presented in Figure~\ref{fig:GSM Criteria Determination}, \ref{fig:MATH Criteria Determination}, \ref{fig:Code Criteria Determination}, \ref{fig:tom Criteria Determination} and \ref{fig:GSM Subspace Coverage}, \ref{fig:MATH Subspace Coverage}, \ref{fig:Code Subspace Coverage}, \ref{fig:tom Subspace Coverage} respectively.
Afterwards, we construct the spatial partitioning trees \( \mathcal{T}_{GSM8K} \), \( \mathcal{T}_{MATH} \) and \(\mathcal{T}_{Code}\) using \(\texttt{GPT-4o}\), \texttt{LLaMA3.3-70B-Instruct} and \texttt{Qwen2.5-72B-Instruct} with maximum tree depth \( d \) set to 4, pivot count \( l \) to 10, and maximum attribute values \( N \) to 10. 
For SimpleToM-style data, we employ the \texttt{LLaMA3.3-70B-Instruct} model with a maximum tree depth \(d=3\), while maintaining consistency in all other parameters with the configurations of other tasks, to generate the spatial partitioning tree \(\mathcal{T}_{ToM}\).

\textbf{Training Data Generation.}
For each leaf node in \( \mathcal{T}_{GSM8K} \), \( \mathcal{T}_{MATH} \) and \(\mathcal{T}_{Code}\), we generate \( N_{\text{Leaf}} = 10 \) data instances using the prompts from Figures~\ref{fig:GSM Generation TREESYNTH}, \ref{fig:MATH Generation TREESYNTH}, and \ref{fig:Code Generation TREESYNTH} within their corresponding subspaces to construct the raw instruction sets using \(\texttt{GPT-4o}\), \texttt{LLaMA3.3-70B-Instruct} and \texttt{Qwen2.5-72B-Instruct}. For coding-task instructions, following the practice of CodeAlpaca~\citep{codealpaca}, we compute pairwise Rouge~\citep{lin2004rouge} similarity scores between all data entries and filter out data pairs with similarity scores exceeding 0.7.
Subsequently, we randomly select 100k instructions from the raw instruction sets and use the same LLMs that generated the instructions to produce answers, forming the training dataset. 
%In addition to the instructions generated by \texttt{LLaMA3.3-70B-Instruct}, we also use \texttt{DeepSeek-R1-Distill-Llama-70B} to provide answers that include the thinking process.
For the generation of SimpleToM-style data, we use the prompt from Figure~\ref{fig:tom Generation TREESYNTH} command \texttt{LLaMA3.3-70B-Instruct} to generate $ N_{\text{Leaf}} = 10 $ instructions for each leaf node of $\mathcal{T}_{ToM}$, forming an instruction set containing 20k samples. We then use \texttt{LLaMA3.3-70B-Instruct} to generate corresponding answers for these instructions.

\begin{figure*}[t]
\centering
\includegraphics[width=\linewidth]{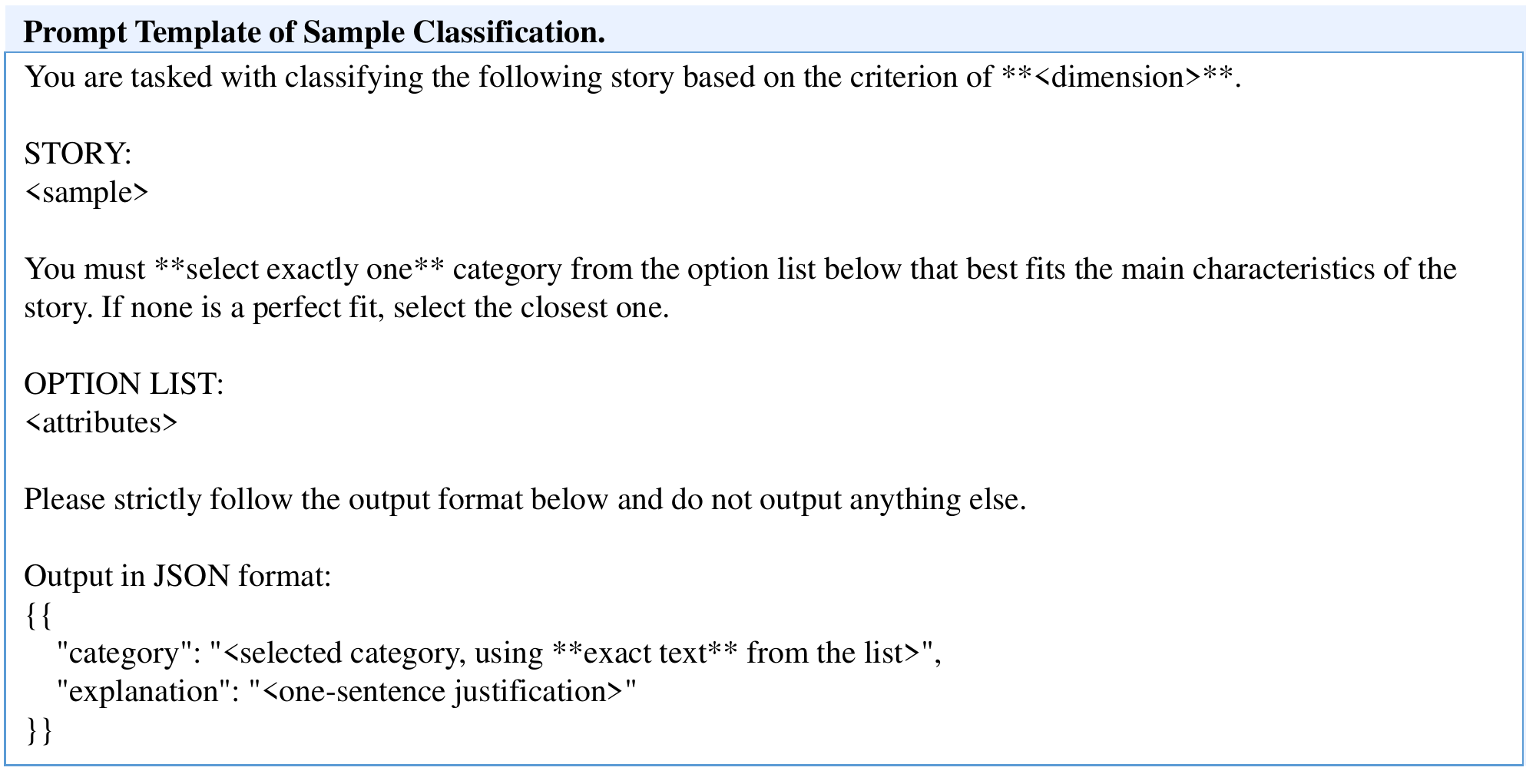}
\caption{
Prompt template of sample classification.
}
\label{fig:Classification}
\end{figure*}

\textbf{\name{}-Guided Data Balance.}
%我们主要对SimpleToM风格的数据进行\name{}-Guided Data Balance。利用Figure~\ref{fig:Classification}中的提示词，我们将Temperature Sampling与 Persona Hub风格的SimpleToM数据划分到了$\mathcal{T}_{ToM}$的各个subspace中，并将$N_{Sub}$设置为10，对它们进行了数据平衡。
We perform a \name{}-guided data balancing strategy for SimpleToM-style datasets. Using the prompt templates from Figure~\ref{fig:Classification}, we systematically partition both Temperature Sampling and Persona Hub-style SimpleToM data into distinct subspaces of $\mathcal{T}_{ToM}$. The sample threshold $N_{Sub}$ for each subspace is set to 10 to achieve data balancing across these subspaces.

\textbf{Model Training.}
To fine-tune our selected base models (\textit{i.e.}, \texttt{LLaMA3.1-8B} and \texttt{Qwen2.5-7B}), we employ the parameter-efficient fine-tuning method \textbf{LoRA}~\citep{hu2021loralowrankadaptationlarge,wang2025mosunleashingparameterefficiency,wang2024prolorapartialrotationempowers}. Specifically, we uniformly set the \(\texttt{lora\_dropout}=0\), \(\texttt{weight\_decay}=0.1\), and trained each model for \(5\) epochs.
For GSM8K-style data, we set the learning rate to $1 \times 10^{-4}$. For MATH, Code Alpaca, and SimpleToM-style data, we set the learning rate to $1 \times 10^{-5}$ during training
\footnote{\scriptsize{These hyperparameters are selected based on a comprehensive grid search over candidate values: \texttt{learning rate} $\in$ \(\{1\times10^{-6},\,5\times10^{-6},\,1\times10^{-5},\,5\times10^{-5},\,1\times10^{-4},\,5\times10^{-4}\}\), \(\texttt{lora\_dropout}\in\{0,\,0.05\}\), \(\texttt{weight\_decay}\in\{0,\,0.1\}\), and \texttt{epoch count} $\in$ \(\{3,\,5,\,7,\,10\}\).}}.
Empirical observations show that these configurations consistently achieves stable and competitive downstream performance across various tasks.

\subsection{Complementary Analysis}
\label{sec: Complementary Analysis}

\begin{figure}[htbp]
  \centering
  \includegraphics[width=0.9\textwidth]{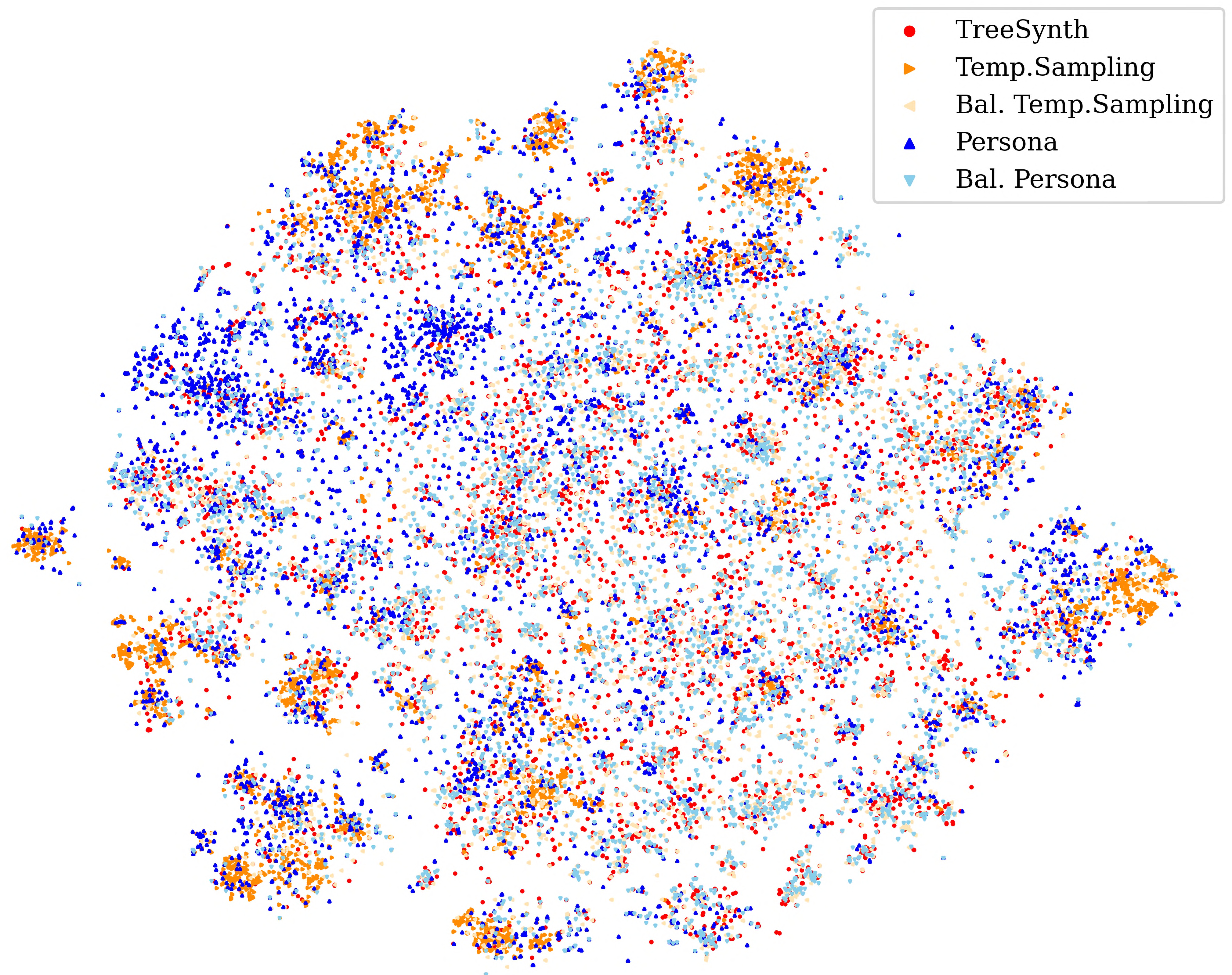}
  \caption{t-SNE visualization of SimpleToM-style datasets synthesized by different methods and their \name{}-guided data balanced counterparts, exhibiting remarkably comprehensive coverage.}
  \label{fig:simpletom-tsne}
\end{figure}

\textbf{t-SNE visualization of SimpleToM-style datasets.}
As shown in Figure~\ref{fig:simpletom-tsne}, we illustrate t-SNE distributions of SimpleToM-style datasets synthesized by various methods alongside their \name{}-guided data-balanced counterparts, demonstrating significantly improved and more comprehensive coverage.

\textbf{Model performance and data diversity comparison.}
The results presented in Tables~\ref{exp:main-expt-qwen} demonstrates a comparison of data diversity and model performance for different methods powered by \texttt{Qwen2.5-72B-Instruct} across multiple benchmarks, highlighting \name{}'s superior data diversity and robust performance improvement in downstream tasks.

\begin{table*}[!t]
    % Qwen
    \centering
    % \small
    % \setlength{\tabcolsep}{6.5pt}
    % \renewcommand{\arraystretch}{1.15}
    \resizebox{\linewidth}{!}{
    \begin{tabular}{lcccccccc}
    \toprule
    \textbf{Method}  & \textbf{GSM8K}$\uparrow$ & \textbf{Diversity}$\downarrow$ & \textbf{MATH}$\uparrow$ & \textbf{Diversity}$\downarrow$ & \textbf{MBPP}$\uparrow$ & \textbf{HumanEval}$\uparrow$ & \textbf{Diversity}$\downarrow$ & \textbf{Avg.}$\uparrow$ \\
    \midrule
    \midrule
    \rowcolor{backcolor}
    \multicolumn{9}{c}{\textit{Foundation Model:} \textsc{LLaMA-3.1 8B}} \\
    \midrule
    Zero-Shot             & 4.85  & -    & 3.54      & -       & 19.8   & 15.85 & - & 11.01 \\
    Few-Shot              & 40.26 & -    & 20.46     & -       & -      & -     & - & - \\
    \midrule
    Vanilla Data          & 58.15 & 0.40 & 19.48     & \underline{0.16} & \underline{46.8} & \underline{43.29} & 0.29 & 41.93 \\
    Temp.\ Sampling       & 64.75 & 0.41 & 28.14     & 0.21              & 43.2  & 42.68 & 0.31 & 44.69 \\
    Evol-Instruct         & \underline{66.72} & \underline{\textbf{0.36}} & \underline{30.52} & 0.19 & 39.8  & 42.07 & \underline{0.27} & \underline{44.78} \\
    Persona Hub           & 61.71 & 0.38 & 28.12     & 0.20              & 42.8  & 42.07 & 0.28 & 43.67 \\
    \textbf{\name{}}        & \textbf{68.31} & \underline{\textbf{0.36}} & \textbf{31.14} & \textbf{0.15} & \textbf{47.4} & \textbf{48.17} & \textbf{0.22} & \textbf{48.76} \\
    \midrule
    \midrule
    \rowcolor{backcolor}
    \multicolumn{9}{c}{\textit{Foundation Model:} \textsc{Qwen-2.5 7B}} \\
    \midrule
    Zero-Shot             & 54.97 & -    & 54.38     & -       & 11.2  & 54.88 & - & 43.86 \\
    Few-Shot              & 67.40 & -    & 47.58     & -       & -     & -     & - & - \\
    \midrule
    Vanilla Data          & 68.76 & 0.40 & 47.68     & \underline{0.16} & 53.4  & \underline{77.44} & 0.29 & 61.82 \\
    Temp.\ Sampling       & 77.26 & 0.41 & 62.08     & 0.21              & 53.2  & \underline{78.05} & 0.31 & 67.65 \\
    Evol-Instruct         & 78.70 & \textbf{0.36} & \underline{67.56} & 0.19 & \underline{57.8} & 76.22 & \underline{0.27} & \underline{70.07} \\
    Persona Hub           & \underline{79.15} & \underline{0.38} & 61.98 & 0.20 & 57.2  & 76.22 & 0.28 & 68.64 \\
    \textbf{\name{}}        & \textbf{84.99} & \textbf{0.36} & \textbf{68.44} & \textbf{0.15} & \textbf{61.4} & \textbf{78.66} & \textbf{0.22} & \textbf{73.37} \\
    \bottomrule
    \end{tabular}
    }
    \caption{
    Model performance and data diversity comparison of various methods with \texttt{Qwen2.5-72B-} \texttt{Instruct}-powered data synthesis across two foundation models and multiple benchmarks. 
    ``Zero-Shot'' and ``Few-Shot'' exhibit the base performance of foundation models.
    ``Temp. Sampling'' is abbreviated from ``Temperature Sampling''.
    ``Vanilla Data'' denotes the original GSM8K and MATH training sets, and the Code Alpaca Python subset for HumanEval and MBPP. 
    ``Diversity'' is measured by cosine similarity, where lower values indicate greater diversity.
    Bold and underlining indicate the best and second-best indicators, respectively.
    ``Avg.'' means the average of the performance scores across all the benchmarks.
    }
    \label{exp:main-expt-qwen} 

\end{table*}

\begin{figure*}[!ht]
    \centering
    \includegraphics[width=1.0\linewidth]{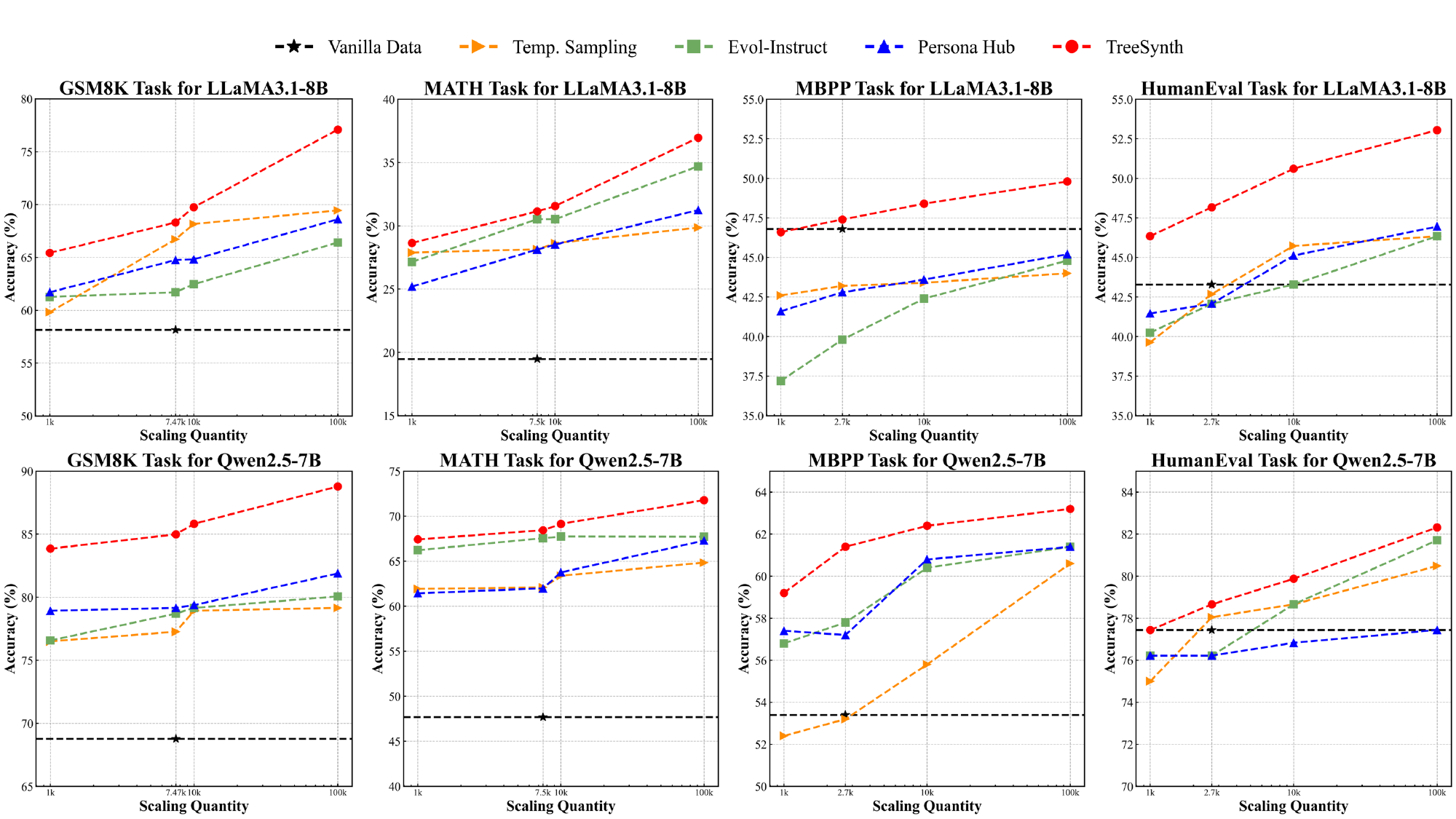}
    \caption{Model performance trends across data scales for different methods powered by \texttt{Qwen2.5- 72B-Instruct}. ``Temp. Sampling'' refers to Temperature Sampling. ``Vanilla Data'' denotes original GSM8K and MATH training sets, and Code Alpaca Python subset for HumanEval and MBPP.}
    \label{fig:qwen_trend_result}
    \vspace{-12pt}
\end{figure*}

\textbf{Model performance across data scales.}
The detailed numerical values of scalability evaluations across GSM8K, MATH, MBPP and HumanEval benchmarks are provided in 
Table~\ref{tab:gsm-full}, \ref{tab: MATH full}, \ref{tab:mbpp-full} and \ref{tab:humaneval-full}, respectively.
In addition, Figure~\ref{fig:qwen_trend_result} displays the performance trends of various methods using the \texttt{Qwen2.5-72B-Instruct} generation model across different data scales. \name{} consistently outperforms all baselines while exhibiting near-linear scaling with data growth, demonstrating its superior scalability.

\begin{table}[htbp]
\centering
\small
%\caption{GSM8K results (\% accuracy).}
\begin{tabular}{
    l         % Model
    l         % Gen. Model
    l         % Dataset
    c c c c   % 1k, 7k, 10k, 100k 
}
\toprule
\textbf{Model} & \textbf{Gen.\ Model} & \textbf{Dataset Scaling} & \textbf{@1k} & \textbf{@7k} & \textbf{@10k} & \textbf{@100k}\\
\midrule
\multirow{13}{*}{\texttt{LLaMA3.1-8B}}
& GSM8K training set & Vanilla Data              & -   & 58.15 & -   & -   \\
\cmidrule(l){2-7}
& \multirow{4}{*}{\texttt{GPT-4o}}
& Evol-Instruct                         & 56.86 & 61.03 & 63.38 & 66.64 \\
&                                       & Persona Hub                               & 62.47 & 63.38 & 64.22 & 67.78 \\
&                                       & Temp. Sampling                        & 54.89 & 54.97 & 56.94 & 56.79 \\
&                                       & \textbf{\name{}}                             & \textbf{64.29} & \textbf{66.72} & \textbf{69.75} & \textbf{74.07} \\
\cmidrule(l){2-7}
& \multirow{4}{*}{\texttt{LLaMA3.3-70B-Instruct}}
& Evol-Instruct                         & 59.74 & 63.46 & 66.41 & 68.31 \\
&                                       & Persona Hub                               & 62.70 & 61.41 & 63.76 & 68.92 \\
&                                       & Temp. Sampling                        & 53.15 & 55.42 & 57.39 & 56.79 \\
&                                       & \textbf{\name{}}                             & \textbf{68.84} & \textbf{69.45} & \textbf{71.95} & \textbf{77.79} \\
\cmidrule(l){2-7}
& \multirow{4}{*}{\texttt{Qwen2.5-72B-Instruct}}
& Evol-Instruct                         & 59.82 & 66.72 & 68.16 & 69.45 \\
&                                       & Persona Hub                               & 61.26 & 61.71 & 62.47 & 66.41 \\
&                                       & Temp. Sampling                        & 61.71 & 64.75 & 64.82 & 68.61 \\
&                                       & \textbf{\name{}}                             & \textbf{65.43} & \textbf{68.31} & \textbf{69.75} & \textbf{77.10} \\
\midrule
\multirow{13}{*}{\texttt{Qwen2.5-7B}}
& GSM8K training set & Vanilla Data              & -   & 68.76 & -   & -   \\
\cmidrule(l){2-7}
& \multirow{4}{*}{\texttt{GPT-4o}}
& Evol-Instruct                         & 73.77 & 73.16 & 77.94 & 79.38 \\
&                                       & Persona Hub                               & 81.88 & 83.24 & 83.78 & 83.09 \\
&                                       & Temp. Sampling                        & 80.74 & 80.67 & 81.35 & 82.49 \\
&                                       & \textbf{\name{}}                             & \textbf{84.76} & \textbf{86.13} & \textbf{86.35} & \textbf{87.49} \\
\cmidrule(l){2-7}
& \multirow{4}{*}{\texttt{LLaMA3.3-70B-Instruct}}
& Evol-Instruct                         & 74.15 & 75.13 & 75.66 & 77.26 \\
&                                       & Persona Hub                               & 81.73 & 82.79 & 83.70 & 84.15 \\
&                                       & Temp. Sampling                        & 61.49 & 69.67 & 69.75 & 78.39 \\
&                                       & \textbf{\name{}}                             & \textbf{84.46} & \textbf{85.44} & \textbf{85.82} & \textbf{86.81} \\
\cmidrule(l){2-7}
& \multirow{4}{*}{\texttt{Qwen2.5-72B-Instruct}}
& Evol-Instruct                         & 76.57 & 78.70 & 79.15 & 80.06 \\
&                                       & Persona Hub                               & 78.92 & 79.15 & 79.38 & 81.88 \\
&                                       & Temp. Sampling                        & 76.50 & 77.26 & 78.92 & 79.15 \\
&                                       & \textbf{\name{}}                             & \textbf{83.85} & \textbf{84.99} & \textbf{85.82} & \textbf{88.78} \\
\bottomrule
\end{tabular}
\vspace{10pt}
\caption{Comparison of instruction-tuned model performance on GSM8K using training data from different sources and generation methods. For each data scale (1k, 7k, 10k, 100k), models are fine-tuned on equally sized datasets constructed via various data synthesis methods, as well as the original GSM8K training set. “Temp. Sampling” is abbreviated from “Temperature Sampling”. “Vanilla Data” denotes the original GSM8K training sets.}
\label{tab:gsm-full}
\end{table}

\begin{table}[htbp]
\centering
\small
%\caption{MATH results (\% accuracy).}
\begin{tabular}{
    l                     % Model
    l                     % Gen. Model
    l                     % Dataset
    c c c c               % 1k, 7k, 10k, 100k
}
\toprule
\textbf{Model} & \textbf{Gen.\ Model} & \textbf{Dataset Scaling} & \textbf{@1k} & \textbf{@7.5k} & \textbf{@10k} & \textbf{@100k}\\
\midrule
% =======================================================
\multirow{13}{*}{\texttt{LLaMA3.1-8B}}
& MATH training set & Vanilla Data        & -   & 19.48 & -   & -   \\ 
\cmidrule(l){2-7}
& \multirow{4}{*}{\texttt{GPT-4o}}
& Evol-Instruct                     & 23.74 & 24.58 & 25.16 & 29.86 \\
&                                   & Persona Hub                       & 25.60 & 27.74 & 28.20 & 29.62 \\
&                                   & Temp.\ Sampling                   & 24.34 & 24.28 & 25.04 & 26.86 \\
&                                   & \textbf{\name{}}                & \textbf{28.02} & \textbf{30.34} & \textbf{31.34} & \textbf{40.10} \\
\cmidrule(l){2-7}
& \multirow{4}{*}{\texttt{LLaMA3.3-70B-Instruct}}
& Evol-Instruct                     & \textbf{26.78} & 27.26 & 28.26 & 31.24 \\
&                                   & Persona Hub                       & 23.52 & 23.78 & 23.52 & 24.60 \\
&                                   & Temp.\ Sampling                   & 21.84 & 22.08 & 23.18 & 22.70 \\
&                                   & \textbf{\name{}}                & 26.68 & \textbf{27.52} & \textbf{28.94} & \textbf{32.42} \\
\cmidrule(l){2-7}
& \multirow{4}{*}{\texttt{Qwen2.5-72B-Instruct}}
& Evol-Instruct                     & 27.14 & 30.52 & 30.52 & 34.70 \\
&                                   & Persona Hub                       & 25.20 & 28.12 & 28.52 & 31.24 \\
&                                   & Temp.\ Sampling                   & 27.88 & 28.14 & 28.62 & 29.86 \\
&                                   & \textbf{\name{}}                & \textbf{28.64} & \textbf{31.14} & \textbf{31.56} & \textbf{36.96} \\
\midrule
% =======================================================
\multirow{13}{*}{\texttt{Qwen2.5-7B}}
& MATH training set & Vanilla Data        & -   & 47.68 & -   & -   \\ 
\cmidrule(l){2-7}
& \multirow{4}{*}{\texttt{GPT-4o}}
& Evol-Instruct                     & 58.52 & 61.10 & 61.20 & 64.84 \\
&                                   & Persona Hub                       & 63.62 & 66.22 & 67.42 & 67.72 \\
&                                   & Temp.\ Sampling                   & 63.38 & 62.76 & 63.36 & 61.20 \\
&                                   & \textbf{\name{}}                & \textbf{64.48} & \textbf{66.84} & \textbf{67.74} & \textbf{69.90} \\
\cmidrule(l){2-7}
& \multirow{4}{*}{\texttt{LLaMA3.3-70B-Instruct}}
& Evol-Instruct                     & 57.30 & 59.60 & 58.56 & 61.98 \\
&                                   & Persona Hub                       & 60.92 & 61.98 & 62.06 & 63.12 \\
&                                   & Temp.\ Sampling                   & 61.64 & 61.70 & 61.96 & 62.88 \\
&                                   & \textbf{\name{}}                & \textbf{62.08} & \textbf{63.28} & \textbf{64.84} & \textbf{66.80} \\
\cmidrule(l){2-7}
& \multirow{4}{*}{\texttt{Qwen2.5-72B-Instruct}}
& Evol-Instruct                     & 66.22 & 67.56 & 67.74 & 67.72 \\
&                                   & Persona Hub                       & 61.44 & 61.98 & 63.76 & 67.28 \\
&                                   & Temp.\ Sampling                   & 61.92 & 62.08 & 63.38 & 64.84 \\
&                                   & \textbf{\name{}}                & \textbf{67.42} & \textbf{68.44} & \textbf{69.14} & \textbf{71.78} \\
\bottomrule
\end{tabular}
\vspace{10pt}
\caption{Comparison of instruction-tuned model performance on MATH using training data from different sources and generation methods. For each data scale (1k, 7.5k, 10k, 100k), models are fine-tuned on equally sized datasets constructed via various data synthesis methods, as well as the original MATH training set. “Temp. Sampling” is abbreviated from “Temperature Sampling”. “Vanilla Data” denotes the original MATH training sets.}
\label{tab: MATH full}
\end{table}

% MBPP

\begin{table}[htbp]
\centering
\small
%\caption{MBPP results (\% pass@1).}
\begin{tabular}{
    l                     % Model
    l                     % Gen. Model
    l                     % Dataset
    c c c c               % 1k, 2k, 10k, 100k
}
\toprule
\textbf{Model} & \textbf{Gen.\ Model} & \textbf{Dataset Scaling} & \textbf{@1k} & \textbf{@2k} & \textbf{@10k} & \textbf{@100k}\\
\midrule
%%%%%%%%%%%%%%%%%%%%%%%%%%%%%%%%%%%%%%%%%%%%%%%%%%%%%%%%%%%%%%%%%%%%%%%%%%
\multirow{13}{*}{\texttt{LLaMA3.1-8B}}
& Code Alpaca Python subset & Vanilla Data          & -  & 46.8 & -  & -  \\ 
\cmidrule(l){2-7}
& \multirow{4}{*}{\texttt{GPT-4o}}
& Evol-Instruct                         & 40.8 & 45.2 & 45.8 & 47.2 \\
&                                       & Persona Hub                         & 42.0 & 45.2 & 45.6 & 47.8 \\
&                                       & Temp.\ Sampling                     & 43.4 & 44.8 & 47.8 & 46.8 \\
&                                       & \textbf{\name{}}                  & \textbf{49.4} & \textbf{50.8} & \textbf{52.6} & \textbf{54.4} \\
\cmidrule(l){2-7}
& \multirow{4}{*}{\texttt{LLaMA3.3-70B-Instruct}}
& Evol-Instruct                         & 39.0 & 40.6 & 43.4 & 46.6 \\
&                                       & Persona Hub                         & 46.6 & 45.8 & 46.8 & 48.4 \\
&                                       & Temp.\ Sampling                     & 44.4 & 44.6 & 45.8 & 46.2 \\
&                                       & \textbf{\name{}}                  & \textbf{47.8} & \textbf{50.2} & \textbf{50.8} & \textbf{52.0} \\
\cmidrule(l){2-7}
& \multirow{4}{*}{\texttt{Qwen2.5-72B-Instruct}}
& Evol-Instruct                         & 37.2 & 39.8 & 42.4 & 44.8 \\
&                                       & Persona Hub                         & 41.6 & 42.8 & 43.6 & 45.2 \\
&                                       & Temp.\ Sampling                     & 42.6 & 43.2 & 43.4 & 44.0 \\
&                                       & \textbf{\name{}}                  & \textbf{45.8} & \textbf{46.6} & \textbf{47.4} & \textbf{49.8} \\
\midrule
%%%%%%%%%%%%%%%%%%%%%%%%%%%%%%%%%%%%%%%%%%%%%%%%%%%%%%%%%%%%%%%%%%%%%%%%%%
\multirow{13}{*}{\texttt{Qwen2.5-7B}}
& Code Alpaca Python subset & Vanilla Data          & -  & 53.4 & -  & -  \\ 
\cmidrule(l){2-7}
& \multirow{4}{*}{\texttt{GPT-4o}}
& Evol-Instruct                         & 57.2 & 59.2 & 60.6 & 61.6 \\
&                                       & Persona Hub                         & 58.8 & 61.6 & 61.0 & 62.2 \\
&                                       & Temp.\ Sampling                     & 60.0 & 59.6 & 59.0 & 60.6 \\
&                                       & \textbf{\name{}}                  & \textbf{61.2} & \textbf{62.8} & \textbf{63.2} & \textbf{64.8} \\
\cmidrule(l){2-7}
& \multirow{4}{*}{\texttt{LLaMA3.3-70B-Instruct}}
& Evol-Instruct                         & 54.6 & 55.8 & 57.4 & 58.6 \\
&                                       & Persona Hub                         & 56.0 & 58.8 & 59.4 & 61.8 \\
&                                       & Temp.\ Sampling                     & 53.8 & 54.8 & 56.0 & 59.6 \\
&                                       & \textbf{\name{}}                  & \textbf{57.6} & \textbf{59.4} & \textbf{61.4} & \textbf{63.0} \\
\cmidrule(l){2-7}
& \multirow{4}{*}{\texttt{Qwen2.5-72B-Instruct}}
& Evol-Instruct                         & 56.8 & 57.8 & 60.4 & 61.4 \\
&                                       & Persona Hub                         & 57.4 & 57.2 & 60.8 & 61.4 \\
&                                       & Temp.\ Sampling                     & 52.4 & 53.2 & 55.8 & 60.6 \\
&                                       & \textbf{\name{}}                  & \textbf{59.2} & \textbf{61.4} & \textbf{62.4} & \textbf{63.2} \\
\bottomrule
\end{tabular}
\vspace{10pt}
\caption{Comparison of instruction-tuned model performance on MBPP using training data from different sources and generation methods. For each data scale (1k, 2k, 10k, 100k), models are fine-tuned on equally sized datasets constructed via various data synthesis methods, as well as the common training set for coding task. “Temp. Sampling” is abbreviated from “Temperature Sampling”. “Vanilla Data” denotes the Code Alpaca training sets.}
\label{tab:mbpp-full}
\end{table}

% Human Eval

\begin{table}[htbp]
\centering
\small
%\caption{HumanEval results (\% pass@1).}
\begin{tabular}{
    l                     % Model
    l                     % Gen. Model
    l                     % Dataset
    c c c c               % 1k, 2k, 10k, 100k
}
\toprule
\textbf{Model} & \textbf{Gen.\ Model} & \textbf{Dataset Scaling} & \textbf{@1k} & \textbf{@2k} & \textbf{@10k} & \textbf{@100k}\\
\midrule
%%%%%%%%%%%%%%%%%%%%%%%%%%%%%%%%%%%%%%%%%%%%%%%%%%%%%%%%%%%%%%%%%%%%%%%%%%
\multirow{13}{*}{\texttt{LLaMA3.1-8B}}
& Code Alpaca Python subset & Vanilla Data          & -  & 43.29 & -  & -  \\ 
\cmidrule(l){2-7}
& \multirow{4}{*}{\texttt{GPT-4o}}
& Evol-Instruct                         & 44.51 & 49.39 & 51.22 & 51.22 \\
&                                       & Persona Hub                         & 43.90 & 45.12 & 48.78 & 49.39 \\
&                                       & Temp.\ Sampling                     & 41.46 & 45.73 & 46.95 & 47.56 \\
&                                       & \textbf{\name{}}                  & \textbf{48.78} & \textbf{50.00} & \textbf{51.83} & \textbf{53.05} \\
\cmidrule(l){2-7}
& \multirow{4}{*}{\texttt{LLaMA3.3-70B-Instruct}}
& Evol-Instruct                         & 40.85 & 41.46 & 42.68 & 45.12 \\
&                                       & Persona Hub                         & 39.63 & 40.24 & 44.51 & 46.34 \\
&                                       & Temp.\ Sampling                     & 40.24 & 41.46 & 42.68 & 45.12 \\
&                                       & \textbf{\name{}}                  & \textbf{47.56} & \textbf{48.17} & \textbf{50.00} & \textbf{54.27} \\
\cmidrule(l){2-7}
& \multirow{4}{*}{\texttt{Qwen2.5-72B-Instruct}}
& Evol-Instruct                         & 40.24 & 42.07 & 43.29 & 46.34 \\
&                                       & Persona Hub                         & 41.46 & 42.07 & 45.12 & 46.95 \\
&                                       & Temp.\ Sampling                     & 39.63 & 42.68 & 45.73 & 46.34 \\
&                                       & \textbf{\name{}}                  & \textbf{46.34} & \textbf{48.17} & \textbf{50.61} & \textbf{53.05} \\
\midrule
%%%%%%%%%%%%%%%%%%%%%%%%%%%%%%%%%%%%%%%%%%%%%%%%%%%%%%%%%%%%%%%%%%%%%%%%%%
\multirow{13}{*}{\texttt{Qwen2.5-7B}}
& Code Alpaca Python subset & Vanilla Data          & -  & 77.44 & -  & -  \\ 
\cmidrule(l){2-7}
& \multirow{4}{*}{\texttt{GPT-4o}}
& Evol-Instruct                         & 78.05 & 80.49 & 80.49 & 81.10 \\
&                                       & Persona Hub                         & 76.22 & 77.44 & 78.66 & 79.88 \\
&                                       & Temp.\ Sampling                     & 79.27 & 80.49 & 81.10 & 81.71 \\
&                                       & \textbf{\name{}}                  & \textbf{79.88} & \textbf{80.49} & \textbf{81.49} & \textbf{83.05} \\
\cmidrule(l){2-7}
& \multirow{4}{*}{\texttt{LLaMA3.3-70B-Instruct}}
& Evol-Instruct                         & 75.61 & 76.83 & 77.44 & 78.05 \\
&                                       & Persona Hub                         & 75.61 & 75.00 & 76.22 & 78.05 \\
&                                       & Temp.\ Sampling                     & 76.22 & 76.83 & 77.44 & 77.44 \\
&                                       & \textbf{\name{}}                  & \textbf{77.44} & \textbf{78.05} & \textbf{79.88} & \textbf{81.71} \\
\cmidrule(l){2-7}
& \multirow{4}{*}{\texttt{Qwen2.5-72B-Instruct}}
& Evol-Instruct                         & 76.22 & 76.22 & 78.66 & 81.71 \\
&                                       & Persona Hub                         & 76.22 & 76.22 & 76.83 & 77.44 \\
&                                       & Temp.\ Sampling                     & 75.00 & 78.05 & 78.66 & 80.49 \\
&                                       & \textbf{\name{}}                  & \textbf{77.44} & \textbf{78.66} & \textbf{79.88} & \textbf{82.32} \\
\bottomrule
\end{tabular}
\vspace{10pt}
\caption{Comparison of instruction-tuned model performance on HumanEval using training data from different sources and generation methods. For each data scale (1k, 2k, 10k, 100k), models are fine-tuned on equally sized datasets constructed via various data synthesis methods, as well as the common training set for coding task. “Temp. Sampling” is abbreviated from “Temperature Sampling”. “Vanilla Data” denotes the Code Alpaca training sets.}
\label{tab:humaneval-full}
\end{table}

\subsection{Case Study}
\begin{figure*}[t]
\centering
\includegraphics[page=1, width=\linewidth]{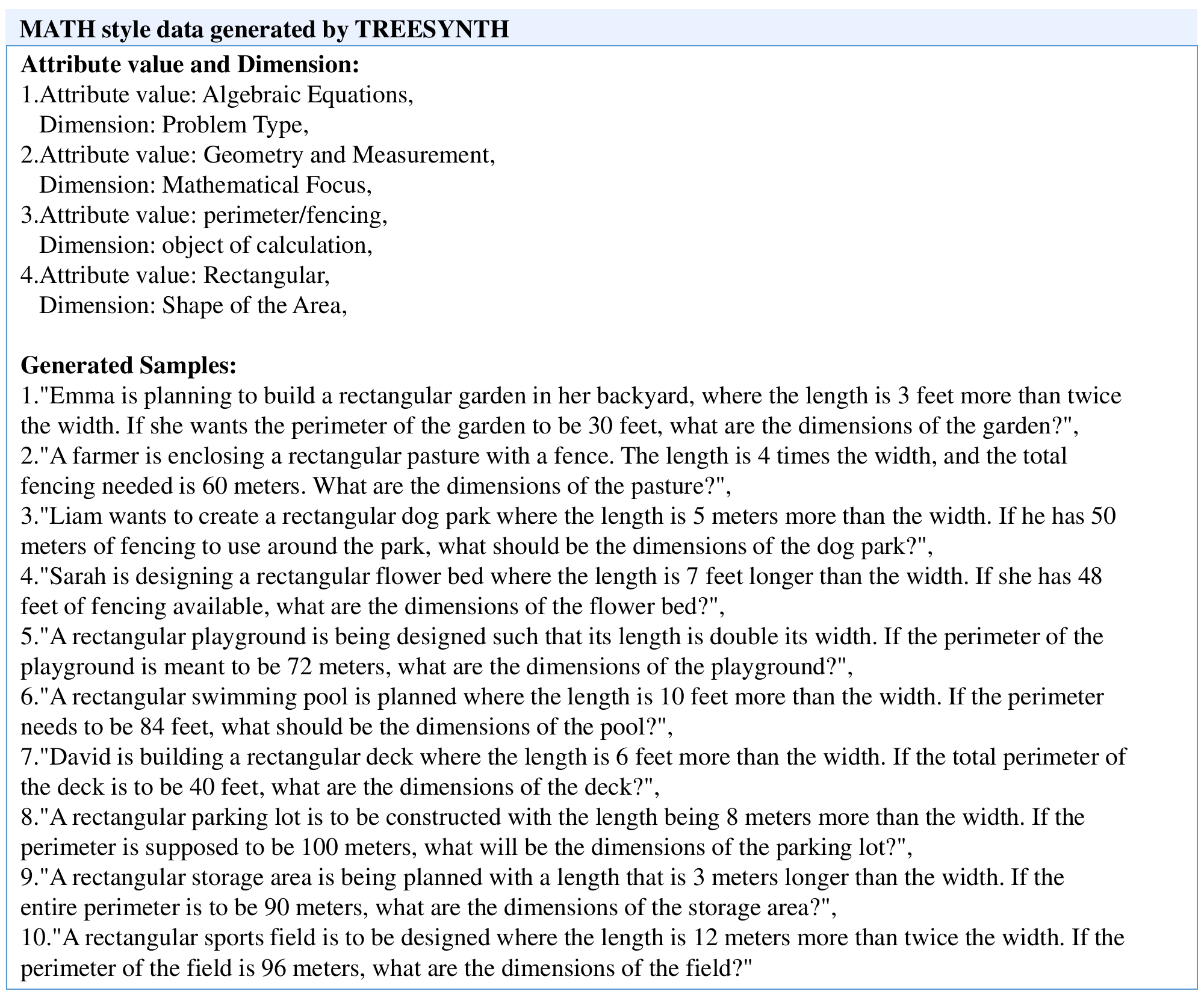}
\caption{
GSM8K style data generated by \name{}.
}
\label{fig:GSM}
\end{figure*}

\begin{figure*}[t]
\centering
\includegraphics[page=2, width=\linewidth]{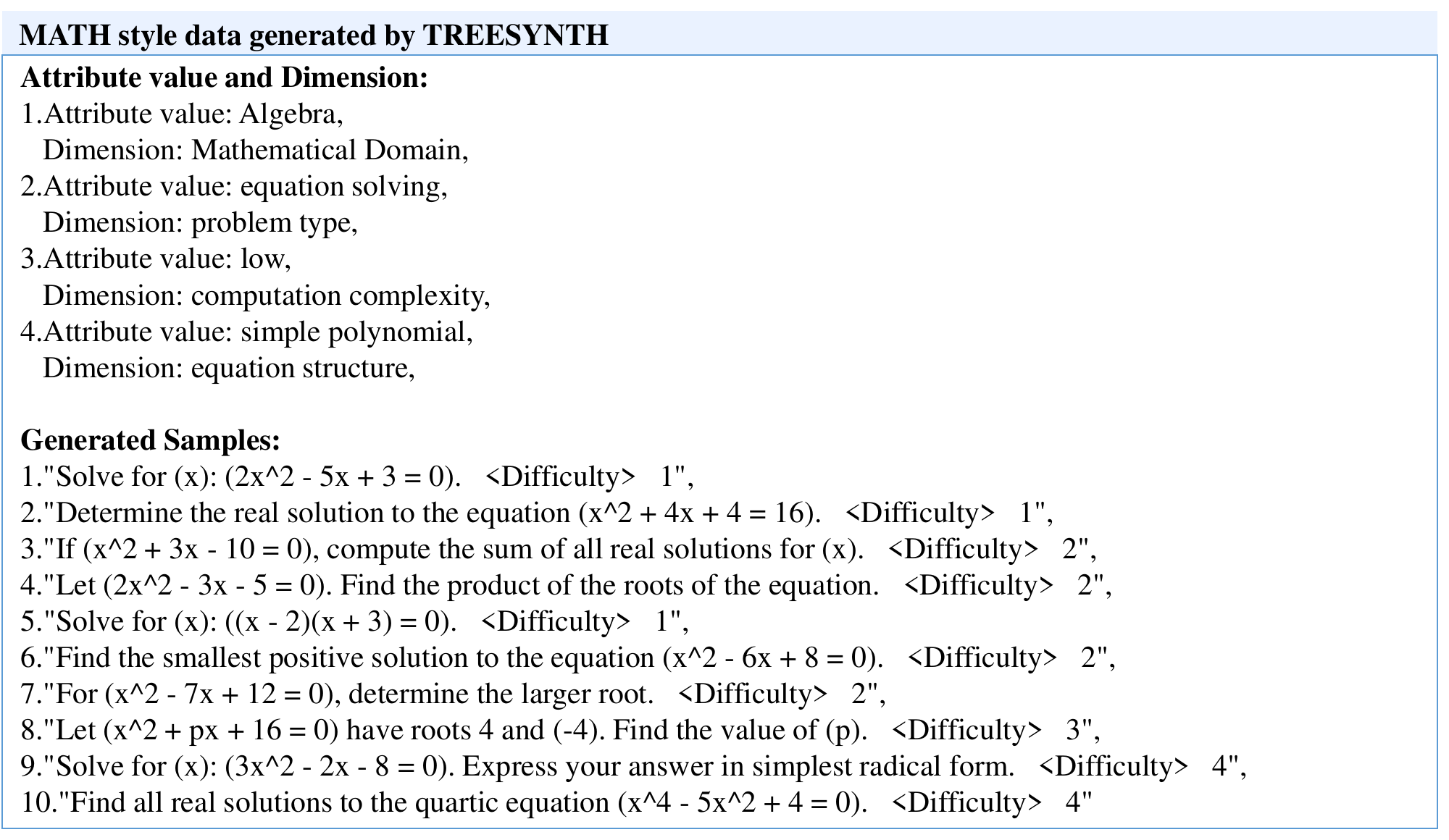}
\caption{
MATH style data generated by \name{}.
}
\label{fig:MATH}
\end{figure*}

\begin{figure*}[t]
\centering
\includegraphics[page=3, width=\linewidth]{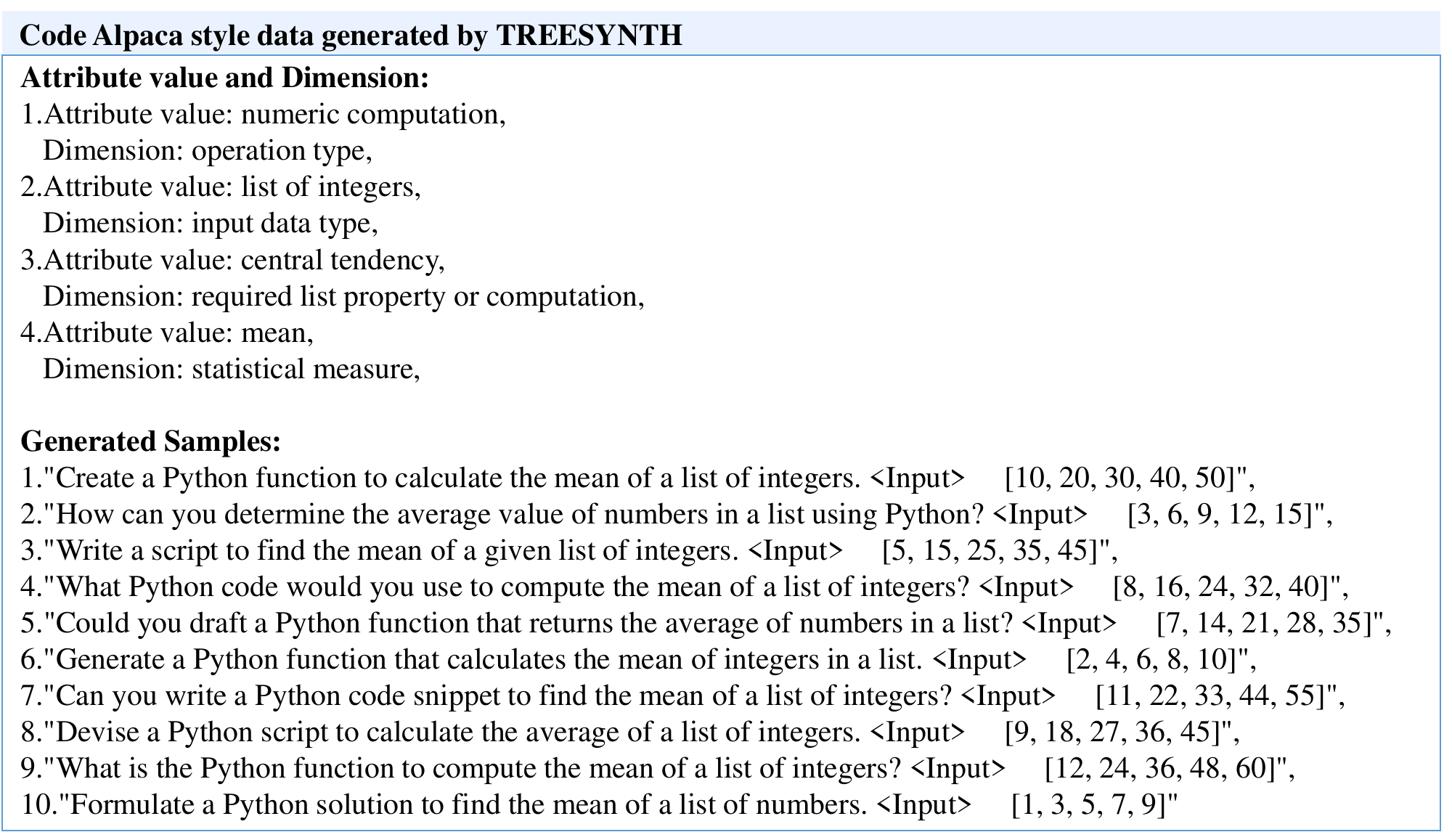}
\caption{
Code Alpaca style data generated by \name{}.
}
\label{fig:Code}
\end{figure*}

\textbf{The data generation process of \name{} demonstrates both high controllability and interpretability.} As illustrated in Figures~\ref{fig:GSM}, \ref{fig:MATH}, and \ref{fig:Code}, the GSM8K, MATH, and Code Alpaca-style datasets generated by \name{} are presented alongside their corresponding criteria  and attribute values. The generated data not only preserves the stylistic features of the original datasets but also strictly adheres to specified attribute values. This highlights \name{}'s key strength: by leveraging attribute values associated with subspaces obtained through data space partitioning, the approach achieves precise control over data generation within each subspace, thereby ensuring effective regulation and interpretability of the synthesis process.

\subsection{Limitations}
\label{sec:limitations}
This paper introduces \name{}, a tree-guided subspace-based synthesis approach that systematically partitions the data space from a global perspective to produce diverse and comprehensive instruction sets. After generating these instructions, LLMs are utilized to synthesize corresponding answers. However, following the practice of Evol-Instruct~\citep{xu2023wizardlm}, the accuracy validation of answers remains unexplored. Despite this potential defect, \name{} still demonstrates consistent improvements on both data diversity and downstream task performance, highlighting its superior effectiveness and robust scalability.

\subsection{Experiments Compute Resources}
\label{sec: Experiments compute resource}
% The computational experiments consume approximately 30,000 GPU-hours using a high-performance computing node configured with the following specifications: eight NVIDIA H100 SXM GPUs (80GB HBM3 each) in a parallel configuration, supported by 128-core CPU and 2TB of system RAM. The software environment utilizes the CUDA 12.4 acceleration approach with the PyTorch 2.6.0 deep learning library.
All experiments are executed on high-performance computing node equipped with eight NVIDIA H100 SXM GPUs (80 GB HBM3 each), dual-socket 128-core CPUs, and 2 TB of system RAM. 
The total compute budget amounted to roughly 30,000 GPU-hours.  
The software stack comprised PyTorch 2.6.0 linked against CUDA 12.1 (NCCL 2.17.1).

\subsection{Broader Impacts} \label{sec:Broader impacts}
\name{} promotes AI advancement by autonomously generating diverse and balanced datasets, reducing dependence on costly human curation and mitigating biases imposed by models, seed data and low-variation prompts. This is important for LLM customization and further enhancing their specific capabilities, especially on domains without source data.

\clearpage

% \begin{figure*}[!ht]
%     \centering
%     \subfloat[Similarity]{
%         \includegraphics[width=1.0\linewidth]{images/Figure_sim.pdf}
%         \label{fig: similarity}
%     }        
%     \caption{xxx}
%     \label{fig:cos_similarity_distribution}
% \end{figure*}

%温度的提示词

%tree数据生成的提示词

%维度确定

%空间覆盖

%例子

\clearpage

\end{document}